\pdfoutput=1
\documentclass[twoside,11pt]{article}

\usepackage[preprint]{jmlr2ekl}
\usepackage{amsmath}
\usepackage{amsthm}

\usepackage{booktabs}
\usepackage{array}
\usepackage{enumitem}
\ShortHeadings{Aggregation Rules from Preference Dynamics}{Vojnovi\'c}
\hypersetup{pdftitle={Aggregation Rules Induced by KL-Regularised Preference Dynamics on Networks},
  pdfauthor={Milan Vojnovi\'c}}
\firstpageno{1}

\newcommand{\proofend}{\hfill\ensuremath{\square}\par\medskip}
\newtheorem{theorem}{Theorem}[section]
\newtheorem{proposition}[theorem]{Proposition}
\newtheorem{lemma}[theorem]{Lemma}
\newtheorem{corollary}[theorem]{Corollary}
\theoremstyle{definition}
\newtheorem{definition}[theorem]{Definition}
\newtheorem{assumption}[theorem]{Assumption}
\newtheorem{example}[theorem]{Example}
\theoremstyle{remark}
\newtheorem{remark}[theorem]{Remark}

\newcommand{\Y}{\mathcal{Y}}
\newcommand{\KL}{D_{\mathrm{KL}}}
\newcommand{\Pone}{\Pi_{\pi}}
\newcommand{\osc}{\operatorname{osc}}
\newcommand{\E}{\mathbb{E}}
\newcommand{\one}{\mathbf{1}}
\newcommand{\Alg}{U}
\newcommand{\Frz}{F}
\newcommand{\AUU}{A_{UU}}
\newcommand{\AUF}{A_{UF}}

\title{Aggregation Rules Induced by KL-Regularised\\
       Preference Dynamics on Networks}

\author{\name Milan Vojnovi\'c \email m.vojnovic@lse.ac.uk \\
       \addr Department of Statistics\\
       London School of Economics and Political Science\\
       London, United Kingdom}

\begin{document}
\maketitle

\begin{abstract}
We study agents that each align a model to their own preference data while
observing one another, and ask which aggregation rule the resulting policies
implement. Each round, an agent takes one KL-regularised step. The step involves a
reference that the agent is penalised for leaving and a challenger against which
its responses are compared, and each of the two may be the agent's own or an
average over the agents it observes; attention need not be reciprocal.
Aggregating the reference makes the rule collective. Agent $v$ then has the weight
$\pi_v/\beta_v$, its share of the network's long-run attention divided by its
temperature. If, in addition, the challenger is fixed, the agents concentrate on
the response maximising $\sum_v(\pi_v/\beta_v)r_v(y)$, where $r_v$ is agent $v$'s
reward, and the rule is cardinal. If the reference is fixed and the challenger
aggregated, the agents' policies at a fixed point generally differ, and any
weighted mean of them is an approximate maximal lottery whatever the weights, so
the dynamics select no weighting. If both are aggregated, the rule is ordinal:
averaged over rounds, the collective policy is a maximal lottery of the
influence-weighted preference kernel, up to a slack that we bound. Where a
collective rule exists, its weights depend only on the attention weights and the
temperatures.
\end{abstract}

\begin{keywords}
AI alignment, preference aggregation, probabilistic social choice, maximal
lotteries, reinforcement learning from human feedback, social learning
\end{keywords}

\bigskip
\section{Introduction}
\label{sec:intro}

Consider $n$ agents with private preferences over a common set of responses, for instance the candidate responses of a language model to one prompt.
Each agent observes the current positions of some of the others, and repeatedly
revises its own position in the direction its preferences favour while staying
close to some anchor. The agents need not agree, coordinate or share their data.
Each has access only to the others' current positions, and uses them. We ask what
such a process converges to, and whose preferences the limit encodes.

The question arises in several parts of the alignment of learned systems.
Reinforcement learning from human feedback (RLHF)~\citep{ziegler2019,ouyang2022}
optimises a policy against a reward model fitted to labels from many annotators.
In federated and privacy-preserving training~\citep{fan2025fedrlhf,wu2024fedrlhf},
reward signals stay inside institutions that cannot pool raw data. Constitutional
AI~\citep{bai2022constitutional} evaluates behaviour against written principles
that act as a fixed external standard, and in multi-agent
deployments~\citep{carichon2025multiagent} models with different objectives see one
another's outputs, so that each one's preferences reach the others indirectly.
Nobody writes down the rule by which the competing preferences are combined, and
what does the combining varies. In RLHF it is the reward model's fit to the
labels, and in Constitutional AI the wording of the principles; in both, the
preferences are reconciled once, before the policy is optimised. In federated
training and in multi-agent deployments the reconciling happens through the
updating itself, as the agents repeatedly revise against what the others
currently hold. This paper is about the second kind, and asks which aggregation
rule that process implements.

\subsection{The dynamics}
\label{sec:intro-dynamics}

Each agent is a node $v$ of a directed graph. It holds a \emph{belief} $q_{v,t}$, a distribution over the responses, which it revises at each round $t$. In the applications above, an agent's belief is the policy it would use to answer the prompt. An agent's preference data are summarised by a kernel $P_v$, where
$P_v(y,y')$ is the probability that its annotators prefer $y$ to $y'$. Only comparisons enter, and no absolute scale of merit is assumed. One agent affects another only through the weight $a_{vu}\geq0$ that $v$
places on $u$'s belief, and $v$ need not weigh $u$ as $u$ weighs
$v$. Section~\ref{sec:setup} gives the construction in full, and
Appendix~\ref{app:modelling} the idealisations it rests on.

\paragraph{One round.}
A round of revision at agent $v$ trades an improvement in expected quality
against a penalty for moving away from an anchor:
\begin{equation}\label{eq:intro-update}
  q_{v,t+1}\;=\;\arg\max_{p}\;\Bigl\{\;
  \underbrace{\E_{y\sim p}\bigl[\mathrm{quality}_{v,t}(y)\bigr]}_{\text{do better}}
  \;-\;\beta_v\,
  \underbrace{\KL\bigl(p\,\big\|\,\mathrm{anchor}_{v,t}\bigr)}_{\text{but stay close}}
  \;\Bigr\}.
\end{equation}
The anchor and the standard on the right are fixed from what is known at the end of round $t$. The temperature $\beta_v>0$ is the exchange rate between the two terms. With a large $\beta_v$ an agent stays close to its anchor, and with a small one it optimises hard and moves away from it. Because the maximiser has a closed form, a round is an explicit reweighting of the anchor by the quality signal, and no approximation enters here. The four dynamics differ in where the anchor and the standard come from, the subject of the next paragraph. The kernels, temperatures, anchors and attention weights are free
parameters within each of them.

Two distributions are needed to make \eqref{eq:intro-update} concrete, and each
may be supplied in either of two ways.
\begin{itemize}[leftmargin=*,itemsep=2pt]
  \item The \emph{reference} is the anchor in the penalty term: what the agent
        must stay close to. It is either \emph{fixed}, a checkpoint the
        agent has validated and will not abandon, or \emph{aggregated},
        that is, recomputed at each round from the current beliefs of the
        agents it observes.
  \item The \emph{challenger} is the distribution that quality is measured against. Preference
        data are comparisons, so quality is defined only relative to a
        baseline: the quality of $y$ is the probability that $y$ is preferred to
        a response drawn from the challenger. It too is either \emph{fixed}, as with a frozen evaluator, or \emph{aggregated} from the neighbours' current beliefs. In standard RLHF the checkpoint never moves; an aggregated reference replaces
it each round by the current consensus of the agents $v$ observes, so the penalty holds an agent near where its neighbours currently are rather
than near a fixed distribution.
\end{itemize}
The maximisation in \eqref{eq:intro-update} has a closed form, and its
solution at agent $v$ is
\begin{equation}\label{eq:intro-explicit}
  q_{v,t+1}(y)\;\propto\;
  \underbrace{\mathcal{R}_{v,t}(y)}_{\text{reference}}\;
  \exp\Bigl(\frac{1}{\beta_v}
  \underbrace{\textstyle\sum_{y'}\mathcal{C}_{v,t}(y')\,P_v(y,y')}
   _{\text{how often }y\text{ beats the challenger}}\Bigr),
\end{equation}
in which $\mathcal{R}_{v,t}$ is either the agent's fixed reference or a weighted
average of its neighbours' beliefs, and $\mathcal{C}_{v,t}$ either its fixed
challenger or a weighted average of the same beliefs. The two averages differ, and Section~\ref{sec:setup} gives them and explains why each role calls for its own. Both use the weights $a_{vu}$ and require only the neighbours' current beliefs. No data, priors or temperatures are exchanged.
Appendix~\ref{app:modelling} reads one application of
\eqref{eq:intro-explicit} as a complete round of local training rather than a
single optimiser step.

\paragraph{The four combinations.}
Each combination induces a different aggregation rule. A fixed challenger means
the standard of quality comes from outside the network and does not move, and
the rule that results is sensitive to how large the agents' rewards are. An
aggregated challenger means the standard of quality is the agents' own current behaviour. Each agent still maximises every round, but the limit is then characterised by
an equilibrium condition, and only the comparisons matter. With a fixed reference, this is the network form of Nash learning from human feedback~\citep{munos2023}. A fixed reference ties each agent to its own starting point, and the others reach it only
through the quality signal. An aggregated reference lets the agents pull each other's anchors, and influence then becomes global.

\paragraph{Influence.}
One quantity attached to the graph appears throughout.
Consider the random walk that steps from an agent $v$ to one of the agents $v$
observes, moving from $v$ to $u$ with probability $a_{vu}$. If every agent can
be reached from every other, this walk has a unique
stationary distribution $\pi$ over the agents, and we call $\pi_v$, the
fraction of time it spends at $v$, the \emph{influence} of agent $v$. It is large when $v$ is observed by agents that
are themselves widely observed, it is computable from the weights alone with no
access to anyone's preference data, and it equals $1/n$ for every agent exactly when the weights are doubly
stochastic, each agent receiving in total as much attention as it gives.

Appendix~\ref{app:deployments} describes the deployments these dynamics
cover and where asymmetric attention arises.

\subsection{The four rules on one small instance}
\label{sec:intro-example}

The four assignments of the two roles give four different answers on identical data. Take three agents and three responses in the shape a federated deployment
has: a central model and two others, each of which places $a=0.2$ of its
attention on the centre while the centre divides $b=0.6$ between them. The
centre therefore attends to each of the others slightly more than it is
attended to, and is the \emph{least} influential of the three, $\pi=(\tfrac28,\tfrac38,\tfrac38)$. It is the hub-and-spoke graph of Section~\ref{sec:running} with two spokes. Let the central agent prefer $y_0$
strongly and the other two prefer $y_1$ mildly, with rewards $r_1=(6,0,0.3)$ and
$r_2=r_3=(0,1,0.5)$, equal temperatures and uniform priors. Each agent's
comparison probabilities follow from its rewards in the standard
way~\citep{bradley1952}: $P_v(y,y')=1/\bigl(1+e^{-(r_v(y)-r_v(y'))}\bigr)$, the logistic function of the reward difference.
In this instance that gives the two mild agents $P(y_1,y_0)=0.73$, from their
reward gap of $1$, and the central agent $P(y_0,y_1)=0.998$, from its gap of
$6$. The scale of the rewards is therefore visible in the kernel only through
how close to certainty it pushes the comparison, and no gap, however large, can
push it past one.

\begin{figure}[t]
\centering
\includegraphics[width=0.95\textwidth]{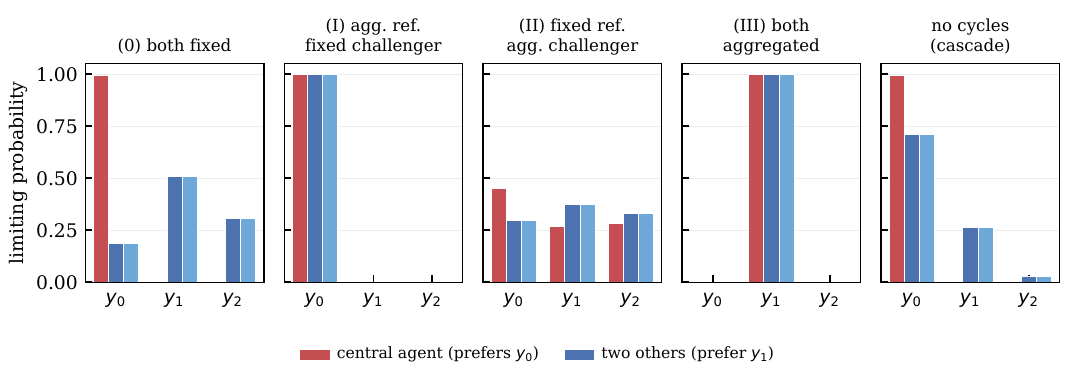}
\caption{Limiting beliefs of the three agents of
Section~\ref{sec:intro-example} under the four assignments of the two roles
and under the acyclic variant, described in the text. Bars are each agent's
limiting probability on each response, the central agent in red.}
\label{fig:intro}
\end{figure}

The five panels read as follows. With both roles fixed, each agent reaches the
distribution that maximises its own reward against its own starting point, an
exponential tilt of that anchor by its reward, and the three disagree. With the reference aggregated and the
challenger fixed, all three concentrate on $y_0$, the central agent's preference, because the rule uses the sizes of the rewards. With the reference
fixed and the challenger aggregated, the agents reach separate interior
limits and the disagreement is permanent. With both aggregated, they concentrate on $y_1$. The rule then uses the preferences only through the comparisons they induce, whose probabilities cannot exceed one, so a reward counts for no more than a sure win. The fifth panel keeps the roles and varies the graph. The central model is a frozen checkpoint that the others distil from. It never updates, and influence flows one way. Each downstream agent then ends at a compromise,
displaced towards a response it disprefers.

In Figure~\ref{fig:intro}, the two panels with an aggregated reference disagree about the winner on the
same data, because one rule uses the rewards and the other only the comparisons they induce. The two with a fixed reference produce no collective outcome at all, because an agent that keeps its own anchor combines only the beliefs it observes directly. No network-wide weighting such as $\pi$ appears in its limit.

The disagreement between the panels is the cardinal--ordinal distinction,
which with three responses can be seen directly. With a fixed challenger and equal temperatures the agents collectively
maximise the reward sum $\sum_v\pi_vr_v(y)$, as Theorem~\ref{thm:case3} shows: the
central agent contributes $0.25\times6=1.5$ towards $y_0$, the other two
$0.75\times1=0.75$ towards $y_1$, and $y_0$ wins. With an aggregated challenger
the signal is instead a probability of winning a comparison, which cannot exceed
one, so an agent's contribution is bounded however strong its preference. The
most the central agent could ever say is ``$y_0$ always beats $y_1$'', worth
$\tfrac12$ above the neutral value, and weighted by its influence that is
$0.25\times\tfrac12=0.125$. The other two are less emphatic. Their kernels give $y_1$ a winning
probability of $0.73$ against $y_0$, that is $0.23$ above neutral, but between them they hold three quarters of the influence, so
$0.75\times0.23=0.17$, and $y_1$ wins. Which side of such a comparison a response falls on is exactly what decides, by
Section~\ref{sec:case4-further}, whether the agents can concentrate on
it: the winner's atom is locally attracting and every loser's is repelling.

Deleting the two edges back to the central model, so that it is a frozen
checkpoint the others distil from, changes the answer again. Each downstream agent ends at $0.71$ on $y_0$ against $0.26$ on $y_1$, pulled towards a response
it disprefers. No $\pi$ appears, because with influence flowing one way the attention matrix is
reducible and has no stationary distribution. Instead a downstream agent takes the checkpoint's
position in full and adds its own preference, amplified by the reciprocal of the attention it pays upstream: with weight $a=0.2$ on the checkpoint, it counts its own preference five times over. Even so, the checkpoint's preference
for $y_0$ is strong enough to prevail.

\subsection{Summary of results}
\label{sec:answers}

The same agents with the same preferences reach different limits according
to which role is aggregated. The reference decides whether the preferences are
aggregated into one rule for the network or into one rule per agent; the challenger decides what that rule is computed from: the magnitudes of the rewards when it is fixed, which makes the rule cardinal, or only the comparisons they induce when it is aggregated, which makes it ordinal. With both roles fixed the agents never
interact, which serves as a baseline. The three remaining cases give the
following.

\begin{itemize}[leftmargin=*,itemsep=3pt]

\item \emph{Aggregated reference, fixed challenger}
(Section~\ref{sec:case3}). Every agent is driven onto the maximiser of $\sum_v(\pi_v/\beta_v)r_v(y)$, or
onto the set of maximisers when there are several, a scoring rule in which agent $v$ has weight $\pi_v/\beta_v$ (Theorem~\ref{thm:case3}). A single agent that does not update halts that collapse, provided every other
can reach it, leaving every agent with a limit that puts positive mass on
every response (Theorem~\ref{thm:case3-partial}).

\item \emph{Fixed reference, aggregated challenger}
(Section~\ref{sec:case2}). Each agent remains anchored to its own $\mu_v$. The only aggregation is then of the beliefs it observes, weighted by its own row of $A$, and the agents need not agree at a fixed point (Proposition~\ref{prop:case2-fp}). The fixed point is unique, and attracts every trajectory geometrically, when $\rho(B^{-1}A)<2$ (Propositions~\ref{prop:case2-fp}(b) and~\ref{prop:case2-geom}), for instance when every $\beta_v>\tfrac12$; with a shared kernel and reversible attention it is unique at every temperature (Proposition~\ref{prop:case2-fp}(d)).  Any weighted
mean of the agents' beliefs at a fixed point is an approximate maximal
lottery of the correspondingly weighted kernel
(Theorem~\ref{thm:case2-lottery}). Unlike in case~(III), where the dynamics select $\pi_v/\beta_v$, the guarantee holds for every weighting and selects none.

\item \emph{Both roles aggregated} (Section~\ref{sec:case4}). The rule becomes ordinal, and interior agreement needs a condition that is overdetermined once the agents differ (Proposition~\ref{prop:case4-fp}). The guarantee that needs no further condition concerns the collective policy, the $\pi$-weighted geometric mean averaged over rounds, which is a maximal lottery of the $\pi_v/\beta_v$-weighted kernel up to a bounded slack (Theorem~\ref{thm:cesaro}). The network thus implements a rule that no participant holds at any round. Above a threshold on the temperatures, set by how far the graph is from
making one round of averaging bring all the agents together, their disagreement contracts geometrically to a residual set by how far their
kernels and temperatures are from agreeing, and so to consensus when they
share both (Theorem~\ref{thm:case4-directed}). The beliefs themselves converge to a single response when it dominates every column of every agent's kernel (Proposition~\ref{prop:case4-atom}), or, above a temperature threshold, every column of the influence-weighted kernel by more than twice the residual disagreement (Proposition~\ref{prop:case4-aggdom}). A Condorcet winner of the influence-weighted kernel attracts them from an explicit neighbourhood, and where that kernel compares every pair strictly and has no Condorcet winner, every atom repels them (Proposition~\ref{prop:case4-condorcet}).

\end{itemize}

\noindent
Two consequences follow wherever the reference is aggregated, and both are
computable from the communication weights before the protocol runs. An agent
whose reward model is wrong, whether through bad data or deliberate manipulation, moves the outcome by $O(\pi_v)$ rather than $O(1/n)$, so one
well-placed agent can determine the outcome of an arbitrarily large network;
and, with a fixed challenger, an agent that does not update fixes where the
others end (Theorem~\ref{thm:case3-partial}), which with an aggregated
challenger it need not (Remark~\ref{rem:frozen-several}). Where influence flows only one way, no single rule governs the network. Each agent's limit is fixed by the agents upstream of it, weighted by the directed paths that reach it.

The results above being theoretical,
Section~\ref{sec:case3-experiment} illustrates one of them on an instance in which the agents optimise
parametric policies rather than attaining their maximisers exactly, a setting
the analysis does not cover: three language models aligned against three
attributes of a human-rated preference dataset arrive at the response the
weighted rule of Theorem~\ref{thm:case3} identifies, in two configurations
that give the same agent its weight by different means, one through its
temperature and one through the attention it receives.

\subsection{Relation to prior work}
\label{sec:positioning}

All four cases are multi-agent extensions of a single-agent primitive, and at $n=1$ they coincide. The graph is then a self-loop, and reference and challenger are
the agent's own belief whether aggregated or not. What remains is
KL-regularised RLHF~\citep{ziegler2019,ouyang2022} with a fixed challenger,
and Nash learning from human feedback~\citep{munos2023} without one, whose results we build on. The graph contributes averaging. An agent's reference is the weighted
geometric mean $\prod_uq_u^{a_{vu}}$ of its neighbours' beliefs, so the
averaging takes place in log-space. It is the logarithmic pooling
of~\citet{genest1986} and the product of experts of~\citet{hinton1999} rather
than the arithmetic mixing $\sum_ua_{vu}q_u$ of DeGroot learning and opinion
dynamics~\citep{degroot1974,golub2010}.

Those literatures study beliefs about an underlying state of the world, and
ask whether the agents converge on it and how fast. Here the agents hold preferences, and there is no ground truth for them to estimate. The question is instead which aggregation of their conflicting
preferences the dynamics implement and what weight it gives each agent, with the answer fixed by two implementation choices. Nobody chooses the rule explicitly. Section~\ref{sec:related} discusses the literature in full and
sets out what is new.

\subsection{Organisation}
\label{sec:approach}

Section~\ref{sec:setup} fixes the model, ending with the structure the four
dynamics share. Sections~\ref{sec:case3} to~\ref{sec:case4} analyse the three
non-degenerate cases in turn: the aggregated reference with a fixed
challenger, the fixed reference with an aggregated challenger, and both
aggregated. Section~\ref{sec:related} compares the results with the
literature on preference learning, opinion dynamics and social choice, and
Section~\ref{sec:conclusion} draws the conclusions and states the questions
left open. Appendix~\ref{app:proofs} holds the notation, the proofs of everything stated
in the body, and the details of the experiment. The Supplementary Material
follows the same division as the body and collects further results, the
graphs that Assumption~\ref{ass:graph} excludes, and the examples.
\section{The Model}
\label{sec:setup}

The definitions used throughout come first. The graph and the objects derived
from it follow, then the update rule and the four dynamics it generates, then
the assumptions the rule embodies, and finally the example used throughout.

\subsection{Two definitions used throughout}
\label{sec:background}

Two definitions and two closed forms from social choice and single-agent alignment are used throughout.

\begin{definition}[Preference kernel]
\label{def:kernel}
A \emph{preference kernel} on $\Y$ is a function $P:\Y\times\Y\to[0,1]$ with
$P(y,y')+P(y',y)=1$. Write $R(y;q)=\E_{y'\sim q}[P(y,y')]$ for the
probability that $y$ beats a response drawn from $q$. We call
$R(\cdot\,;q)$ the \emph{quality signal} against $q$. Every use of ``signal'' below refers to this function, and
$R_v(\cdot\,;\mathcal C_{v,t})$ is the signal agent $v$ faces at round $t$.
\end{definition}

\begin{definition}[Maximal lottery]
\label{def:maxlot}
A distribution $p$ on $\Y$ is a \emph{maximal lottery} of $P$ if
$\E_{y\sim p,\,y'\sim q}[P(y,y')]\geq\tfrac12$ for every $q\in\Delta(\Y)$,
and an \emph{$\varepsilon$-maximal lottery} if the same holds with
$\tfrac12-\varepsilon$ on the right.
\end{definition}

\noindent
Maximal lotteries were introduced by \citet{kreweras1965} and \citet{fishburn1984}. One always exists, by von Neumann's minimax theorem. It is Condorcet-consistent and ordinal, and maximal lotteries are the only probabilistic social choice rule satisfying a pair of consistency axioms~\citep{brandl2016}. The same
object appears as the von Neumann winner of duelling
bandits~\citep{dudik2015} and as the Nash equilibrium that Nash learning from
human feedback converges to~\citep{munos2023,maura2025}: a maximal lottery of
$P$ is exactly a Nash equilibrium of the symmetric zero-sum game with payoff
$P-\tfrac12$, which is the game a single agent plays against itself in the
limit of no regularisation.

Two closed forms are used throughout: for a reward $r:\Y\to\mathbb{R}$,
temperature $\beta>0$ and reference $q_0\in\Delta(\Y)$, the unique maximiser
of $\E_p[r]-\beta\KL(p\|q_0)$ is the \emph{Gibbs policy}
\begin{equation}\label{eq:gibbs}
  p^*(y)=q_0(y)e^{r(y)/\beta}\Big/\textstyle\sum_{y'}q_0(y')e^{r(y')/\beta},
\end{equation}
and replacing the reward by the quality signal against the policy itself
gives the \emph{regularised Nash policy}, the fixed point
\begin{equation}\label{eq:nash}
  p^*(y)\propto q_0(y)\exp\bigl(\beta^{-1}R(y;p^*)\bigr),
\end{equation}
which is the single-agent object the network generalises.

\subsection{Graph, operator, and three derived objects}

Let $G=(V,E)$ be a directed graph with $|V|=n$, the agents being its vertices
and its edges giving who attends to whom: $(u,v)\in E$ exactly when agent
$v$ places positive weight on agent $u$, so that $G$ is determined by the
attention matrix $A$ introduced below and the two are used interchangeably.
Each agent $v$ holds the following objects, fixed once and for all except the
belief.

\begin{itemize}[leftmargin=*,itemsep=1pt]
  \item A \emph{belief} $q_{v,t}\in\Delta(\Y)$, the agent's policy at round $t$,
        and the only object that evolves.
  \item A \emph{preference kernel} $P_v$ in the sense of
      Definition~\ref{def:kernel}, fixed throughout and not shared between
      agents.
  \item A \emph{temperature} $\beta_v>0$, collected into
        $B=\mathrm{diag}(\beta_v)$, weighting the KL penalty in the update of Section~\ref{sec:fourcases} and so
      fixing how far one round may carry the agent
        from its reference. A large $\beta_v$ keeps it close to that reference,
        whatever the reference happens to be. As $R_v$ takes values in $[0,1]$, $\beta_v$ is measured against a quantity of
      order one.
  \item A \emph{fixed reference} $\mu_v\in\Delta(\Y)$, used as
        $\mathcal{R}_{v,t}$ in the two dynamics whose reference is not
        aggregated. It is the distribution the agent is anchored to, and the one it holds when the
      quality signal is uninformative; operationally, its pre-trained or
      validated checkpoint.
  \item A \emph{fixed challenger} $\nu_v\in\Delta(\Y)$, used as
        $\mathcal{C}_{v,t}$ in the two dynamics whose challenger is not
        aggregated. It is the baseline that quality is measured against, and it
        determines the reward $r_v(y)=R_v(y;\nu_v)$ appearing in
        Section~\ref{sec:case3}.
\end{itemize}

\noindent
Write $\mathcal{N}^{\mathrm{in}}_v=\{u:(u,v)\in E\}$ for the agents whose
beliefs $v$ observes, and $a_{vu}\geq0$ for the weight $v$ places on $u$'s
belief, with
\begin{equation}\label{eq:weights}
  \sum_{u\in\mathcal{N}^{\mathrm{in}}_v\cup\{v\}}a_{vu}=1\qquad\text{for every }v.
\end{equation}

Collect the weights into the $n\times n$ matrix $A=(a_{vu})$, so that
$(A\mathbf f)_v$ is the average agent $v$ forms of any quantity $\mathbf f$
held one value per agent. By \eqref{eq:weights} $A$ is row-stochastic, $A\one=\one$. Throughout, \emph{attention} refers to these weights and \emph{influence} to
the measure $\pi$ they induce.

The constant vectors are therefore exactly what $I-A$ annihilates, which is the fact Section~\ref{sec:fixedpoints} rests on. Throughout, $A$ multiplies column vectors indexed by the agents, as in $A\mathbf f$, and $\pi$ multiplies $A$ from the left, as in $\pi^\top A$. The distinction matters, since $A\one=\one$ while $\pi^\top A=\pi^\top$.

\begin{assumption}[Influence circulates]
\label{ass:graph}
Information reaches every agent from every other, and does so at every
sufficiently late round: for some $m$, every agent can be reached from every
other along a directed path of exactly $m$ steps. Equivalently, $A^m>0$ entrywise for some
$m$: the graph is strongly connected and aperiodic, and $A$ is
\emph{primitive}. Strong connectivity alone is not enough, since a graph on
which influence alternates between two groups has $A^m$ with zeros for every $m$. Any agent retaining weight on itself rules that out.
\end{assumption}

Two consequences are used repeatedly below. First, $A$ has a unique
\emph{stationary distribution}: weights $\pi_v>0$ summing to one with
$\pi^\top A=\pi^\top$, which is the $\pi$ already described in
Section~\ref{sec:intro}. Second, repeated averaging retains only that weighting:
$A^t\to\one\pi^\top$, so after many
rounds each agent's average of any fixed quantity is the $\pi$-weighted average
of it, and the discrepancy falls off geometrically at a rate we write $|\lambda_2(A)|$, the largest modulus among the eigenvalues
of $A$ other than $1$, which
Assumption~\ref{ass:graph} makes simple and strictly dominant.

The assumption excludes two kinds of graph. One is a network in which influence fails to circulate. The conclusions differ
there, and the Supplementary Material treats two such cases. If $A$ is
reducible, the graph splits into closed components, each running these
dynamics on its own with its own stationary distribution, while the remaining
agents take a convex combination of the components' outcomes with weights
given by the probabilities that attention leads from them to each component.
If instead no agent is reachable from itself through any other, as in a
distillation cascade, there is no stationary distribution and no collective rule: each agent's limit is fixed
by the agents upstream of it, weighted by the directed paths that reach it,
and solvable in closed form by working down the graph. The weights
$\pi_v/\beta_v$ have no counterpart in either case, which is why
Assumption~\ref{ass:graph} is imposed throughout. Another is a periodic graph, on which agents are
reachable only at multiples of some number of steps. Periodicity is not only a technical nuisance. The star in which every client replaces its model by the server's and the
server by the clients' average is periodic. Giving any agent weight on itself removes it, and this exclusion costs little, unlike the failure of influence to circulate, which changes the conclusions as described above.

\begin{definition}[Symmetric and balanced weights]
\label{def:symmetric}
The weights are \emph{symmetric} if
\[
  a_{vu}=a_{uv}\qquad\text{for every pair }u,v,
\]
that is, if each agent attends to another exactly as much as it is attended to
by it; equivalently $A=A^\top$. They are \emph{balanced} if $A$ is \emph{doubly stochastic}, its columns
summing to one as well as its rows. This is weaker than symmetry and is exactly the condition $\pi=\one/n$ that
makes every agent equally influential.
\end{definition}

Symmetry is the standing assumption of the literature this paper departs from,
and dropping it is necessary for the question to arise at all. With symmetric weights every agent is equally influential, so the preferences of all agents count equally.

Two further quantities built from $A$ appear throughout.

\begin{definition}[Deviation matrix, ergodic coefficient]
\label{def:resolvent}
The \emph{deviation matrix} and \emph{ergodic coefficient} of $A$ are
\begin{equation}\label{eq:D}
  D=\sum_{t\geq0}\bigl(A^t-\one\pi^\top\bigr),
  \qquad
  \tau_1(A)=\tfrac12\max_{v,v'}\sum_u\bigl|a_{vu}-a_{v'u}\bigr| ,
\end{equation}
the sum converging by Assumption~\ref{ass:graph}.
\end{definition}

The Supplementary Material gives a form of $\tau_1$ that is easier to
evaluate, and evaluates it on several graphs as a guide to the size of the
thresholds in which it appears.

\noindent
By Assumption~\ref{ass:graph}, $A^t\mathbf f$ approaches the $\pi$-weighted
average of $\mathbf f$, and $D$ sums the differences over all rounds, so
$d_{vu}$ is the excess weight $v$'s repeated averages place on $u$ beyond
$u$'s long-run share $\pi_u$. The coefficient $\tau_1(A)\in[0,1]$ is the largest total-variation distance
between the weightings two agents use, and is small when any two attend to
similar sources. For example, it is $0$ on the complete graph with uniform
weights, where every agent averages the same way, and close to $1$ when two
agents share almost no sources. It acts below as a contraction factor~\citep[Sec.~4.3]{seneta2006}. One round of averaging multiplies the spread of any quantity across the agents by at most
$\tau_1(A)$
(Lemma~\ref{lem:dobrushin}). On the complete graph the spread is gone after
a single round, and the closer $\tau_1(A)$ is to one the slower the agents
come together.

Since $I-A$ annihilates the constant vectors it has no inverse, and $D$ stands
in for one. These are standard properties of a primitive chain, proved in
Lemma~\ref{lem:Dprops}.

\subsection{The update rule and the four dynamics}
\label{sec:cases}

Write the neighbourhood geometric and arithmetic means as
\[
  \bar q_{v,t}(y)\propto\prod_uq_{u,t}(y)^{a_{vu}},
  \qquad
  \check q_{v,t}(y)=\sum_ua_{vu}q_{u,t}(y),
\]
the bar and the check being reminders of which average is meant and carrying no
other significance. Which mean goes with which role is not arbitrary. Section~\ref{sec:aggregates} gives the reason. Restating \eqref{eq:intro-explicit}
with these aggregates named, the dynamics are
\begin{equation}\label{eq:general}
  q_{v,t+1}(y)\;\propto\;\mathcal{R}_{v,t}(y)\,
  \exp\Bigl(\tfrac{1}{\beta_v}R_v\bigl(y;\mathcal{C}_{v,t}\bigr)\Bigr),
\end{equation}
with $\mathcal{R}_{v,t}\in\{\mu_v,\bar q_{v,t}\}$ and
$\mathcal{C}_{v,t}\in\{\nu_v,\check q_{v,t}\}$. Equivalently, the step taken by agent $v$ at round $t$ is
\[
  q_{v,t+1}=\arg\max_{p\in\Delta(\Y)}
  \Bigl\{\E_pR_v(\cdot;\mathcal{C}_{v,t})
  -\beta_v\KL\bigl(p\,\|\,\mathcal{R}_{v,t}\bigr)\Bigr\} .
\]

This is a best response. Treat the agents as players of a game in which
$v$'s payoff, given what the others currently hold, is $\E_pR_v(\cdot;\mathcal{C}_{v,t})-\beta_v\KL(p\|\mathcal{R}_{v,t})$; then \eqref{eq:general} is $v$'s optimal reply to the
current profile, and \eqref{eq:general} run at every agent at once is
simultaneous best-response dynamics. Its rest points are the Nash equilibria of that game. A run is then fixed by the agents' kernels, temperatures and references together with the attention weights, with no step size or schedule to choose. Two of the paper's findings are known behaviours of such dynamics. Best reply converges when the reply map contracts and need not otherwise. The classical example of non-convergence in a related process is \citeauthor{shapley1964}'s three-by-three game, in which fictitious play cycles indefinitely~\citep{shapley1964}. The temperature threshold of Proposition~\ref{prop:case2-fp} is a contraction condition of that kind, and below it the iterates can cycle. And the time average is better behaved than the last iterate, the fact
Section~\ref{sec:cesaro} rests on.

Write $f_{v,t}=\log q_{v,t}$ for the \emph{log-belief} of agent $v$ at round $t$. Initial beliefs and fixed references are taken to put positive mass on every response, so that every log-belief is finite. The update preserves this. It is convenient to work instead with
$\tilde f_{v,t}=f_{v,t}+c_{v,t}$, in which the normalising constant has not
been subtracted, so that the shift $c_{v,t}$ depends on the agent and the
round but not on the response $y$. Nothing is lost by this. A belief determines its log-belief only up to such a shift, every statement below involves
$\tilde{\mathbf f}$ through differences at two responses or through the
softmax, and both are unchanged by it
(Remark~\ref{rem:normalisation}). In these terms,
\begin{equation}\label{eq:logspace-vec}
\begin{aligned}
  \tilde{\mathbf f}_{t+1}(y)&=A\tilde{\mathbf f}_t(y)+B^{-1}\mathbf R_t(y)
  &&\text{(aggregated reference)},\\
  \tilde{\mathbf f}_{t+1}(y)&=\log\boldsymbol\mu(y)+B^{-1}\mathbf R_t(y)
  &&\text{(fixed reference)} ,
\end{aligned}
\end{equation}
where $\log\boldsymbol\mu(y)$ and $\mathbf R_t(y)$ stack $\log\mu_v(y)$ and
$R_{v,t}(y)$ over the agents, and $B=\mathrm{diag}(\beta_v)$. The two differ only in the first term. With an aggregated reference the previous round
enters through $A$, and with a fixed one it does not enter.

\begin{remark}[Normalisation]
\label{rem:normalisation}
The normalised and unnormalised iterations differ by a $y$-independent shift at
each agent and time: writing $\Phi_{v,t}$ for the $y$-independent log-normaliser
of the update at agent $v$ and time $t$, if $f_{v,t}=\tilde f_{v,t}-c_{v,t}$ with
$c_{v,t}$ $y$-independent then
$c_{v,t+1}=\sum_ua_{vu}c_{u,t}+\Phi_{v,t}$ is again $y$-independent, so by induction the shifts cancel in every log-odds difference
$f_{v,t}(y)-f_{v,t}(y')$, and in every difference $\tilde f_{u,t}(y)-\tilde f_{v,t}(y)$ between two agents at a common $y$.
All statements below are of these two kinds, and we work with
\eqref{eq:logspace-vec}. The one quantity that does see the shifts is the
disagreement measure of Section~\ref{sec:case4-conv}, where it is treated.
\end{remark}

\label{sec:fourcases}
The four dynamics are labelled as follows, in the order in which they are treated below. The baseline in which neither role is aggregated is labelled
$(0)$.

\begin{center}
\renewcommand{\arraystretch}{1.3}
\begin{tabular}{@{}lll@{}}
\toprule
& \emph{Fixed challenger} & \emph{Aggregated challenger}\\
\midrule
\emph{Fixed reference}      & (0) decoupled Gibbs  & (II) anchored, Sec.~\ref{sec:case2}\\
\emph{Aggregated reference} & (I) peer-regularised, Sec.~\ref{sec:case3} & (III) fully endogenous, Sec.~\ref{sec:case4}\\
\bottomrule
\end{tabular}
\end{center}

Dynamics~(0) is trivial. With both roles fixed, $r_v(y)=R_v(y;\nu_v)$ does not change from round to
round, so $q_{v,t+1}\propto\mu_ve^{r_v/\beta_v}$ is independent of $t$ and
identical to \eqref{eq:gibbs} at each agent: the fixed point is reached in one
step, is unique for every vector of temperatures with $\beta_v>0$, and is unaffected by the graph. The
agents then solve $n$ separate single-agent problems of the form
\eqref{eq:gibbs}, the idealisation of KL-regularised alignment discussed in
Appendix~\ref{app:deployments}, and inherit its sensitivity to the scale of the
rewards. We use it only as a baseline.

A remark on the role of $\beta_v$. It weights an agent's quality signal against the penalty for leaving its reference. A large $\beta_v$ keeps $q_{v,t+1}$
close to $\mathcal R_{v,t}$. It does not by itself bound
$\|q_{v,t+1}-q_{v,t}\|$, since that also depends on how far the reference has
moved. Under a fixed reference the two are equivalent, as
$\mathcal R_{v,t}=\mu_v$ at every round and a large $\beta_v$ gives an agent
whose belief changes little. Under an aggregated reference a large $\beta_v$
gives an agent that tracks its neighbours' current consensus closely, and that
consensus may itself have moved a long way.

\subsection{Where the two aggregates come from}
\label{sec:aggregates}

Neither aggregate has to be posited. Both follow from a single objective, in
which agent $v$ weighs the reward it gets against each neighbour by the
attention it pays that neighbour, and pays a penalty for departing from each
neighbour's belief with the same weights:
\begin{equation}\label{eq:pairwise-objective}
  \max_{p\in\Delta(\Y)}\;\Bigl\{\,\sum_ua_{vu}\,\E_{y\sim p,\;y'\sim q_u}
  \bigl[P_v(y,y')\bigr] \;-\;\beta_v\sum_ua_{vu}\KL\bigl(p\,\|\,q_u\bigr)\Bigr\}.
\end{equation}
The first sum is linear in the neighbour's belief, so it collapses to a single
comparison against the arithmetic mean, $\E_{y\sim p}R_v(y;\check q_{v})$. The
second collapses to a single divergence against the geometric mean, since
$\sum_ua_{vu}\KL(p\|q_u)=\KL(p\,\|\,\bar q_{v})-\log Z_v$ with $Z_v$
normalising $\bar q_v$ and independent of $p$. So
\eqref{eq:pairwise-objective} is exactly \eqref{eq:general} with
$\mathcal R_{v,t}=\bar q_{v,t}$ and $\mathcal C_{v,t}=\check q_{v,t}$.

Each mean therefore follows from the term it comes from. An expectation of expectations is again an
expectation, so averaging the neighbours' rewards is the same as facing the arithmetic mixture of their beliefs. A divergence does not compose that way,
and averaging the neighbours' divergences is the same as facing one divergence
from their geometric mean. The two means are linear and logarithmic pooling. Appendix~\ref{app:deployments} places them in that literature and names the restrictions built into the update rule.

\subsection{A running example}
\label{sec:running}

One concrete graph is used throughout for illustration, the simplest
arrangement in which attention need not be reciprocal: a hub and $k$ spokes, with the weight in each direction left free, so that
either side may attend to the other more. Let $V=\{h\}\cup\{1,\dots,k\}$, so $n=k+1$. Each leaf places weight
$a\in(0,1)$ on the hub and $1-a$ on itself; the hub places $b/k$ on each leaf and
$1-b$ on itself. The ratio $a/b$ measures how much more the spokes attend to the hub than the reverse. The arrangement stands for a frontier model that many downstream deployments
track while it samples them only occasionally, or for a widely-followed annotator pool. The instance of Section~\ref{sec:intro-example} is the case
$k=2$, $a=0.2$, $b=0.6$, in which the ratio runs the other way.

The one quantity we can already compute is the influence measure. Solving
$\pi^\top A=\pi^\top$ with $\pi_h=p$ and $\pi_\ell=q$ for each leaf, the balance
at the hub reads $p=p(1-b)+kqa$, so $pb=kqa$, and with $p+kq=1$ this gives
\begin{equation}\label{eq:hub-pi}
  \pi_h=\frac{a}{a+b},\qquad \pi_\ell=\frac{b}{k(a+b)} .
\end{equation}
Two features of \eqref{eq:hub-pi} matter for what follows.
The hub's influence depends only on the ratio $a/b$ and not on $k$. A hub attended to twice as hard as it attends has weight $2/3$ whether it has ten followers or ten thousand. Its share is independent of the number of spokes.
And the weights are symmetric in the sense of Definition~\ref{def:symmetric}
only when $ka=b$, which is exactly when $A$ is doubly stochastic and
\eqref{eq:hub-pi} returns $\pi_h=1/n$. Every other choice of $a$ and $b$ gives a
graph in which every edge is matched in the opposite direction and yet the
agents are not equally influential. Section~\ref{sec:running} returns to this
graph once the results are in place.
\subsection{Fixed points of the update rule}
\label{sec:fp-solvable}
\label{sec:fixedpoints}

The fixed points of the four dynamics divide along the two roles, and that division separates the four rules: \emph{the reference determines the
weights, and the challenger determines what is weighted}. With the reference
fixed, the agents decouple and each limit is settled by
\eqref{eq:logspace-vec} on its own; with it aggregated, $A$ multiplies the
state, so the trajectory accumulates powers of $A$ and those converge to
$\one\pi^\top$ by Assumption~\ref{ass:graph}, which is why $\pi$ appears
wherever it does. Independently of that, the signal $\mathbf R^*$ contains either the magnitudes of the rewards, giving a cardinal rule, or only the
comparisons they induce, giving an ordinal one. Because the two choices are independent, the four combinations give four rules.

The rest of this subsection concerns the two dynamics with an aggregated
reference, cases~(I) and~(III), whose fixed-point equation is
\begin{equation}\label{eq:fp-compact}
  (I-A)\,\mathbf f^*(y)=B^{-1}\mathbf R^*(y)-\boldsymbol\Phi^* .
\end{equation}
The star marks evaluation at the candidate fixed point: $\boldsymbol\Phi^*$ are
its $y$-independent normalisers and $R^*_v(y)=R_v(y;\check q^*_v)$ the signal
agent $v$ perceives there. With a fixed challenger $\mathbf R^*$ is the given
reward and \eqref{eq:fp-compact} is linear in $\mathbf f^*$; with an aggregated
one $\check q^*_v$ is built from the profile, so $\mathbf R^*$ depends on
$\mathbf f^*$ and \eqref{eq:fp-compact} is a fixed-point condition. We write
$\mathbf R^*(y)$ throughout, with that dependence understood. Only interior profiles satisfy \eqref{eq:fp-compact}. Those on the boundary have infinite log-odds and are treated separately in
Sections~\ref{sec:case3} and~\ref{sec:case4}, where they arise as limits. They
do not arise in Section~\ref{sec:case2}, since a fixed reference keeps the
beliefs away from the boundary at every round
(Proposition~\ref{prop:case2-confine}). Since
$I-A$ is singular, the first question is whether a solution exists at all.

\begin{theorem}[Solvability, and the solution when it exists]
\label{thm:fredholm}
Let the reference be aggregated, as in cases~(I) and~(III) of
Section~\ref{sec:cases}, and let Assumption~\ref{ass:graph} hold.
\begin{enumerate}[label=(\alph*),leftmargin=*]
\item \eqref{eq:fp-compact} has an interior solution if and only if the
  $\pi$-weighted signal does not depend on the response, that is if and only if
  $\sum_v(\pi_v/\beta_v)R^*_v(y)$ is the same for every $y\in\Y$. The common
  value is then pinned by the normalisers,
  \begin{equation}\label{eq:solvability}
    \sum_{v\in V}\frac{\pi_v}{\beta_v}R^*_v(y)
    =\sum_{v\in V}\pi_v\Phi^*_v\qquad\text{for all }y\in\Y .
  \end{equation}
\item Every solution has the form
  \begin{equation}\label{eq:resolvent-sol}
    \mathbf f^*(y)=D\,B^{-1}\mathbf R^*(y)-D\boldsymbol\Phi^*+c(y)\,\one ,
  \end{equation}
  where $c(y)$ is a scalar, the same for every agent but free to vary with the
  response, and $\mathbf R^*,\boldsymbol\Phi^*$ are those of the solution
  itself. For a fixed right-hand side, two solutions differ only in $c$.
\end{enumerate}
\end{theorem}

The solvability condition \eqref{eq:solvability} requires one weighted sum
to score every response equally. Each agent contributes the signal it perceives,
divided by its own temperature and weighted by its influence.

In the solution \eqref{eq:resolvent-sol} only the first term separates the
agents. The scalar $c(y)$ is common to all of them, and
$-D\boldsymbol\Phi^*$ does not depend on $y$, so it shifts each agent's
log-beliefs by one constant that normalisation removes. What distinguishes
the agents is therefore $D\,B^{-1}\mathbf R^*(y)$: the deviation matrix $D$
of Definition~\ref{def:resolvent} records how much more weight agent $v$'s
repeated averages place on each agent than the long-run share $\pi$, and
applying it to the temperature-scaled signals turns those discrepancies into
differences between the agents' log-beliefs. Since the $c(y)$ are undetermined, solvability leaves a whole family of fixed points.
The two cases meet \eqref{eq:solvability} differently, and this is the first
place the challenger axis makes a difference. With a fixed challenger
$\mathbf R^*$ is data, so the condition
imposes $|\Y|-1$ equalities of rewards that were not chosen to satisfy them. It can hold only in particular cases, and Section~\ref{sec:case3} treats both
the case in which it holds and the case in which it fails. With an
aggregated challenger $\mathbf R^*$ moves with the profile, so the condition becomes an equation the profile must solve, and
Section~\ref{sec:case4} analyses it in that form.

\section{An External Standard: Aggregated Reference, Fixed Challenger}
\label{sec:case3}

We take the four dynamics in turn, beginning with the \emph{peer-regularised} dynamics, case~(I), in which the reference is aggregated and the challenger fixed. Quality is set outside the network, for instance by a fitted reward model, and held fixed during training. The graph supplies the anchor, recomputed each round from the beliefs each agent observes. They are also the case that yields the most, because the iteration stays affine in the log-beliefs. The trajectory can then be written down at finite $t$, and the limit has an explicit description as an aggregation of the agents' rewards.

Take $\mathcal{R}_{v,t}=\bar q_{v,t}$ and $\mathcal{C}_{v,t}=\nu_v$, so that with
$r_v(y)=R_v(y;\nu_v)$ fixed,
\begin{equation}\label{eq:case3}
  q_{v,t+1}(y)\propto\bar q_{v,t}(y)\,e^{r_v(y)/\beta_v},
  \qquad
  \tilde{\mathbf f}_{t+1}(y)=A\tilde{\mathbf f}_t(y)+B^{-1}\mathbf r(y).
\end{equation}
Each step solves
$\arg\max_p\bigl\{\E_pr_v-\beta_v\KL(p\|\bar q_{v,t})\bigr\}$, a local RLHF
problem whose anchor is the neighbourhood consensus rather than the agent's
own prior. Since that anchor moves every round, the sequence of steps
optimises no fixed objective. By Theorem~\ref{thm:fredholm} the fixed-point
equation generically has no interior solution here, the right-hand side being the reward
data and \eqref{eq:solvability} a constraint on it.

\subsection{Exact trajectory}

With a fixed challenger the quality signal never changes, so the log-space
recursion is affine and can be solved outright. The result is an exact
formula for every agent at every round: a term growing linearly at a rate common to
every agent, a constant offset particular to the agent, and a remainder that vanishes at the rate the graph mixes.

\begin{theorem}[Exact trajectory]
\label{thm:case3}
Assume that Assumption~\ref{ass:graph} holds, and for two responses $y,y'$ let
$\phi_{v,t}=\log\bigl(q_{v,t}(y)/q_{v,t}(y')\bigr)$. Then for every agent, every
pair of responses and every finite $t$,
\begin{equation}\label{eq:case3-logodds}
  \phi_{v,t}
  \;=\;\underbrace{t\sum_u\frac{\pi_u}{\beta_u}\bigl(r_u(y)-r_u(y')\bigr)
       +\pi^\top\boldsymbol\phi_0}
  _{\text{the same at every agent; the first part grows with }t}
  \;+\;\underbrace{\Bigl[D\,B^{-1}\bigl(\mathbf r(y)-\mathbf r(y')\bigr)\Bigr]_v}
   _{\text{differs between agents; constant in }t}
  \;+\;\varepsilon_{v,t},
\end{equation}
where $\boldsymbol\phi_0=\bigl(\log(q_{v,0}(y)/q_{v,0}(y'))\bigr)_{v\in V}$
collects the agents' initial log-odds for the same pair, and
$\varepsilon_{v,t}$ decays at the rate $|\lambda_2(A)|^t$ of
Assumption~\ref{ass:graph}, up to a factor polynomial in $t$.
\end{theorem}

Two consequences follow, both concerning how the agents differ from one another. The agents do not become equal. Their log-odds come to differ by the fixed offsets
$[D\,B^{-1}\mathbf r]_v$, which vanish only when $r_v/\beta_v$ is the same at
every agent. And when $\bar r$ has a unique maximiser $y^*$, with margin
$\Delta_{\pi,\beta}$, every agent's belief converges to $\delta_{y^*}$, the
odds against $y^*$ falling as $e^{-t\Delta_{\pi,\beta}}$.

\begin{figure}[t]
\centering
\includegraphics[width=0.62\textwidth]{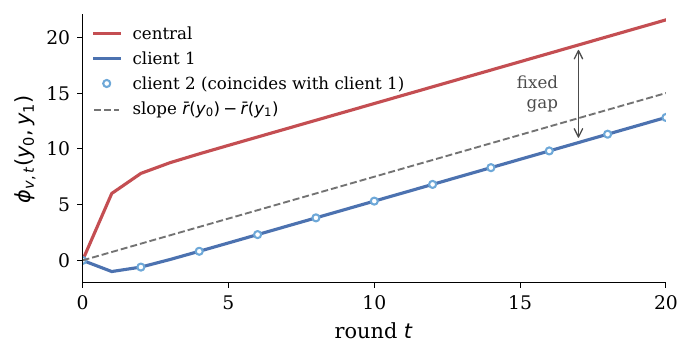}
\caption{The three terms of Theorem~\ref{thm:case3} on the instance of Section~\ref{sec:intro-example}; see Example~\ref{ex:rates}. The two clients' curves coincide, by symmetry, and client~2's is marked by circles.}
\label{fig:rates}
\end{figure}

The rate of the growing term is the influence-weighted reward difference, and the remainder decays at the rate the graph mixes. The remainder has the closed form $\varepsilon_{v,t}=\bigl[(A^t-\one\pi^\top)\bigl(\boldsymbol\phi_0-D\,B^{-1}(\mathbf r(y)-\mathbf r(y'))\bigr)\bigr]_v$, and so is also at most $\tau_1(A)^t$ times the oscillation of $\boldsymbol\phi_0-D\,B^{-1}(\mathbf r(y)-\mathbf r(y'))$, a bound with no constant (Appendix~\ref{app:caseI}). So an agent's position is
decided by where the network as a whole is heading, adjusted by a fixed amount
of its own, with the influence of its starting point dying away.

\begin{example}[The three terms of Theorem~\ref{thm:case3}]
\label{ex:rates}
Run \eqref{eq:case3} on the instance of Section~\ref{sec:intro-example}.
Every agent's log-odds of $y_0$ against $y_1$ grow at the same rate, the
influence-weighted margin $\bar r(y_0)-\bar r(y_1)$, and the three curves of Figure~\ref{fig:rates} become parallel. The two clients' curves coincide, by symmetry, but the central agent's does not. Each agent keeps the offset $[D\,B^{-1}\mathbf r]_v$ the theorem assigns it, and the central agent's exceeds the clients'. The remainder has decayed within a few rounds,
leaving the two terms the theorem separates.
\end{example}

Three consequences of Theorem~\ref{thm:case3} follow. Each is read off \eqref{eq:case3-logodds} directly, and the derivations are
collected after the example that follows them.

\begin{corollary}[Ties]
\label{cor:case3-ties}
If $M=\arg\max_y\bar r(y)$ has more than one element, the limit puts no mass
outside $M$, and on $M$ it satisfies
\[
  \log\frac{q^*_v(y)}{q^*_v(y')}
  =\pi^\top\boldsymbol\phi_0
  +\Bigl[D\,B^{-1}\bigl(\mathbf r(y)-\mathbf r(y')\bigr)\Bigr]_v ,
  \qquad y,y'\in M ,
\]
where $\boldsymbol\phi_0$ is the vector of initial log-odds of $y$
against $y'$, as in Theorem~\ref{thm:case3}. Both terms on the right are
differences of a single function of $y$, so the limit is a Gibbs distribution
on $M$:
\[
q^*_v(y)\;\propto\;\bar q_0(y)\,
  \exp\Bigl(\bigl[D\,B^{-1}\mathbf r(y)\bigr]_v\Bigr),
  \qquad y\in M,
  \qquad\text{where }\;
  \bar q_0(y)\;\propto\;\prod_{u\in V}q_{u,0}(y)^{\pi_u} ,
\]
In this form the limit has the shape of \eqref{eq:gibbs}: the $\pi$-weighted geometric mean $\bar q_0$ of the agents' starting beliefs acts as the reference, and agent $v$'s offset
$[D\,B^{-1}\mathbf r(y)]_v$ as the tilt. The tie is broken by where the network
began, pooled logarithmically with the influence weights, together with each
agent's own offset. \end{corollary}

A tie removes the growing term, since $\bar r$ does not separate $y$ from $y'$
within $M$.

 What remains is where the agents started, carried through the graph as
$\pi^\top\boldsymbol\phi_0$ and common to all of them, plus each agent's own
offset. Ties are therefore one of the few configurations in this case where the
initial beliefs still affect the limit. Whether the agents end up agreeing depends on
the offset, which vanishes exactly when $r_v(y)/\beta_v$ does not depend on $v$.

\begin{remark}[Scope of the theorem]
\label{rem:unbounded}
The recursion is affine in $\mathbf r$ and the theorem uses no bound on it, so
although $r_v=R_v(\cdot;\nu_v)$ takes values in $[0,1]$ here, the statement
holds verbatim for an arbitrary real reward vector at each agent, as produced
by a reward model trained in the usual
way~\citep{ziegler2019,ouyang2022}. Nor does it impose any condition on the temperatures, whereas the convergence
results of Sections~\ref{sec:case2} and~\ref{sec:case4} require the
temperatures to be bounded below. The polynomial factor in
the remainder is absent whenever $A$ is diagonalisable, in particular whenever
it is symmetric.
\end{remark}

\subsection{The rule the network implements}
\label{sec:case3-reading}

\paragraph{The induced rule and its weights.} Assume that $\bar r$ has a
unique maximiser; the tied case is Corollary~\ref{cor:case3-ties}. Every agent then ends at that maximiser. The rule the network implements selects the response maximising
$\sum_v(\pi_v/\beta_v)r_v(y)$: a weighted sum of rewards, hence a scoring
rule.
An agent's weight is its influence $\pi_v$ divided by its caution $\beta_v$. A cautious central agent and an aggressive peripheral one can therefore count equally. The temperature enters inversely because the graph acts on $r_v/\beta_v$. By \eqref{eq:intro-explicit} agent $v$ adds $r_v/\beta_v$ to its log-belief each round, and its observers average log-beliefs. The quantity an agent contributes to those averages is thus its reward divided by its temperature.
Equal attention alone does not equalise the weights. Making $A$ doubly stochastic
gives every agent the same $\pi_v$, and the agents still count unequally
unless their temperatures agree. Because the rule is cardinal, an agent can multiply its own weight by rescaling a reward model only it can see. This matters in practice, since each agent fits its own reward model with its own normalisation. Under the ordinal rule of Section~\ref{sec:case4}, an agent's contribution stays bounded however its rewards are scaled.

The weights also determine what a group of agents can and cannot do to change
the outcome.

\begin{remark}[Edge-dependent temperatures]
\label{rem:edge-temps}
An agent need not weigh its departure from every neighbour at the same rate.
Replacing $\beta_v$ in the penalty of \eqref{eq:pairwise-objective} by a
temperature $\beta_{vu}>0$ for each edge leaves the derivation of
Section~\ref{sec:aggregates} intact, since $\sum_uw_u\KL(p\|q_u)=W\KL(p\,\|\,\tilde q)-W\log Z$ for any positive weights $w_u$, with $W=\sum_uw_u$, $\tilde q=\prod_uq_u^{w_u/W}/Z$ and $Z=\sum_y\prod_uq_u(y)^{w_u/W}$; the last term does not depend on $p$. Taking
$w_u=a_{vu}\beta_{vu}$ returns \eqref{eq:case3} with
\[
  \tilde a_{vu}=\frac{a_{vu}\beta_{vu}}{\sum_u a_{vu}\beta_{vu}},
  \qquad
  \tilde\beta_v=\sum_u a_{vu}\beta_{vu} ,
\]
so every result of this section holds with $A$, $\pi$, $D$ and $\beta_v$
replaced by their tilded counterparts. The matrix $\tilde A$ is again
row-stochastic by construction and has the same support as $A$, so
Assumption~\ref{ass:graph} passes to it. Attention and trust therefore enter
only through the product $a_{vu}\beta_{vu}$, whose direction reweights the
graph and whose magnitude sets the temperature. The reward term involves no temperature, and the substitution gives nothing for case~(II); in case~(III) the
substitution leaves the reference and challenger on different graphs, which
is the situation of the Supplementary Material.
\end{remark}

\paragraph{The collective belief at every round.} Corollary~\ref{cor:case3-ties}
and the two consequences above describe the limit. One consequence of Theorem~\ref{thm:case3} holds at every
$t$, and needs only $\pi^\top A=\pi^\top$. Applying $\pi^\top$ to the log-space update gives
$\pi^\top\tilde{\mathbf f}_t=\pi^\top\mathbf f_0+t\,\bar r$ exactly, so the
$\pi$-weighted log-beliefs accumulate the same increment every round. Write
$p_t$ for the \emph{collective policy}, the $\pi$-weighted geometric mean $p_t(y)\propto\prod_vq_{v,t}(y)^{\pi_v}$, whose log-weights are
$\pi^\top\tilde{\mathbf f}_t$. Then
\begin{equation}\label{eq:case3-finite}
  p_t(y)=p_0(y)\,e^{\,t\,\bar r(y)}\Big/\textstyle\sum_{y'}p_0(y')\,e^{\,t\,\bar r(y')}
  \qquad\text{for every }t\geq0 .
\end{equation}
This is \eqref{eq:gibbs}, the single-agent Gibbs policy, with $p_0$ as the
reference, $\bar r$ in place of $r$, and temperature $1/t$. The collective
policy therefore \emph{is} a Gibbs policy over $\bar r$ at every round, with
the round index entering only as the reciprocal temperature. The ranking of responses is settled immediately, and a longer run only widens the margin.
The statement concerns the differences alone, not any individual agent, whose
offset
$[D\,B^{-1}\mathbf r]_v$ does not decay with $t$.

\paragraph{Keeping part of the anchor.} The aggregated reference of this case is an extreme, in which an agent's own anchor receives no weight. Assume instead that it
anchors to $\mu_v^{1-\theta}\bar q_{v,t}^{\,\theta}$, keeping a fraction
$1-\theta$ of its own. For every $\theta<1$ the dynamics have a unique
interior fixed point, reached geometrically, at which the log-odds of $y$
against $y'$ are $\tfrac{1}{1-\theta}(\bar r(y)-\bar r(y'))$ plus terms that
stay bounded as $\theta\to1$. The
influence weights $\pi_v/\beta_v$ therefore order the responses at every
$\theta$, not only at $\theta=1$. What the fully aggregated reference adds is
that the leading term grows without bound, turning a tilt towards the
$\bar r$-maximiser into concentration on it. Theorem~\ref{thm:case3} is the $\theta\to1$ limit of this family.

\subsection{When some agents are frozen}
\label{sec:case3-partial}

Theorem~\ref{thm:case3} describes a network all of whose members update. Some may not, for example a frozen checkpoint or an annotator pool whose labels do not change. The concentration then disappears and
every updating agent is left with an interior limit. Write $\AUU$ for the block of $A$ among the updating agents and $\AUF$ for
the block from updating to frozen. When every updating agent can reach a frozen one by following attention, through a chain of agents each placing positive weight on the next, $\rho(\AUU)<1$, so
\[
  Z=\sum_{s\geq0}\AUU^{\,s}=(I-\AUU)^{-1},
  \qquad
  H=Z\,\AUF
\]
both exist, $z_{vu}$ counting the expected visits to $u$ by a walk from $v$
that follows attention until it reaches a frozen agent, and $h_{vb}$ giving
the probability that the walk stops at $b$. The rows of $H$ sum to one.

\begin{theorem}[Closed-form limit]
\label{thm:case3-partial}
Assume every updating agent can reach a frozen one. Then for every updating $v$ the beliefs converge to
\begin{equation}\label{eq:case3-partial}
  \log q^*_v(y)=\sum_{b\in\Frz}h_{vb}\log q_{b,0}(y)
  +\sum_{u\in\Alg}\frac{z_{vu}}{\beta_u}\,r_u(y)+\mathrm{const}_v ,
\end{equation}
a distribution putting positive mass on every response, and independent of how
the updating agents were initialised. Convergence is geometric, at the rate $\rho(\AUU)^t$ up to a factor polynomial in $t$.
\end{theorem}

The probability that the walk of Lemma~\ref{lem:substoch} from $v$ has not been absorbed after $t$ steps is $(\AUU^t\one)_v$, which decays at the rate $\rho(\AUU)^t$. The rate is set by how long an agent can avoid the boundary. Exponentiating puts the limit
in the Gibbs form of a single agent,
\[
q^*_v(y)\;\propto\;\underbrace{\prod_{b\in\Frz}q_{b,0}(y)^{h_{vb}}}
  _{\text{inherited reference}}\;
  \exp\Bigl(\underbrace{\textstyle\sum_{u\in\Alg}(z_{vu}/\beta_u)\,r_u(y)}
  _{\text{accumulated reward}}\Bigr),
\]
with the first factor as the reference and the second as the tilt. The reference is inherited from the frozen agents. By Lemma~\ref{lem:substoch} the weights $h_{vb}$ are the probabilities that a
walk following attention from $v$ stops at $b$. Each agent adopts the boundary's positions in proportion to how its paths reach them. The second is
a reward, accumulated along the way. Here $z_{vu}$ counts expected visits to $u$ before absorption. An agent's own tilt and those of the agents it passes through are thus added, each divided by its temperature.

An agent behind a boundary therefore ends where one round of single-agent
alignment at temperature one would have put it, from a reference and a reward
neither of which is its own. The reference is the weighted geometric mean of the
frozen agents' beliefs, with weights $H_{v\cdot}$. The reward is the network's
rewards summed, each divided by its owner's temperature and weighted by how
often the walk from $v$ passes through that owner.

The limit does not depend on the updating agents' initial beliefs, which
\eqref{eq:case3-logodds} shows it would in the absence of frozen agents, and
it depends permanently on the frozen agents' beliefs. The agents reach different limits,
since $Z_{v\cdot}$ and $H_{v\cdot}$ differ across agents. Their disagreement is determined by their positions in the graph relative
to the frozen agents, and not only by their preferences.
The limit is also interior. The solvability obstruction of
Theorem~\ref{thm:fredholm} is absent and attaching a boundary deliberately
restores an interior fixed point that the data generically deny.

\subsection{The rule with parametric policies}
\label{sec:case3-experiment}

This subsection reports an experiment illustrating
Theorem~\ref{thm:case3} on an instance that violates two of its assumptions,
which indicates whether the conclusion extends beyond the setting in which it
is proved. The experiment is not part of the theory, and none of the results
above depends on it.

 Theorem~\ref{thm:case3} assumes that each
agent attains the maximiser of \eqref{eq:general} over all distributions on
$\Y$, and that it does so separately for each prompt. An agent running a
language model satisfies neither. It optimises a parametric policy, which
only approximates the maximiser, and the parameters are shared across
prompts, which couples them. The experiment asks whether the collective
policy still concentrates on the $\bar r$-maximiser under those two
relaxations.

\paragraph{Setup.} The agents are three copies of a small instruction-tuned
language model. The data come from the HelpSteer2
dataset~\citep{wang2024helpsteer2}, which pairs each prompt with two candidate
responses and with human ratings of those responses on several attributes.
Each agent is fine-tuned towards a different attribute, so that $r_v$ is that
agent's attribute score. The three are arranged on a hub-and-spoke graph and run for $25$ rounds. In each round every agent forms the geometric mean of its neighbours' current
policies as its reference, then makes one pass over the prompts, ascending the gradient of $\E_q[r_v]-\beta_v\KL(q\,\|\,\bar q_{v,t})$ with respect to its own parameters. This is a step towards the maximiser that \eqref{eq:case3} assumes
is attained exactly, and the gap between the two is one of the relaxations
being tested. Section~\ref{app:experiment} gives the
model, the fine-tuning, the update and two controls for a possible confound
with response length.

The experiment is run in two configurations. An agent's weight $\pi_v/\beta_v$
combines its attention share with its temperature, and the hub is given a
dominant weight in each of the two ways separately. In the first, attention favours the spokes,
$\pi=(\tfrac14,\tfrac38,\tfrac38)$, and the temperatures are
$\boldsymbol\beta=(\tfrac14,1,1)$, so that the hub optimises four times as
aggressively as the spokes; the weights $\pi_v/\beta_v$ are then
$(1,\tfrac38,\tfrac38)$. In the second, the temperatures are equal,
$\boldsymbol\beta=(1,1,1)$, and the attention is reversed to favour the hub,
$\pi=(0.70,0.15,0.15)$, which are also the weights. The two configurations separate the two sources of an agent's weight. In the
first the hub receives the least attention of the three, yet has the largest
weight, which its temperature supplies. A reader might then attribute
the outcome to the temperatures alone. In the second the temperatures are equal. The weights are then the attention shares, and nothing else distinguishes the agents. The two examples show that either component of $\pi_v/\beta_v$ can decide the outcome on its own.

\begin{figure}[t]
\centering
\includegraphics[width=0.95\textwidth]{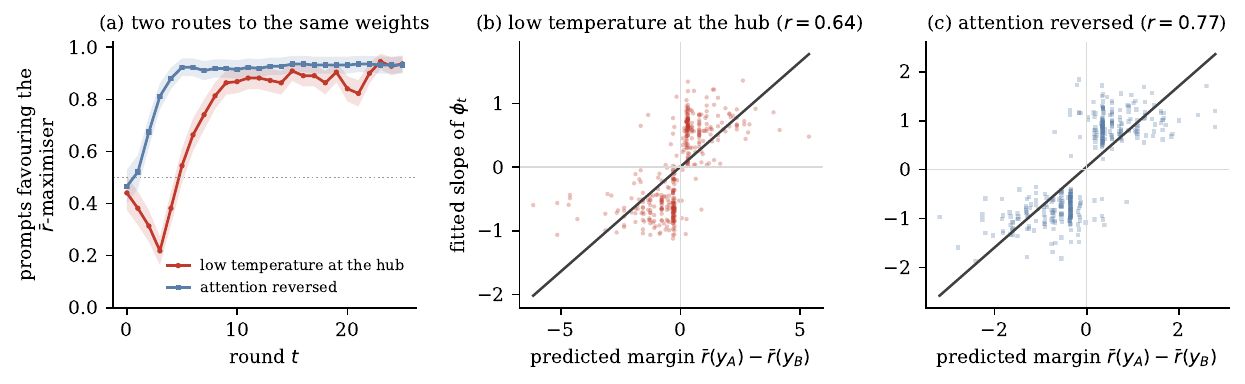}
\caption{Three language models, each aligned towards a different human-rated
attribute, converge on the response Theorem~\ref{thm:case3} predicts
(Section~\ref{sec:case3-experiment}). (a) Share of discriminating prompts on
which the collective policy favours the $\bar r$-maximiser, with $95\%$ bands.
(b), (c) For each prompt of the same two runs, the fitted slope of the
collective log-odds against the predicted margin, least-squares fit in
grey.}
\label{fig:experiment}
\end{figure}

\paragraph{Results.} Theorem~\ref{thm:case3} predicts that the collective
policy concentrates on the maximiser of $\bar r$, and the runs confirm it: after $25$ rounds the collective policy favours that response on
$94\%$ and $93\%$ of the prompts in the two configurations. Panel~(a) of Figure~\ref{fig:experiment} shows the share over rounds. It begins below $\tfrac12$, since the agents start from a
common policy favouring the other response, crosses $\tfrac12$ by round five,
and settles above $0.9$. The figures are over the prompts on which $\bar r$ and the
unweighted sum $\sum_vr_v$ favour different responses, which is where the
weighting makes a difference to the prediction: $220$ of the $400$ prompts in
the first configuration and $234$ in the second.

Theorem~\ref{thm:case3} predicts not only which response the agents end on
but how fast they get there. Each agent's log-odds grow linearly in $t$ at a
rate proportional to the margin $\bar r(y_A)-\bar r(y_B)$, so prompts on
which the weighted rewards differ more should separate faster. Fitting a line in $t$ to each prompt's collective
log-odds accounts for a median of $85\%$ of the variation in both configurations. The growth is close to linear, as predicted, and the fitted slopes correlate with the predicted margins at $0.64$ and $0.77$
(panels~(b) and~(c)). The correlation strengthens with the horizon, from
$0.49$ over four rounds to $0.83$ over twenty-five, as the theorem implies,
since the linear term dominates the transient only as $t$ grows.

Panels~(b) and~(c) show one feature the theorem does not account for. The
fitted slopes cluster according to the sign of the predicted margin rather
than varying continuously with its magnitude. The sign is therefore predicted
better than the rate.
Where that difference comes from is not determined by these runs.

Throughout, an agent's belief is its policy restricted to the two rated responses. The runs track the log-odds between a fixed pair and not the
responses an agent would generate on its own.

\subsection{Further results}
\label{sec:case3-further}

Two sections of the Supplementary Material contain further results for this
case. Section~\ref{supp:shared} anchors each agent to the weighted
geometric mean $\mu_v^{1-\theta}\bar q_{v,t}^{\,\theta}$ of its own reference and its
neighbours', interpolating between the fixed reference at $\theta=0$ and the
aggregated one at $\theta=1$. For every $\theta<1$ the matrix $I-\theta A$ is
invertible, so the fixed-point equation is solvable for any rewards, with a
unique fixed point reached from every start. The solvability condition of
Theorem~\ref{thm:fredholm} is therefore a feature of the fully aggregated
reference alone, and not of aggregation as such.

Section~\ref{supp:case3} contains four further results. The first two are the
proofs of the consequences stated after Theorem~\ref{thm:case3}. The third
shows that Theorems~\ref{thm:case3} and~\ref{thm:fredholm} are consistent:
the latter denies an interior fixed point when $\bar r$ is not constant, and
the limit the agents reach lies on the boundary of the simplex. The fourth
accounts for the difference between Theorems~\ref{thm:case3}
and~\ref{thm:case3-partial}, which reach opposite conclusions about the same dynamics. The whole difference is whether the updating agents' weights sum to
one, as they do when every agent updates and do not when any is frozen.
Three examples illustrate the theorems: two show what the graph and the temperatures each contribute to an agent's weight in Theorem~\ref{thm:case3}, and one gives the limits of Theorem~\ref{thm:case3-partial} on a line with both ends frozen. Two more treat ties in $\bar r$: the first shows a tie broken agent by agent, in opposite directions for the central agent and the clients, and the second separates the part of the outcome decided by the starting beliefs from the part decided by the rewards.

\section{Fixed Anchors: Fixed Reference, Aggregated Challenger}
\label{sec:case2}

Exchanging the two roles of Section~\ref{sec:case3} gives the \emph{anchored}
dynamics, case (II) of Section~\ref{sec:fourcases}: each agent keeps its own
reference fixed and takes its standard of quality from the agents it observes.
Whereas in Section~\ref{sec:case3} every agent was driven onto the maximisers
of $\bar r$, so that the agents agreed in the limit, a fixed
anchor leaves each agent with its own $\mu_v$ and the agents need not agree
at all.

Take $\mathcal{R}_{v,t}=\mu_v$ and
$\mathcal{C}_{v,t}=\check q_{v,t}$:
\begin{equation}\label{eq:case2}
  q_{v,t+1}(y)\propto\mu_v(y)
  \exp\Bigl(\tfrac{1}{\beta_v}\sum_ua_{vu}\sum_{y'}q_{u,t}(y')P_v(y,y')\Bigr).
\end{equation}

\subsection{Fixed points}
\label{sec:case2-fixedpoints}

The first questions are whether a fixed point exists, whether it is unique,
and whether at a fixed point the agents hold the same belief, and the three come apart. Existence holds at every temperature, with no condition on
the graph or the preferences.
Uniqueness holds under a condition bounding
how strongly the agents respond to one another. Agreement is characterised
exactly: the agents hold a common belief at a fixed point precisely when a
single distribution is simultaneously each agent's best response to it.

\begin{proposition}[Existence, uniqueness, consensus]
\label{prop:case2-fp} Consider the dynamics \eqref{eq:case2}, on any attention matrix $A$.
\begin{enumerate}[label=(\alph*),leftmargin=*]
  \item A fixed point exists for every vector of temperatures
        $(\beta_v)_{v\in V}$ with $\beta_v>0$.
  \item If $\rho(B^{-1}A)<2$, the fixed point is unique. Since $A$ is stochastic,
        $\rho(B^{-1}A)\leq1/\beta_{\min}$ with
        $\beta_{\min}=\min_v\beta_v$, so $\beta_{\min}>\tfrac12$ suffices.
  \item An interior consensus fixed point $q^*_v\equiv q^*$ exists if and only if
        \[
          q^*(y)\;\propto\;\mu_v(y)\,
          \exp\Bigl(\tfrac{1}{\beta_v}R_v\bigl(y;q^*\bigr)\Bigr)
          \qquad\text{for every }v ,
        \]
        that is, if and only if $q^*$ is the Gibbs policy \eqref{eq:gibbs} of every
        agent against its own anchor and the common challenger $q^*$.
        Equivalently, in log-odds,
        \[
          \log\frac{q^*(y)}{q^*(y')}
          =\log\frac{\mu_v(y)}{\mu_v(y')}
          +\frac{R_v(y;q^*)-R_v(y';q^*)}{\beta_v}
        \]
        at every agent, for every pair $y,y'$. In particular the right-hand
        side takes the same value at every agent; that agreement is
        necessary but not sufficient, since with a single agent it holds for
        every $q^*$. The attention weights do not appear, each agent's
        challenger at a consensus profile being $q^*$ itself.
  \item If the agents share a kernel, $P_v\equiv P$, and $A$ is reversible, that is $\omega_va_{vu}=\omega_ua_{uv}$ for every pair $u,v$ and some positive weights $(\omega_v)_{v\in V}$ (for instance $A$ symmetric, with $\omega_v\equiv1$), then the fixed point is unique for every vector of temperatures, so the threshold in part~(b) is not needed.
\end{enumerate}
\end{proposition}

Part~(a) follows from Brouwer's fixed-point theorem~\citep[Ch.~6]{border1985}, \eqref{eq:case2} being a continuous
self-map of the compact convex set $\Delta(\Y)^n$.

Part~(b) is a contraction condition: a change of $\varepsilon$ in the beliefs
of $v$'s neighbours moves $v$'s own belief by at most
$\varepsilon/(2\beta_v)$, by Lemmas~\ref{lem:signal-lip}
and~\ref{lem:softmax}, and $\rho(B^{-1}A)<2$ means that composing these factors around the network shrinks any difference between two profiles. The simpler
$\beta_{\min}>\tfrac12$ makes every factor smaller than one individually.
Neither condition involves the kernels, because the bound
$\varepsilon/(2\beta_v)$ on how far $v$ moves holds for every preference
kernel. Both are sufficient and neither is necessary: the Supplementary
Material gives an instance below the threshold whose fixed point is still
unique. Below the threshold there may be several: the Supplementary Material gives two
agents with different kernels and three fixed points.

Part~(c) is an equivalence, and it shows that consensus is restrictive. With
$g_v=\log\mu_v+R_v(\cdot\,;q^*)/\beta_v$ it requires $g_v-g_u$ to be constant
in $y$ for every pair of agents, which is $(n-1)(|\Y|-1)$ equations on the
$|\Y|-1$ free parameters of $q^*$, and more equations than unknowns as soon as
two agents differ. At a generic fixed point the agents therefore hold
different beliefs. The condition constrains anchors, kernels and temperatures together. Agents with different anchors can still agree if their signals
compensate, whereas agents sharing a kernel and an anchor but not a
temperature cannot, each being tilted from the common anchor in proportion to
its own $1/\beta_v$.

Part~(d) removes the threshold of part~(b) when the agents share a kernel and the attention is reversible. Their temperatures and anchors may differ. Summing $\omega_va_{vu}=\omega_ua_{uv}$ over $v$ gives $\sum_v\omega_va_{vu}=\omega_u$, so the normalised weights are a stationary distribution of $A$, and reversibility is the detailed-balance condition for the chain on the agents with transition matrix $A$. Symmetric attention is the case of uniform weights. Reversibility alone is not enough: the Supplementary Material gives two agents on a symmetric graph with different kernels and several fixed points.

Whether the dynamics reach a fixed point is a separate question, taken up in
Section~\ref{sec:case2-convergence}.

\paragraph{Weighted means of the beliefs.}
\label{sec:case2-lottery}
Proposition~\ref{prop:case2-fp} describes no collective object, and with a
fixed anchor there is no obvious candidate, since each agent is a tilt of its own $\mu_v$ and of nothing shared. There is nonetheless something to say. Any weighted mean of the agents' beliefs is an approximate
maximal lottery of the correspondingly weighted kernel, in the sense of
Definition~\ref{def:maxlot}. Unlike the results for the aggregated-reference cases, it holds for every weighting and singles none out.

\begin{theorem}[Weighted means are approximate maximal lotteries]
\label{thm:case2-lottery} Let $\mathbf q^*$ be any fixed point of \eqref{eq:case2}, on any attention matrix $A$, and let $w$ be any probability vector on the agents. Write
\[
  \bar P_w=\sum_vw_vP_v,
  \qquad
  \bar q_w=\sum_vw_vq^*_v,
  \qquad
  M_v=\log\frac{1}{\min_y\mu_v(y)},
  \qquad
  \mathcal D_w=\max_v\bigl\|q^*_v-\bar q_w\bigr\|_1
\]
for the $w$-weighted kernel and policy, for the \emph{anchor spread} of
agent $v$, and for the disagreement about $\bar q_w$. Each $M_v$ is finite because
the anchors are interior by assumption, and
$M_v=\log|\Y|$ exactly when $\mu_v$ is uniform, its smallest possible
value. Then for every $q\in\Delta(\Y)$,
\begin{equation}\label{eq:case2-lottery}
  \E_{y\sim\bar q_w,\;y'\sim q}\bigl[\bar P_w(y,y')\bigr]\;\geq\;\tfrac12-\varepsilon_w ,
  \qquad
  \varepsilon_w\;\leq\;\sum_vw_v\beta_vM_v+\tfrac32\,\mathcal D_w .
\end{equation}
That is, $\bar q_w$ is an $\varepsilon_w$-maximal lottery of $\bar P_w$ in the sense of
Definition~\ref{def:maxlot}.
\end{theorem}

Two weightings are of interest. Taking $w=\pi$, a stationary distribution of $A$ (the unique one under Assumption~\ref{ass:graph}), gives the attention-weighted policy $q_\pi=\sum_v\pi_vq^*_v$ and the kernel $P_\pi=\sum_v\pi_vP_v$, with slack
$\sum_v\pi_v\beta_vM_v+\tfrac32\mathcal D_\pi$. Taking $w_v=(\pi_v/\beta_v)/\eta$, with
$\eta=\sum_v\pi_v/\beta_v$, gives the \emph{influence-weighted} kernel
$\bar P=\eta^{-1}\sum_v(\pi_v/\beta_v)P_v$ and the matching policy
$\bar q=\eta^{-1}\sum_v(\pi_v/\beta_v)q^*_v$, which Section~\ref{sec:case4} uses throughout. The slack is then
$\eta^{-1}\sum_v\pi_vM_v+\tfrac32\mathcal D_{\bar q}$. The
two coincide when the temperatures agree, and neither is singled out by the
dynamics: the bound holds for every $w$ alike. If the agents share a kernel, an anchor and a temperature
then $\mathcal D_w=0$ for both and $\varepsilon_w\leq\beta M$, so the
collective policy approaches a maximal lottery of that kernel as the
temperature falls.

\noindent
The two terms bounding $\varepsilon_w$ measure different things. The
anchor term $\sum_vw_v\beta_vM_v$ falls as the agents optimise more
aggressively, and is smallest when the anchors are uniform, since $M_v$ is
the largest penalty $\KL(p\,\|\,\mu_v)$ can impose, which is large when
$\mu_v$ places little mass on some response. The disagreement term $\tfrac32\mathcal D_w$ measures how far apart the
agents' beliefs are, and vanishes only when they agree. It cannot be
replaced by a measure of how far apart the \emph{kernels} are, because agents whose kernels are close can still hold beliefs that are not, at a fixed
temperature.

Theorem~\ref{thm:case2-lottery} is a statement about the fixed points of
\eqref{eq:case2} and not about its trajectories. Whether the dynamics reach a
fixed point, and what the theorem gives when they do not, are taken up in
Section~\ref{sec:case2-convergence}.

\paragraph{Homogeneous agents.}
\label{sec:case2-homog}

Consider now agents that share a kernel, an anchor and a temperature, the
case the multiplayer literature analyses and the regime in which the agents
may be run as a distributed computation of a single policy, each holding only
its own data. Sharing only two of the three is generically not enough for the agents to
agree, since by Proposition~\ref{prop:case2-fp}(c) consensus requires
$\log\mu_v+R_v(\cdot\,;q^*)/\beta_v$ to agree across agents up to a constant.
With a common kernel and anchor, agents with different temperatures are tilted
from the anchor by different multiples of the same signal; with a common kernel
and temperature, different anchors enter the condition directly; and with a
common anchor and temperature, different kernels generically give different
signals at $q^*$. The exception is a \emph{degenerate} kernel, one whose
quality signal ranks the responses the same way whatever the opposition,
which removes the only nonlinearity in the dynamics. The Supplementary Material characterises these, and every kernel on two responses is one.

\begin{corollary}[Homogeneous agents]
\label{cor:case2-homog}
Assume that $P_v\equiv P$, $\mu_v\equiv\mu$ and $\beta_v\equiv\beta$.
\begin{enumerate}[label=(\alph*),leftmargin=*]
  \item For every $\beta>0$ there is exactly one consensus fixed point, and it
        is the single-agent regularised Nash policy \eqref{eq:nash} of $P$: the
        distribution each agent would reach in isolation. The graph plays no
        part in it.
  \item That policy is an $\varepsilon$-maximal lottery of $P$ with
        $\varepsilon=\beta\log(1/\mu_{\min})$ and
        $\mu_{\min}=\min_y\mu(y)$.
\item If in addition $\beta>\tfrac12$, it is the \emph{only} fixed point,
        consensus or otherwise, and by Proposition~\ref{prop:case2-geom} every
        trajectory converges to it geometrically at rate $1/(2\beta)$.
\end{enumerate}
\end{corollary}

By part~(a), the graph drops out entirely. The agents agree on the policy
each would have reached alone, whatever the attention weights. Part~(b)
measures how far that policy is from a maximal lottery, the $\varepsilon$
being the largest value the penalty $\beta\KL(\cdot\,\|\,\mu)$ can take. Since
a maximal lottery beats every alternative with probability at least
$\tfrac12$, an $\varepsilon$-maximal lottery with $\varepsilon\geq\tfrac12$
constrains nothing, and the bound is informative only while
$\beta\log(1/\mu_{\min})<\tfrac12$.

\subsection{Convergence and confinement}
\label{sec:case2-convergence}
\label{sec:case2-conv}

Section~\ref{sec:case2-fixedpoints} identifies the fixed points. This one asks whether the dynamics arrive at one, and where the beliefs lie in the
meantime. The condition that made the fixed point unique also makes the iteration a contraction, and it answers the first question as well.

\begin{proposition}[Geometric convergence]
\label{prop:case2-geom}
Assume that $\rho(B^{-1}A)<2$, the condition of
Proposition~\ref{prop:case2-fp}(b). Let
$\mathbf z=(I-\tfrac12B^{-1}A)^{-1}\one$, whose entries are finite and satisfy
$z_v\geq1$, and let
$d_{\mathbf z}(\mathbf q,\mathbf q')=\max_v\|q_v-q'_v\|_1/z_v$. Then the map
$T$ defined by \eqref{eq:case2} is a contraction for $d_{\mathbf z}$ with
modulus
\[
  \gamma=1-\frac{1}{\|\mathbf z\|_\infty}<1 ,
\]
so the iteration converges to the unique fixed point from any
initialisation, at the rate
\[
  d_{\mathbf z}(\mathbf q_t,\mathbf q^*)\;\leq\;\gamma^t\, d_{\mathbf z}(\mathbf q_0,\mathbf q^*) .
\] Under the stronger hypothesis $\beta_v>\tfrac12$ at every $v$, one may take
$\mathbf z=\one$, which recovers the unweighted metric
$\max_v\|q_v-q'_v\|_1$ with modulus $1/(2\beta_{\min})$.
\end{proposition}

The same condition that gives uniqueness therefore also gives convergence, at
a geometric rate whose modulus depends on the graph through $\mathbf z$. Weighting the metric by agent lets the proof work under
$\rho(B^{-1}A)<2$ rather than only under the stronger $\beta_{\min}>\tfrac12$,
for which the unweighted metric suffices.

The next result shows that the beliefs never leave the interior of the
simplex, at any temperature.

\begin{proposition}[The beliefs never leave the interior]
\label{prop:case2-confine}
For the dynamics \eqref{eq:case2}, every vector of temperatures with
$\beta_v>0$, every round $t\geq1$, every agent $v$ and every response $y$
satisfy
\begin{equation}\label{eq:case2-confine}
  q_{v,t}(y)\;\geq\;\mu_v(y)\,e^{-1/\beta_v} .
\end{equation}
\end{proposition}

The iterates therefore stay in a compact subset of the interior, fixed in
advance by the anchors and the temperatures. No agent approaches a point mass, however aggressively it optimises, at any
temperature and from any start. This contrasts with
Section~\ref{sec:case3}, where the beliefs leave every compact subset of the
interior, converging to a point mass when $\bar r$ has a unique maximiser and
to a distribution on the tied set otherwise
(Theorem~\ref{thm:case3} and Corollary~\ref{cor:case3-ties}). There the
reference is rebuilt each round and imposes no lower bound on the beliefs,
whereas here each agent's own anchor supplies one.

Theorem~\ref{thm:case2-lottery} bounds the weighted mean of the beliefs at a
fixed point, so it describes a trajectory only where that fixed point
attracts. Above the threshold of Proposition~\ref{prop:case2-geom} the fixed
point is unique and globally attracting, so the bound applies to the limit of
every trajectory, but it is weak there, $\varepsilon_w$ being of order
$\beta_{\min}>\tfrac12$. Reducing the temperatures tightens the bound and
withdraws the guarantee of convergence. Which of the two happens depends on
the game the agents play in the limit $\beta_v\to0$, in which each maximises
its quality signal outright. If that game has a strict pure equilibrium,
every agent's belief converges to a point mass and the fixed point is locally
asymptotically stable at every low enough temperature, by
Proposition~\ref{prop:local-strict}: the regime in which the bound is sharp
is then one in which it is attained. If the game has no such equilibrium the
beliefs remain in the interior and the fixed point can be unstable, as for
the cyclic instance of Proposition~\ref{prop:case2-below}. The point masses may differ across agents. A stable fixed point is therefore compatible with a mixed $\bar q_w$, because stability constrains the individual beliefs, whereas
the theorem describes their weighted mean.

Where the fixed point is unstable, the paper gives no guarantee for the time
average of the trajectory. The argument that supplies one in
Section~\ref{sec:cesaro} rests on the log-space recursion
\eqref{eq:logspace-vec} carrying $\tilde{\mathbf f}_t$ forward, so that the
signals accumulate; with a fixed reference it does not, since
$\tilde{\mathbf f}_{t+1}=\log\boldsymbol\mu+B^{-1}\mathbf R_t$.

\subsection{Further results}
\label{sec:case2-further}

Neither Proposition~\ref{prop:case2-fp} nor Proposition~\ref{prop:case2-geom}
gives the fixed point explicitly, and for general kernels it has no closed
form, the challenger being built from the profile being solved for. There are
two exceptions. For a degenerate kernel the quality signal does not depend on
the opposition, so each agent settles after one round at
$q^*_v\propto\mu_ve^{c_v/\beta_v}$, with $c_v$ the kernel's score vector, independently of the graph. This covers every problem with two responses. For cautious agents, $\beta_v>\tfrac12$ at every $v$, the
Supplementary Material expands the fixed point about the anchors as
$q^{(1)}_v(y)\propto\mu_v(y)\exp\bigl(\tfrac{1}{\beta_v}\sum_ua_{vu}R_v(y;\mu_u)\bigr)$,
to within $O(1/\beta_{\min}^2)$, so that each agent's anchor is tilted by the
anchors of the agents it observes, weighted by its own row of $A$. Neither
expression contains $\pi$ or any other quantity defined on the network as a
whole, which is the sense in which a fixed reference yields no aggregation
across the network. The remaining case, a non-degenerate kernel below the
threshold, has no closed form. The Supplementary Material records what is known there, including the cycling instance at $n=1$.

\section{Both Roles Aggregated: Aggregated Reference and Challenger}
\label{sec:case4}

This section treats case~(III) of Section~\ref{sec:fourcases}, in which the
distribution an agent is penalised for leaving and the one its responses are
compared against are both averages over the agents it observes. We call
these the \emph{fully endogenous} dynamics:
\begin{equation}\label{eq:case4}
  q_{v,t+1}(y)\propto\bar q_{v,t}(y)\,
  \exp\Bigl(\tfrac{1}{\beta_v}\E_{y'\sim\check q_{v,t}}[P_v(y,y')]\Bigr).
\end{equation}
In log space this is the recursion
$\tilde{\mathbf f}_{t+1}=A\tilde{\mathbf f}_t+B^{-1}\mathbf R_t$ of
\eqref{eq:logspace-vec}, with $\mathbf R_t$ now evaluated at the challengers
$\check q_{v,t}$ and so depending on $\tilde{\mathbf f}_t$ itself. No other case has that dependence.

One object recurs throughout, and we write
\begin{equation}\label{eq:pbar}
  \bar P=\frac1\eta\sum_v\frac{\pi_v}{\beta_v}P_v
\end{equation}
for the \emph{influence-weighted kernel}, with
$\eta=\sum_v\pi_v/\beta_v$ the constant that keeps $\bar P$ a
preference kernel. Each agent is weighted by its influence over its caution, which makes $\bar P$ the ordinal counterpart of the weighted reward
$\bar r$ of Section~\ref{sec:case3}, with the same weights.

Every atom is a fixed point of these dynamics. If every agent holds
$\delta_y$, every reference and every challenger is $\delta_y$, and reweighting
a point mass leaves it unchanged. This holds for every graph, kernel and temperature, so the update has at least $|\Y|$ fixed points and is never a contraction; no single profile attracts every initial condition, since each atom is its own constant trajectory. The result of Section~\ref{sec:case2-convergence}, convergence to one fixed point from every
initialisation, has no counterpart here. In case~(II) the fixed anchors keep
every belief off the boundary and so exclude the atoms. The multiplicity of
fixed points does not by itself exclude every trajectory converging to
\emph{some} fixed point, but no such result is known. Where the question has been settled, the answer is negative: with $A=I$, outside Assumption~\ref{ass:graph}, each agent plays multiplicative weights against itself, and its divergence from an interior equilibrium is non-decreasing
\citep{bailey2018}. The analysis below therefore does not assume
that the profile converges.

What can be analysed is how the averaging acts on two parts of the
log-beliefs. It contracts the differences between the agents. Their disagreement can then be bounded without the profile converging, which
Section~\ref{sec:case4-conv} does. And it leaves their $\pi$-weighted average
$\pi^\top\tilde{\mathbf f}_t$, a vector with one entry per response,
unchanged. Since $\pi$ is the stationary distribution of $A$,
$\pi^\top A=\pi^\top$; in coordinates,
$\sum_v\pi_v(A\tilde{\mathbf f})_v=\sum_u\bigl(\sum_v\pi_va_{vu}\bigr)\tilde f_u
=\sum_u\pi_u\tilde f_u$, so agent $u$'s log-belief enters the averaged profile
with the same total weight $\pi_u$ that it had before. Applying $\pi^\top$ to
the recursion therefore gives
\begin{equation}\label{eq:case4-common}
  \pi^\top\tilde{\mathbf f}_{t+1}=\pi^\top\tilde{\mathbf f}_t
  +\pi^\top B^{-1}\mathbf R_t ,
\end{equation}
so the $\pi$-weighted average changes only through the signal. In terms of the
collective policy $p_t(y)\propto\prod_uq_{u,t}(y)^{\pi_u}$, which is
proportional to $\exp\bigl((\pi^\top\tilde{\mathbf f}_t)(y)\bigr)$,
\eqref{eq:case4-common} reads
\[
  p_{t+1}(y)\;\propto\;p_t(y)\,\exp\bigl((\pi^\top B^{-1}\mathbf R_t)(y)\bigr),
\]
an exponential-weights update. Exponential weights come with a guarantee for their time average whether or not the iterates converge, which Section~\ref{sec:cesaro} establishes. This is how the network aggregates
without any assumption that the profile converges. Section~\ref{sec:case4-exist} first characterises the fixed
points. Interior ones can exist, since the signals move with the profile and
the condition that Theorem~\ref{thm:fredholm} places on them can be met at
some profiles, but no condition is known under which they attract.
Section~\ref{sec:case4-further} collects further results, among them
conditions under which an atom attracts.

\subsection{Fixed points: existence, consensus, and structure}
\label{sec:case4-exist}

A fixed point always exists, and so does a consensus one; the agents agree on
an \emph{interior} profile only under a condition that generically fails.

\begin{proposition}[Existence, atoms, and the consensus condition]
\label{prop:case4-fp}
Assume that Assumption~\ref{ass:graph} holds, and consider the dynamics
\eqref{eq:case4}.
\begin{enumerate}[label=(\alph*),leftmargin=*]

\item For any $\beta_v>0$ a fixed point exists in the closed simplex.
\item For each response $y$, the profile at which every agent holds
      $\delta_y$ is a fixed point, whatever the kernels, attention weights and
      temperatures.
\item A consensus profile $q^*_v\equiv q^*$, with support
      $S=\operatorname{supp}q^*$, is a fixed point if and only if
      \begin{equation}\label{eq:unanimity}
        R_v(y;q^*)=R_v(y';q^*)\qquad\text{for every }v\text{ and all }y,y'\in S .
      \end{equation}
      Each agent's own signal must be constant on $S$, separately for each $v$;
      off $S$ no condition is imposed. Part~(b) is the case $|S|=1$: the only pair available is $y=y'$, so
      \eqref{eq:unanimity} holds automatically and every atom is a fixed point.
\end{enumerate}
\end{proposition}

Part~(a) follows from
part~(b), which is immediate: a geometric mean of copies of $\delta_y$ is
$\delta_y$, and a point mass is unchanged by reweighting, so each of the
$|\Y|$ atoms is a fixed point whatever the kernels, attention weights and
temperatures. The substantive part is~(c), whose condition \eqref{eq:unanimity} requires that at $q^*$ every agent's quality signal scores the responses in $S$ equally,
and for $S=\Y$ that value must be $\tfrac12$, since averaging the signal
against $q^*$ itself gives $\tfrac12$. Then $R_v(y;q^*)=\tfrac12$ for every
$y$, so every response, and hence every mixture $p$, beats $q^*$ under $P_v$
with probability exactly $\tfrac12$; since $P_v+P_v^\top=\one\one^\top$,
$q^*$ in turn beats every $p$ with probability $\tfrac12$. By
Definition~\ref{def:maxlot}, $q^*$ is therefore a maximal lottery of every
$P_v$, and by linearity of any weighted average of them, in particular of
$\bar P$. The restriction to $S=\Y$ matters. On a smaller support
\eqref{eq:unanimity} constrains the signal only on $S$, and a response outside
$S$ may beat the ones on it, so $q^*$ need not be a maximal lottery. Weighting by
$\pi_v/\beta_v$ and summing gives the strictly weaker requirement that $q^*$ be
a Nash equilibrium of $\bar P$ alone; weighting by $\pi_v$ would target $\sum_v\pi_vP_v$, the same kernel only when
the temperatures agree.

Interior consensus is nonetheless rare, since requiring one agent's quality
signal to take the same value at every response is $|\Y|-1$ equations, so over all $n$ agents \eqref{eq:unanimity} is
$n(|\Y|-1)$ equations in the $|\Y|-1$ free parameters of $q^*$, which for
$n\geq2$ has no solution unless the agents' equations happen to coincide.
Under Bradley--Terry preferences the condition reduces to agreement on the
best response: if each $r_v$ has a unique maximiser $y^*_v$, then
\eqref{eq:unanimity}
demands $y^*_1=\dots=y^*_n$, so every agent must already prefer the same
response. No weighting across agents can relax this, since
\eqref{eq:unanimity} constrains each agent's own signal.

One case is worth separating, that in which $\bar P$ has a strict Condorcet
winner $y^*$:
its only maximal lottery is $\delta_{y^*}$, so no interior consensus fixed
point exists. This leaves open fixed points at which the agents hold different policies, since those
face different challengers and part~(c) does not apply to them.

\label{sec:spread}

When the agents cannot agree, a fixed point still assigns them different
beliefs, and the next proposition describes how different.

\begin{proposition}[Structure of a fixed point]
\label{prop:case4-spread}
Assume that Assumption~\ref{ass:graph} holds, so that $\pi$ and the deviation matrix $D$
of Definition~\ref{def:resolvent} are defined, and let $\mathbf q^*$ be any
fixed point of \eqref{eq:case4}. Write $\check q^*_v=\sum_ua_{vu}q^*_u$ for the
challenger agent $v$ faces and $R^*_v(y)=R_v(y;\check q^*_v)$ for the quality
signal it receives. Then:
\begin{enumerate}[label=(\alph*),leftmargin=*]
\item \emph{(Common support.)} All agents share one support,
      $S=\operatorname{supp}q^*_v$, the same for every $v$.
\item \emph{(Flatness of the weighted signal on $S$.)} There is a constant $c$
      with
      \begin{equation}\label{eq:case4-solv}
        \sum_v\frac{\pi_v}{\beta_v}R^*_v(y)\;=\;c
        \qquad\text{for every }y\in S ,
      \end{equation}
      and no condition is imposed at $y\notin S$: the signal there may take any
      value. If \eqref{eq:case4-solv} fails on $\Y$ at every interior profile,
      no interior fixed point exists.
\item \emph{(Spread.)} Writing $\mathbf f^*=\log\mathbf q^*$, for every $v$ and
      every $y\in S$
      \begin{equation}\label{eq:case4-spread}
        f^*_v(y)-\sum_u\pi_uf^*_u(y)
        =\bigl[D\,B^{-1}\mathbf R^*(y)\bigr]_v+k_v ,
      \end{equation}
      with $k_v$ independent of $y$. Equivalently, writing
      $p^*(y)\propto\prod_uq^*_u(y)^{\pi_u}$ for the collective policy,
      \begin{equation}\label{eq:case4-spread-gibbs}
        q^*_v(y)\;\propto\;p^*(y)\,
        \exp\Bigl(\bigl[D\,B^{-1}\mathbf R^*(y)\bigr]_v\Bigr) ,
      \end{equation}
      which holds at every $y\in\Y$, both sides vanishing off $S$.
\end{enumerate}
\end{proposition}

Part~(a) holds because zeros propagate. The reference
$\bar q_{v,t}(y)\propto\prod_uq_{u,t}(y)^{a_{vu}}$ vanishes whenever one
observed agent assigns $y$ no mass, so a zero is inherited along every edge
and, under Assumption~\ref{ass:graph}, reaches every agent.

Part~(b) constrains only the weighted sum of the signals.
This is the difference from Proposition~\ref{prop:case4-fp}(c). Consensus requires every agent's signal to be flat on $S$, whereas a fixed point at
which the agents hold different beliefs requires only the sum
$\sum_v(\pi_v/\beta_v)R^*_v$ to be. When even that fails at every interior
profile, no interior fixed point exists, and what the dynamics do instead is
taken up in Section~\ref{sec:cesaro}.

Part~(c) expresses each agent's belief at a fixed point as the collective
policy $p^*$, the geometric mean $\prod_uq^*_u(y)^{\pi_u}$ that
Section~\ref{sec:cesaro} also uses, multiplied by an agent-specific tilt $\exp([D\,B^{-1}\mathbf R^*(y)]_v)$. Writing
$c$ for the common value in \eqref{eq:case4-solv}, the exponents
$h_v(y)=[D\,B^{-1}\mathbf R^*(y)]_v$ are the solution, for each $y\in S$, of
\[
  h_v(y)\;=\;\sum_ua_{vu}h_u(y)+\frac{R^*_v(y)}{\beta_v}-c ,
  \qquad \sum_v\pi_vh_v(y)=0 ,
\]
so an agent's tilt is the average tilt of the agents it observes plus the
amount by which its own scaled signal exceeds the network-wide level $c$. The condition \eqref{eq:case4-solv} makes that level the same at every $y\in S$. The exponent is $\sum_ud_{vu}R^*_u(y)/\beta_u$, which, since $R^*_u$ takes values in
$[0,1]$, is at most $\|D\|_\infty/\beta_{\min}$ in absolute value, so the tilts
vanish as the temperatures rise. Individual beliefs are therefore close to the
collective policy when the temperatures are high. The graph enters through
$\|D\|_\infty$, which the ergodic coefficient of Definition~\ref{def:resolvent}
controls. Row $v$ of $A^t-\one\pi^\top$ is $(e_v-\pi)^\top A^t$, because
$\pi^\top A^t=\pi^\top$, and a vector $x$ summing to zero satisfies
$\|x^\top A\|_1\leq\tau_1(A)\,\|x\|_1$ by Lemma~\ref{lem:dobrushin}. Since
$\|e_v-\pi\|_1\leq2$, the row has $\ell_1$-norm at most $2\,\tau_1(A)^t$, and
summing over $t\geq0$ gives
\[
  \|D\|_\infty\;\leq\;\frac{2}{1-\tau_1(A)} .
\]
The $t=0$ term alone contributes up to $2$, so $\|D\|_\infty$ cannot be made
small by the graph: on the complete graph with uniform weights
$D=I-\tfrac1n\one\one^\top$ and $\|D\|_\infty=2(n-1)/n$, each agent being
tilted by its own signal relative to the average. Section~\ref{app:tau1} of
the Supplementary Material gives $\|D\|_\infty$ for the other example graphs.

\subsection{Contraction of the disagreement}
\label{sec:case4-conv}
The agents do not in general reach consensus, a fixed point assigning them
different beliefs, but the averaging contracts the differences between them,
down to a level set by how far their kernels and temperatures are from
agreeing. The theorem below measures those differences by $\mathcal D_t$, the
largest gap between two agents' unnormalised log-beliefs about any one
response. It controls how far apart their beliefs are: for every pair $u,v$,
$\|q_{u,t}-q_{v,t}\|_1\leq\mathcal D_t$, by Lemma~\ref{lem:softmax} and
inequality~\eqref{eq:D-controls} of Appendix~\ref{app:caseIII}, which also
explains why the unnormalised log-beliefs are used. In each round the averaging multiplies
$\mathcal D_t$ by at most $\tau_1(A)$, by Lemma~\ref{lem:dobrushin}, while the
agents' differing signals add to it.

\begin{theorem}[Contraction of the disagreement]
\label{thm:case4-directed}
Consider the dynamics \eqref{eq:case4}, on any attention matrix $A$. Write $\mathcal D_t=\max_y\osc_v\bigl(\tilde f_{v,t}(y)\bigr)$, in the
unnormalised log-beliefs of \eqref{eq:logspace-vec}, for the disagreement at round $t$, and let
\[
  \Delta_P=\max_y\max_{u,v}\ \sup_{q\in\Delta(\Y)}
  \bigl|R_u(y;q)-R_v(y;q)\bigr| ,
  \qquad
  \Delta_\beta=\max_{u,v}\Bigl|\frac1{\beta_u}-\frac1{\beta_v}\Bigr| ,
\]
which measure how far the agents' kernels and temperatures are from agreeing,
$\Delta_P$ vanishing exactly when all agents share a kernel and
$\Delta_\beta$ exactly when they share a temperature. Then for every $t$,
\begin{equation}\label{eq:case4-cond}
  \mathcal D_{t+1}\;\leq\;\gamma\,\mathcal D_t
  +\frac{1}{\beta_{\min}}\Delta_P+\Delta_\beta,
  \qquad
  \gamma=\tau_1(A)\Bigl(1+\frac{1}{2\beta_{\min}}\Bigr) .
\end{equation}
Consequently, if
\begin{equation}\label{eq:case4-threshold}
  \tau_1(A)<1
  \qquad\text{and}\qquad
  \beta_{\min}>\frac{\tau_1(A)}{2\bigl(1-\tau_1(A)\bigr)} ,
\end{equation}
so that $\gamma<1$, then
\[
  \mathcal D_t\;\leq\;\gamma^{\,t}\mathcal D_0+\bar{\mathcal D},
  \qquad
  \bar{\mathcal D}=\frac{1}{1-\gamma}\Bigl(\frac{1}{\beta_{\min}}\Delta_P+\Delta_\beta\Bigr) .
\]
The disagreement therefore contracts at rate $\gamma$ to within the residual
$\bar{\mathcal D}$ of zero, and to zero itself when the agents share a kernel
and a temperature.
\end{theorem}

\begin{figure}[t]
\centering
\includegraphics[width=0.62\textwidth]{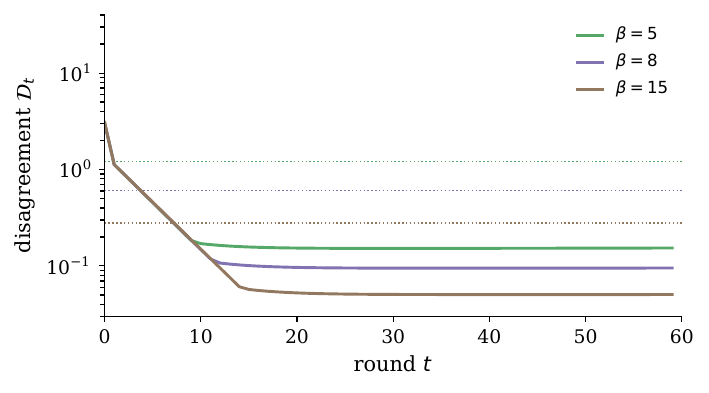}
\caption{Case~(III) on the instance of Section~\ref{sec:intro-example}: the
disagreement $\mathcal D_t$ of Theorem~\ref{thm:case4-directed} at three common
temperatures, with the theorem's residual $\bar{\mathcal D}$ at each as a dotted
line. See Example~\ref{ex:contract}.}
\label{fig:contract}
\end{figure}

\begin{example}[Contraction to a residual]
\label{ex:contract}
Run \eqref{eq:case4} on the instance of Section~\ref{sec:intro-example}, for
which $\tau_1(A)=0.8$, so that \eqref{eq:case4-threshold} requires $\beta_{\min}>2$. Take a common temperature $\beta\in\{5,8,15\}$, which makes
$\Delta_\beta=0$, and an initial profile at which the agents disagree, with
$\mathcal D_0=3.07$. Figure~\ref{fig:contract} shows $\mathcal D_t$. It falls
geometrically over the first thirty or so rounds and then stays well below the
residual, at about $0.15$, $0.09$ and $0.05$ against
$\bar{\mathcal D}=1.21$, $0.61$ and $0.28$. The bound holds with a wide margin on this instance. The level is lower at higher temperature, as the
residual is. Over longer runs $\mathcal D_t$ drifts slightly upwards, which the theorem allows, since it bounds each $\mathcal D_t$ only by
$\gamma^t\mathcal D_0+\bar{\mathcal D}$.
\end{example}

Theorem~\ref{thm:case4-directed} bounds how far the agents are from \emph{each other}. It does not bound the distance of any of them from a fixed point. The agents may draw closer to one another while none of them converges. It constrains the differences between the agents' log-beliefs, which the averaging contracts. Their $\pi$-weighted average is left unchanged by the averaging and moves only through the signal, by \eqref{eq:case4-common}. The theorem therefore gives no information about where the agents go. The disagreement is
nonetheless what the rest of the section runs on. In
Section~\ref{sec:cesaro} it bounds the slack $\bar\Delta_T$ of
Theorem~\ref{thm:cesaro}, and in Section~\ref{sec:case4-further}, under
\eqref{eq:case4-threshold}, a response that beats every other in every column
of $\bar P$ by more than $2\bar{\mathcal D}$ attracts every agent from every
initialisation.

\subsection{Time averages approximate the aggregated maximal lottery}
\label{sec:cesaro}

The introduction to this section showed that the collective policy
$p_t\propto\prod_uq_{u,t}^{\pi_u}$ evolves by exponential weights. That turns
into a guarantee for its time average, which holds whether or not the beliefs converge. The average is an approximate maximal lottery of $\bar P$, with an
explicit slack. The object is the average of the collective policy over
rounds, which is neither any agent's belief nor the policy at any round, and
the guarantee rests on the regret bound for exponential weights.

\begin{theorem}[The time average is an $\varepsilon$-maximal lottery]
\label{thm:cesaro} Let $\pi$ be a stationary distribution of $A$, a probability vector with $\pi^\top A=\pi^\top$; under Assumption~\ref{ass:graph} it is the unique one. Let $\tilde{\mathbf f}_t$ be the unnormalised log-beliefs of \eqref{eq:logspace-vec}, write $\eta=\sum_v\pi_v/\beta_v$, and let $\bar P$ be the influence-weighted kernel of \eqref{eq:pbar}. Set \[ F_t=\sum_v\pi_v\tilde f_{v,t},\qquad p_t(y)\propto e^{F_t(y)},\qquad \bar p_T=\frac1T\sum_{t<T}p_t , \]
so that $p_t$ is the collective policy of \eqref{eq:case3-finite}, the
$\pi$-weighted geometric mean of the agents' beliefs, and $\bar p_T$ its
average over the first $T$ rounds, and write
$M_0=\max_y\log\bigl(1/p_0(y)\bigr)$ for the starting term, which equals
$\log|\Y|$ when the collective policy starts uniform.
Define the \emph{aggregation defect}
\[
  \Delta_t=\sum_v\frac{\pi_v/\beta_v}{\eta}
  \bigl\|\check q_{v,t}-p_t\bigr\|_1,
  \qquad
  \bar\Delta_T=\frac1T\sum_{t<T}\Delta_t .
\]
Then for every $q\in\Delta(\Y)$ and every $T\geq1$,
\begin{equation}\label{eq:cesaro}
  \E_{y\sim\bar p_T,\;y'\sim q}\bigl[\bar P(y,y')\bigr]
  \;\geq\;\frac12-\varepsilon_T ,
  \qquad
  \varepsilon_T=\underbrace{\frac{M_0}{\eta T}}_{\text{horizon}}
  +\underbrace{\frac{\eta}{8}}_{\text{step size}}
  +\underbrace{\bar\Delta_T}_{\text{disagreement}} .
\end{equation}
That is, $\bar p_T$ is an $\varepsilon_T$-maximal lottery of $\bar P$ in
the sense of Definition~\ref{def:maxlot}. In particular, if the agents reach consensus, so that $\bar\Delta_T\to0$,
then $\limsup_T\varepsilon_T\leq\eta/8$. Scaling every temperature by a common factor so that $\eta=\eta_T=1/\sqrt T$, which makes each $\beta_v$ grow as $\sqrt T$ and leaves $\bar P$ unchanged, gives $\varepsilon_T=O\bigl(M_0/\sqrt T\bigr)+\bar\Delta_T$, so that
$\bar p_T$ converges to the set of maximal lotteries of $\bar P$
whenever $\bar\Delta_T\to0$.
\end{theorem}

Theorem~\ref{thm:cesaro} identifies the rule that case~(III) implements. The
time average of the collective policy is an approximate maximal lottery of
the influence-weighted kernel $\bar P$. The rule is therefore ordinal, using the
agents' preferences only through their comparison probabilities, and each
agent counts with the weight $\pi_v/\beta_v$ it also has in case~(I), where the
reference is likewise aggregated. The lottery is of $\bar P$ rather than of any agent's own kernel, and the property belongs to the time average $\bar p_T$; a single $p_t$ need not have it. The slack has three sources. The horizon term
$M_0/(\eta T)$ vanishes as the run lengthens. The step-size term $\eta/8$ is the
price of taking finite steps. The disagreement term $\bar\Delta_T$ is the
networked content: it measures how far each agent's challenger is from the
collective policy, and the next result bounds it by the disagreement of
Theorem~\ref{thm:case4-directed}.

\begin{proposition}[Controlling the aggregation defect]
\label{prop:defect} Let $\pi$ be as in Theorem~\ref{thm:cesaro}, and recall from Theorem~\ref{thm:case4-directed} the disagreement
$\mathcal D_t$ between the agents, its contraction factor $\gamma$ and the
spreads $\Delta_P$ and $\Delta_\beta$. The defect of Theorem~\ref{thm:cesaro}, how far each agent's challenger is from the collective policy, obeys $\Delta_t\leq2\mathcal D_t$, so that whenever
$\gamma<1$,
\[
  \bar\Delta_T\;\leq\;\frac{2\mathcal D_0}{(1-\gamma)T}
  +\frac{2}{1-\gamma}\Bigl(\frac{1}{\beta_{\min}}\Delta_P+\Delta_\beta\Bigr) .
\]
In particular $\bar\Delta_T=O(1/T)$ when the agents share a kernel and a
temperature, and otherwise $\limsup_T\bar\Delta_T\leq2\bar{\mathcal D}$, with $\bar{\mathcal D}$ the residual of Theorem~\ref{thm:case4-directed}.
\end{proposition}

\noindent
Unlike the other two terms, $\bar\Delta_T$ is removed by a longer horizon only
when the agents share a kernel and a temperature. Otherwise its limit is at most $2\bar{\mathcal D}$, so that
\[
  \limsup_T\varepsilon_T\;\leq\;\frac{\eta}{8}+2\bar{\mathcal D}
  =\frac{\eta}{8}+\frac{2}{1-\gamma}\Bigl(\frac{1}{\beta_{\min}}\Delta_P+\Delta_\beta\Bigr),
\]
and the residual falls as the temperatures rise together. The guarantee is informative only while this bound is below $\tfrac12$, since every distribution is an $\varepsilon$-maximal lottery at $\varepsilon=\tfrac12$.

The scaling $\eta=1/\sqrt T$ of Theorem~\ref{thm:cesaro} is the standard step size for exponential weights, read here as a temperature. It balances the horizon and step-size terms, whose sum $M_0/(\eta T)+\eta/8$ is smallest at $\eta=\sqrt{8M_0/T}$. Since $\eta=\sum_v\pi_v/\beta_v$, the scaling makes every temperature grow as $\sqrt T$, and the signal moves each log-belief by $O(1/\sqrt T)$ per round. Because the factor is common,
$\bar P$ and with it the target of the guarantee stay fixed. The scaling also
removes the disagreement. When $\tau_1(A)<1$, the threshold
\eqref{eq:case4-threshold} holds for all large $T$, $\gamma$ tends to
$\tau_1(A)$ and $\bar{\mathcal D}=O(1/\sqrt T)$, so
Proposition~\ref{prop:defect} gives $\bar\Delta_T=O(1/\sqrt T)$. All three
terms of the slack are then $O(1/\sqrt T)$, and $\bar p_T$ converges to the set
of maximal lotteries of $\bar P$ with no separate assumption on the disagreement.

That the property belongs to the time average matters for measurement. It gives no guarantee for a single checkpoint $p_t$, which can be far from $\bar p_T$
when the collective policy has not settled, as in Example~\ref{ex:maxlot} of
the Supplementary Material.

Theorem~\ref{thm:case2-lottery} gives a bound of the same shape for
case~(II), but for a different object, a fixed point and the arithmetic mean
of the beliefs, and the temperature enters it with the opposite sign: there
the term $\sum_vw_v\beta_vM_v$ falls as the agents optimise harder, while here
the step-size term $\eta/8$ grows. Case~(III) carries the log-beliefs forward
and accumulates the quality signals, so $1/\beta_v$ acts as a step size and
must be small; case~(II) does not carry them forward, by
\eqref{eq:logspace-vec}, and there $\beta_v$ only sets how sharply each agent
responds.

\subsection{Further results}
\label{sec:case4-further}

The fixed points of \eqref{eq:case4} include every atom. Which of them the beliefs reach is a separate question. We give three conditions on the kernels, each weaker than the one before. If $y^*$ dominates every column of every $P_v$ by a margin $\delta$, every agent converges to $\delta_{y^*}$, geometrically, with no condition on the graph or the temperature. If it dominates every column of the influence-weighted kernel
$\bar P$ by a margin exceeding twice the agents' residual disagreement, the same conclusion holds, again geometrically, now under Assumption~\ref{ass:graph} and the temperature condition of Theorem~\ref{thm:case4-directed}, which bounds that disagreement.
And if $y^*$ is merely a strict Condorcet winner of $\bar P$, the atom
$\delta_{y^*}$ attracts from nearby, while an atom whose margin is negative
repels; so the Condorcet condition is close to necessary for an atom to attract. Where $\bar P$ compares every pair strictly and has no Condorcet winner, every atom has a negative margin and repels, which is the case
Section~\ref{sec:cesaro} addresses through the time average. The statements
and proofs are in Section~\ref{supp:condorcet} of the Supplementary Material.

If some agents hold a fixed interior belief that the others observe, as a frozen reference model or an annotator pool with fixed standards would, and every updating agent attends to one of them, directly or through a chain of other agents, the solvability condition disappears: the updating agents have an interior fixed point for every choice of kernels and temperatures, determined by the frozen beliefs and the signals of the kernels. It is unique and reached geometrically when the temperatures are large enough relative to the weight each updating agent places on the updating agents, for instance when each such temperature exceeds $a/(2(1-a))$, with $a<1$ that weight; a temperature merely exceeding $a$ does not suffice. The atoms remain fixed points when every updating agent places some weight on the updating agents, but no trajectory started from an interior profile approaches them. Section~\ref{sec:case4-frozen} of the Supplementary Material gives the statements.

Section~\ref{supp:interior} of the Supplementary Material takes up interior
fixed points. It shows that the uniform profile is one whenever every kernel is balanced against it, for every graph and every vector of temperatures; that when the agents share a kernel and a temperature every interior consensus fixed point is unstable, at every temperature, unless every comparison is a tie; that otherwise one can be locally asymptotically stable, already with two agents and three responses; and that fixed points at which the agents disagree can also exist. Section~\ref{app:two-graphs} shows that the fixed-point results do not need the two roles to use the same weights: with a separate attention graph for references and another for challengers, Propositions~\ref{prop:case4-fp} and~\ref{prop:case4-spread} hold with $\pi$ and $D$ taken from the reference graph and the signals from the challenger graph. The convergence argument of Theorem~\ref{thm:case4-directed} does need a single graph.

\section{Related Work}
\label{sec:related}

\paragraph{Multiplayer preference optimisation.} \citet{wu2025mnpo} generalise
Nash learning from human feedback beyond two players. Their objective
regularises each policy towards a fixed reference while it competes against a
population of opponents, which is case~(II). The update they analyse regularises instead towards the geometric mean of the opponents, which is
case~(III). That update is \eqref{eq:case4} with $a_{vu}=1/(n-1)$ for $u\neq v$,
a shared kernel, and $\beta=1/\eta$ for their learning rate $\eta$, so the
comparison below concerns it rather than the objective. They obtain an
$O(1/\sqrt T)$ guarantee for the time-averaged policy, which is
Theorem~\ref{thm:cesaro} on that graph, where the agents are identical and
start together so the disagreement term vanishes. They identify the
heterogeneous case, in which distinct oracles make the game no longer
constant-sum, as lacking formal guarantees. Here the graph is arbitrary, which brings the weights $\pi_v/\beta_v$ into play, and kernels and temperatures may differ: Theorem~\ref{thm:cesaro} keeps
the maximal-lottery conclusion for the influence-weighted aggregate,
Proposition~\ref{prop:defect} bounds the slack the heterogeneity adds, and
under the same $1/\sqrt T$ scaling that slack vanishes whenever
$\tau_1(A)<1$. The bound here extends theirs without improving on it.

\paragraph{Aggregating heterogeneous preferences.} Social choice has been
applied to alignment directly~\citep{conitzer2024social,park2024rlhf}, and the
finding that the opinions language models express align with some
demographic groups far more than others motivates the
question~\citep{santurkar2023}. \citet{maura2025} propose maximal lotteries
as the target of alignment and show that game-theoretic methods such as Nash
learning from human feedback emulate them for a single model trained on a
population's comparisons. Here a maximal lottery is not chosen as the target
but arises from the dynamics, for a kernel whose weights the graph and the
temperatures determine. The electorate is taken as given in that literature. Here it is determined by who attends to whom. \citet{brandl2024}
show that a natural stochastic process converges to maximal lotteries, a
mechanism different from the one here but with the same limit.

\paragraph{Preference-based alignment of a single model.} At $n=1$ the
dynamics reduce to this setting, and both of its branches appear as special
cases: the reward-based branch, in which a reward model is fitted to
comparisons and maximised under a KL penalty towards a fixed
reference~\citep{ziegler2019,ouyang2022}, with direct methods such as
DPO~\citep{rafailov2023}; and the preference-based branch, in which the policy
is compared with its own current behaviour~\citep{munos2023}. In both the
preferences are a single agent's, and no question of aggregation arises. The
distinction persists in the network as the challenger axis of
Section~\ref{sec:fourcases} and decides whether the rule is computed from the rewards or from the comparisons they induce. The reference
axis has no counterpart at $n=1$, yet it decides whether a collective rule exists.

\paragraph{Opinion dynamics and social learning.} Applied to log-beliefs, the
averaging recursion is DeGroot's model of opinion~\citep{degroot1974,golub2010},
in which the stationary distribution of the communication matrix determines whose view prevails. That recursion is linear and unforced, so its limit is
the $\pi$-weighted average of the initial beliefs. Non-Bayesian social
learning adds a private signal at every round. \citet{jadbabaie2012} combine
the beliefs arithmetically, whereas the log-linear rule of \citet{molavi2018}
combines them geometrically, each agent adding the log-likelihood of its own
signal to the weighted average of its neighbours' log-beliefs. The recursion
\eqref{eq:case3} has that log-linear form, with the log-likelihood replaced by
the scaled reward $r_v/\beta_v$, so its limit depends on the agents'
preferences as well as on the graph. In case~(III) the added term depends on the beliefs themselves. The closest single result is~\citet{acemoglu2013},
who show that stubborn agents prevent consensus and that the \emph{expected}
beliefs of the others converge to the harmonic vector with boundary values at
the stubborn ones. Theorem~\ref{thm:case3-partial} has the same structure, the
matrix $H$ being exactly those absorption probabilities, and reduces to
harmonic interpolation when the rewards vanish. Relative to that result, the additions are the reward accumulated along the paths to the boundary and the
geometric pooling, both consequences of the KL penalty.

\paragraph{Learning dynamics in games.} The fully endogenous case is a game
played by learning agents. \citet{freund1999} show that when both sides of a
zero-sum game update by multiplicative weights the average of play approaches
the value, which is the mechanism behind Theorem~\ref{thm:cesaro};
\citet{bailey2018} show the last iterate instead diverges, which is why the
collective rule is stated for the time average, and \citet{cheung2018}
obtains the same divergence, independently, for graphical constant-sum games.
The difference here is that the game is played across a graph, and the aggregate each agent faces is its own neighbourhood's.

\paragraph{Best-response dynamics.} This paper inherits from best-reply
dynamics the convergence of the iteration under a contraction
condition~\citep{rosen1965,gabay1980}. Here the temperature supplies the contraction, and below the threshold the iteration can cycle. The question also differs. That literature fixes the game and asks whether best reply reaches its equilibrium and how fast. In this paper the equilibrium is the object of study, and the question is which
aggregation of conflicting preferences it represents. The game is also
unusual, its payoffs being preferences against neighbourhood aggregates rather
than a fixed matrix.

\paragraph{Aggregation in deployed systems.} Aggregated references already
occur in deployed algorithms: ensemble distillation for federated
learning~\citep{lin2020} trains a server against an average of its clients'
outputs, which is the aggregated reference studied here, with later variants
reweighting the ensemble~\citep{chen2021fedbe,sattler2021fedaux}. Model
fusion~\citep{wan2024knowledge} applies the same aggregation to sources that
are read but never updated, and on the reward side ensembles of reward models
reduce overoptimisation during RLHF~\citep{coste2023}. None of this work asks
which aggregation rule the resulting policies implement.

\paragraph{Distributed learning.} Distributed
optimisation~\citep{nedic2009,sayed2014} and federated
learning~\citep{mcmahan2017} average parameters rather than distributions.
Federated RLHF shares the motivating setting of this paper but not its
dynamics: \citet{fan2025fedrlhf} keep a personalised policy at each client and
prove convergence and sample-complexity bounds, while \citet{wu2024fedrlhf}
aggregate \emph{preference models} trained at the clients and evaluate the
result empirically. Neither regularises an agent towards an average of the other agents' policies. In both, the aggregation is performed on the reward
side by an explicit choice of pooling rule rather than induced by the
dynamics. In network games, influence is measured by Katz--Bonacich
centrality~\citep{ballester2006,bramoulle2014}, a different functional of the
graph from the stationary distribution that governs these dynamics.

\section{Conclusion}
\label{sec:conclusion}

The paper asks which rule of preference aggregation a network of agents
implements when each optimises its own policy against the others', and finds
that the kind of rule is determined by two modelling choices. The distribution
an agent is penalised for leaving, the reference, decides whether the
preferences are aggregated into a single rule for the network or into one rule
per agent. The distribution its responses are compared against, the
challenger, decides what that rule is computed from: the magnitudes of the
rewards, giving a cardinal rule, or only the comparisons they induce, giving
an ordinal one. That is the division social choice draws between scoring rules
and Condorcet-consistent ones, and the four cases are the four ways of
combining the two choices. The two choices also have different effects. Aggregating the challenger makes an approximate maximal lottery appear. One appears in case~(II) as well, for any weighting of
the agents (Theorem~\ref{thm:case2-lottery}). Aggregating the reference selects a particular weighting, the one we call the network's rule. Where a collective rule exists, an agent's weight is its share
of the network's long-run attention divided by how cautiously it optimises,
fixed by the communication weights and the temperatures alone, with no access
to anybody's preferences.

Agents that receive equal attention can therefore have unequal weight, since the one that optimises harder counts for more. Where the rule is cardinal,
an agent's influence also grows with the scale of its rewards, doubling them
doubling its contribution; where the rule is ordinal, its contribution stays
bounded however its rewards are scaled. Disagreement, finally, need not be transient. With a fixed reference the agents converge, above the threshold of
Proposition~\ref{prop:case2-geom}, to policies that in general differ, and the differences persist.

The collective rules rest on influence circulating, as
Assumption~\ref{ass:graph} requires. Where it does not, as with agents that
never update or a graph through which influence flows one way, no single
weighting applies. With a fixed challenger each agent then aggregates the
others through the paths that reach it (Theorem~\ref{thm:case3-partial}); with
an aggregated challenger a boundary of agents that never update can leave
several fixed points (Remark~\ref{rem:frozen-several}). And where both roles
are aggregated, the guarantee that holds without further conditions is for the
time average of the collective policy. The beliefs themselves come together
above a temperature threshold (Theorem~\ref{thm:case4-directed}), and converge to a single response under the conditions of Section~\ref{sec:case4-further}.

Several directions remain open. The first concerns case~(III) beyond the time average. How large the set of initial beliefs that converge to a single response is, and what the beliefs do when no response wins the influence-weighted comparison, are open for networks of several agents; interior fixed points can be
stable or unstable (Proposition~\ref{prop:case4-interior}). In case~(II), uniqueness below the temperature threshold is settled only in special cases.

The second concerns dynamics between the four extremes. Each role here is either
fixed or aggregated, and the Supplementary Material treats only one intermediate
setting (Proposition~\ref{prop:interpolate}). A promising candidate mixes both roles
between an agent's anchor and its neighbours' aggregate. For a single agent this is
Nash-MD~\citep{munos2023}, whose last iterate converges to the regularised Nash policy
at every temperature. Whether a network version keeps that property, and which
weighting of the agents its limit carries, would show whether last-iterate guarantees can be had without losing the collective rule.

Two further directions are to let the two roles use different attention weights,
which the fixed-point results already allow but the contraction of
Theorem~\ref{thm:case4-directed} does not, and to replace the exact maximisation in
each round by the approximate, sample-based optimisation used in practice, which
the experiment of Section~\ref{sec:case3-experiment} illustrates but the analysis
does not cover.

\section*{Acknowledgments}

Anthropic's Claude Opus 5.5 was used to assist with literature research, mathematical verification, the revision of proofs and, for some results, their initial derivation, exposition, and the generation of code for numerical computations. The author reviewed and verified all AI-assisted content and takes full responsibility for the contents of this paper.

\newpage
\begin{center}
{\large\bf Supplementary Material to\\[2pt]
``Aggregation Rules Induced by KL-Regularised\\
Preference Dynamics on Networks''}
\end{center}
\vspace{1em}

\appendix

\section{Notation, Proofs and Experimental Details}
\label{app:proofs}

This appendix fixes the notation, proves the results stated in the body, and
gives the details of the experiment of Section~\ref{sec:case3-experiment}.

\subsection{Notation}
\begin{center}
\small
\renewcommand{\arraystretch}{1.2}
\begin{tabular}{@{}l p{0.66\textwidth}@{}}
\toprule
$\Y$, $\Delta(\Y)$, $\Y_0$ & set of responses; simplex; zero-sum vectors on $\Y$\\
$A$, $a_{vu}$ & averaging matrix, row-stochastic; weight agent $v$ places on agent $u$\\
$\pi$, $D$, $\tau_1(A)$ & stationary distribution; deviation matrix \eqref{eq:D}; Dobrushin coefficient\\
$\lambda_2(A)$, $N$, $m$ & eigenvalue of $A$ of largest modulus other than $1$; $A-\one\pi^\top$; largest Jordan block at modulus $|\lambda_2(A)|$\\
$P_v$, $R_v(y;q)$, $B_P$ & preference kernel; $\E_{y'\sim q}[P_v(y,y')]$; bilinear form of the kernel\\
$\mu_v$, $\nu_v$ & fixed reference (prior); fixed challenger\\
$\bar q_{v,t}$, $\check q_{v,t}$ & neighbourhood geometric mean; arithmetic mean\\
$\beta_v$, $B$, $\eta$ & temperature; $\mathrm{diag}(\beta_v)$; $\sum_v\pi_v/\beta_v$\\
$\bar r$, $\bar P$ & influence-weighted reward $\sum_v(\pi_v/\beta_v)r_v$; influence-weighted kernel \eqref{eq:pbar}\\
$\Alg$, $\Frz$, $Z$, $H$ & updating and frozen agents; fundamental and harmonic matrices\\
$\mathcal D_t$, $\bar{\mathcal D}$, $\gamma$ & disagreement; its residual (Theorem~\ref{thm:case4-directed}); contraction modulus\\
$\Delta_P$, $\Delta_\beta$ & largest difference between two agents' signals; between their inverse temperatures\\
$p_t$, $\bar p_T$, $M_0$ & collective policy; its time average; $\max_y\log(1/p_0(y))$\\
$\bar\Delta_T$ & time-averaged distance of the challengers from the collective policy\\
$\osc$, $\mathcal{N}^{\mathrm{in}}_v$, $\Phi_{v,t}$ & oscillation $\max-\min$; in-neighbourhood of $v$; $y$-independent log-normaliser\\
$\sigma$, $\Pi$, $J$, $J_v$ & softmax on $\mathbb R^{\Y}$; projector onto $\Y_0$; all-ones matrix; $\mathrm{diag}(q_v)-q_vq_v^\top$\\
$\phi_{v,t}(y,y')$ & log-odds of $y$ against $y'$ at agent $v$ and round $t$\\
$\varepsilon$ & slack or remainder; see below\\
\bottomrule
\end{tabular}
\end{center}

The symbol $\varepsilon$ is used in three ways, each local to its statement:
the slack of an approximate maximal lottery, $\varepsilon_w$ in
Theorem~\ref{thm:case2-lottery} and $\varepsilon_T$ in
Theorem~\ref{thm:cesaro}; the remainder $\varepsilon_{v,t}$ of
Theorem~\ref{thm:case3}; and, inside proofs, a small positive number.

\subsection{Section~\ref{sec:setup}: shared structure}
\label{app:shared}

This subsection proves the results of Section~\ref{sec:fixedpoints}, which
apply to all four dynamics, and gives the two lemmas that the contraction
estimates of Sections~\ref{sec:case2} and~\ref{sec:case4} rest on.

\paragraph{The deviation matrix.} The deviation matrix $D$ of \eqref{eq:D} is the object through which the graph
enters every limit below, and the next lemma collects what is used of it. Its first two parts concern $A$:
the kernels of $I-A$, and the geometric convergence of $A^t$ to
$\one\pi^\top$. The last two concern $D$ itself: its closed form as a
resolvent, and that it inverts $I-A$ in every direction except that of the
stationary distribution, which it annihilates. Parts~(a) and~(b) are standard facts about a primitive
matrix~\citep{seneta2006,horn1985}, and parts~(c) and~(d) the standard
properties of the deviation matrix, which is the group inverse of
$I-A$~\citep{kemeny1960,meyer1975}; the proof is included so that the
constants and the form of the decay used later are explicit.

\begin{lemma}[Properties of $D$]
\label{lem:Dprops}
Assume that Assumption~\ref{ass:graph} holds. Then
\begin{enumerate}
\item[(a)] up to scaling, $\one$ is the only vector with $A\mathbf x=\mathbf x$ and $\pi$ the only one with $\pi^\top A=\pi^\top$; adding the rank-one term makes
  $I-A+\one\pi^\top$ invertible;
\item[(b)] $\|A^t-\one\pi^\top\|\leq Ct^{m-1}|\lambda_2(A)|^t$ for all $t\geq m$, where $m$ is the largest Jordan block among the eigenvalues of modulus $|\lambda_2(A)|$;
\item[(c)] the series in \eqref{eq:D} converges absolutely, and
  $D=(I-A+\one\pi^\top)^{-1}-\one\pi^\top$;
\item[(d)] $(I-A)D=D(I-A)=I-\one\pi^\top$, and $D\one=0$, $\pi^\top D=0$.
\end{enumerate}
\end{lemma}

\begin{proof}
Write $\Pone=\one\pi^\top$ for the projection onto the direction of the
stationary distribution and put
\[
  N=A-\Pone .
\]
\emph{Step 1: $N$ separates the two parts of $A$.} From $A\one=\one$, $\pi^\top A=\pi^\top$ and $\pi^\top\one=1$ we have
$\Pone^2=\Pone$ and
\[
  \Pone N=N\Pone=0, \qquad N\one=0, \qquad \pi^\top N=0,
  \qquad\text{hence}\qquad A^t=\Pone+N^t\quad (t\geq1).
\]

\emph{Step 2: $N$ is a contraction.} By the Perron--Frobenius theorem for
primitive matrices~\citep{seneta2006}, the eigenvalue $1$ of $A$ is simple and
every other eigenvalue has modulus below one. Since $A=\Pone+N$ with
$\Pone N=N\Pone=0$, the spectrum of $N$ is that of $A$ with the eigenvalue $1$
replaced by $0$, so
\[
  \rho(N)=|\lambda_2(A)|<1 . \] Writing $N=S\mathcal JS^{-1}$ in Jordan form, a block of size $k$ at an eigenvalue $\lambda$ has $t$-th power with entries $\binom{t}{j}\lambda^{t-j}$ for $0\leq j<k$, each bounded by a constant times $t^{k-1}|\lambda|^t$ once $t\geq k$ (for $\lambda=0$ the power vanishes from $t=k$ on). Taking the
largest block at modulus $\rho(N)$ and absorbing $S$ gives
$\|N^t\|\leq Ct^{m-1}\rho(N)^t$ for $t\geq m$, with $m$ the largest Jordan block among the
eigenvalues of modulus $|\lambda_2(A)|$. With
Step~1 this is the decay asserted in~(b), and being summable it also gives the
absolute convergence asserted in~(c).

\emph{Step 3: the kernels, giving~(a).} Because $\rho(N)<1$, the matrix
$I-N=I-A+\one\pi^\top$ is invertible, which is the second assertion of~(a).
For the first, a vector in the kernel of $I-A$ satisfies
$(I-N)x=\Pone x=(\pi^\top x)\one$, so $x=(\pi^\top x)(I-N)^{-1}\one=(\pi^\top x)\one$,
using $N\one=0$ from Step~1, and hence $x\in\mathrm{span}(\one)$; the same
argument on the left, with $\pi^\top N=0$, gives the left kernel.

\emph{Step 4: part~(c).} Summing Step~1 and using
$\rho(N)<1$,
\[
  D=\sum_{t\geq0}\bigl(A^t-\Pone\bigr)
   =(I-\Pone)+\sum_{t\geq1}N^t
   =(I-N)^{-1}-\Pone ,
\]
and $I-A+\Pone=I-N$, so the resolvent expression for $D$ is the same matrix.

\emph{Step 5: part~(d).} Using $\Pone N=0$ and $\Pone^2=\Pone$,
\[
  (I-A)D=(I-N-\Pone)\bigl[(I-N)^{-1}-\Pone\bigr]=I-\Pone ,
\]
and $D(I-A)=I-\Pone$ by the same computation on the other side. Finally
$(I-N)^{-1}\one=\one$ and $\pi^\top(I-N)^{-1}=\pi^\top$, which with
$D=(I-N)^{-1}-\Pone$ give $D\one=0$ and $\pi^\top D=0$.
\end{proof}

\paragraph{Proof of Theorem~\ref{thm:fredholm}.}

The theorem concerns \eqref{eq:fp-compact}, the fixed-point equation
$(I-A)\mathbf f^*(y)=B^{-1}\mathbf R^*(y)-\boldsymbol\Phi^*$ of the two
dynamics whose reference is aggregated. Part~(a) asserts that it is solvable exactly when the $\pi$-weighted signal
$\sum_v(\pi_v/\beta_v)R^*_v(y)$ does not depend on the response, with
\eqref{eq:solvability} then fixing its common value; part~(b) gives the solutions as
\eqref{eq:resolvent-sol}. Write $\mathbf g(y)$ for the right-hand side, so that
the equation reads $(I-A)\mathbf f^*(y)=\mathbf g(y)$.

\emph{Part (a), necessity.} Suppose a solution exists. Applying $\pi^\top$ and
using $\pi^\top A=\pi^\top$ from Assumption~\ref{ass:graph} annihilates the
left-hand side, so $\pi^\top\mathbf g(y)=0$ for every $y$. Written out, this is
$\sum_v(\pi_v/\beta_v)R^*_v(y)=\sum_v\pi_v\Phi^*_v$, which is
\eqref{eq:solvability}; since $\boldsymbol\Phi^*$ does not involve $y$, the
left-hand side is the same at every response.

\emph{Part (a), sufficiency.} Suppose $\pi^\top\mathbf g(y)=0$. By
Lemma~\ref{lem:Dprops}(d), $(I-A)D=I-\one\pi^\top$, so
\[
  (I-A)\,D\mathbf g(y)=\mathbf g(y)-\one\,\pi^\top\mathbf g(y)=\mathbf g(y),
\]
and $D\mathbf g(y)$ solves the equation. Solvability is therefore equivalent to
\eqref{eq:solvability}.

\emph{Part (b).} Let $\mathbf f^*$ be a solution, that is a function
$y\mapsto\mathbf f^*(y)$ satisfying \eqref{eq:fp-compact} at every response,
and write $\mathbf g(y)=B^{-1}\mathbf R^*(y)-\boldsymbol\Phi^*$ for the
right-hand side it produces. Here $\boldsymbol\Phi^*$ is determined by the
whole function, not by its value at one response; with $\mathbf g$ so fixed,
the argument runs at each $y$ separately.

By the sufficiency argument above, $D\mathbf g(y)$ also solves
$(I-A)\mathbf x=\mathbf g(y)$. Subtracting the two equations, the right-hand
sides cancel and $\mathbf x=\mathbf f^*(y)-D\mathbf g(y)$ satisfies $A\mathbf x=\mathbf x$, and Lemma~\ref{lem:Dprops}(a) identifies such vectors as the multiples of $\one$. So
$\mathbf f^*(y)=D\mathbf g(y)+c(y)\one$, the scalar free to vary with $y$
because $y$ was arbitrary. Expanding $\mathbf g(y)$ gives
\eqref{eq:resolvent-sol}.
\proofend

\paragraph{Averaging closes a spread.}
\label{app:dobrushin}

One round of averaging over a graph brings the agents closer together, at a
rate the ergodic coefficient measures. The bound below is used wherever a
disagreement is contracted.

\begin{lemma}[Averaging closes a spread]
\label{lem:dobrushin}
Let $A$ be row-stochastic and let each agent hold a number $b_v$. After one
round of averaging, the spread of those numbers across agents has shrunk by at
least the factor $\tau_1(A)$ of Definition~\ref{def:resolvent}:
\[
  \osc_v\bigl((A\mathbf b)_v\bigr)\;\leq\;\tau_1(A)\,\osc_u(b_u).
\]
If $A$ is primitive then $\tau_1(A^m)<1$ for some finite $m$, so the spread does
strictly shrink over $m$ rounds even when it need not over one.

Dually, for any $x\in\mathbb R^n$ with $\sum_vx_v=0$,
\[
  \|x^\top A\|_1\;\leq\;\tau_1(A)\,\|x\|_1 .
\]
Both are standard \citep[Sec.~4.3]{seneta2006}.
\end{lemma}

\begin{proof}
For any $v,v'$, $(A\mathbf b)_v-(A\mathbf b)_{v'}=\sum_u(a_{vu}-a_{v'u})b_u$, and
$\sum_u(a_{vu}-a_{v'u})=0$, so we may replace $b_u$ by
$b_u-\min_ub_u\in[0,\osc(b)]$ without changing the value. Bounding the
contribution of the positive part of $a_{vu}-a_{v'u}$ by $\osc(b)$ and that of
the negative part by $0$,
\[
  (A\mathbf b)_v-(A\mathbf b)_{v'}\leq\osc(b)\sum_u(a_{vu}-a_{v'u})^+
  =\tfrac12\|a_{v\cdot}-a_{v'\cdot}\|_1\,\osc(b),
\]
the last equality because the positive and negative parts of a zero-sum vector
carry equal mass. Maximising over $v,v'$, and over which of the two is the
larger, gives $\osc_v((A\mathbf b)_v)\leq\tau_1(A)\osc(b)$. If $A$ is primitive
then $A^m>0$ entrywise for some $m$, and any strictly positive stochastic matrix
has $\tau_1<1$ since distinct rows overlap.

For the dual statement, write $x=x^+-x^-$ with $x^\pm\geq0$; since $x$ sums
to zero, $\|x^+\|_1=\|x^-\|_1=h$ with $h=\tfrac12\|x\|_1$. If $h=0$ there is
nothing to prove. Otherwise, writing $a_{v\cdot}$ for row $v$ of $A$,
\[
  x^\top A=\sum_vx^+_va_{v\cdot}-\sum_{v'}x^-_{v'}a_{v'\cdot}
  =\frac1h\sum_{v,v'}x^+_vx^-_{v'}\bigl(a_{v\cdot}-a_{v'\cdot}\bigr) ,
\]
and since $\|a_{v\cdot}-a_{v'\cdot}\|_1\leq2\tau_1(A)$ and
$\sum_{v,v'}x^+_vx^-_{v'}=h^2$, this gives
$\|x^\top A\|_1\leq2h\,\tau_1(A)=\tau_1(A)\|x\|_1$.
\end{proof}

\paragraph{Sensitivity of the quality signal.}
\label{app:signal-lip}

Wherever an agent's challenger is aggregated, a round's signal depends on the
other agents' current beliefs, so bounding how far the signal can move when
those beliefs move is the first step in every contraction argument in the
paper. The lemma below does that. With Lemma~\ref{lem:softmax} it gives, in Section~\ref{sec:case2}, that an agent's belief moves by at most $1/(2\beta_v)$ times its neighbours' movement; the condition $\rho(B^{-1}A)<2$ of Proposition~\ref{prop:case2-fp}(b) controls these factors across the network. It is used again in the proof of Theorem~\ref{thm:case4-directed}, and through that theorem in the slack of Theorem~\ref{thm:cesaro}. The factor of $\tfrac12$ matters: the obvious estimate $\|q-q'\|_1$ would
double every temperature threshold stated in this paper.

\begin{lemma}[Sensitivity of the signal]
\label{lem:signal-lip}
For any preference kernel $P$ and any two distributions $q,q'$ on $\Y$,
\[
  \bigl|R(y;q)-R(y;q')\bigr|\;\leq\;\tfrac12\|q-q'\|_1
  \qquad\text{for every }y,
\]
the right-hand side being the total variation distance between $q$ and $q'$.
\end{lemma}

\begin{proof}
The signal is linear in the challenger, so that
\[
  R(y;q)-R(y;q')=\sum_{y'}\bigl(q(y')-q'(y')\bigr)P(y,y') .
\]
Both $q$ and $q'$
are distributions, so $\sum_{y'}(q-q')(y')=0$ and any constant may be
subtracted from every entry of $P$ without changing the sum; subtracting
$\tfrac12$,
\[
  R(y;q)-R(y;q')=\sum_{y'}\bigl(q(y')-q'(y')\bigr)
  \bigl[P(y,y')-\tfrac12\bigr] .
\]
A preference kernel takes values in $[0,1]$, so
$|P(y,y')-\tfrac12|\leq\tfrac12$, and bounding the sum term by term gives
\[
  \bigl|R(y;q)-R(y;q')\bigr|
  \;\leq\;\tfrac12\sum_{y'}\bigl|q(y')-q'(y')\bigr|
  \;=\;\tfrac12\|q-q'\|_1 . \qedhere
\]
\end{proof}

\subsection{Section~\ref{sec:case3}: aggregated reference, fixed challenger}
\label{app:caseI}

\paragraph{Proof of Theorem~\ref{thm:case3}.}

Iterating \eqref{eq:case3} gives
$\tilde{\mathbf f}_t=A^t\tilde{\mathbf f}_0+\sum_{s<t}A^sB^{-1}\mathbf r$, and by
Remark~\ref{rem:normalisation} log-odds differences of the normalised beliefs
equal those of $\tilde{\mathbf f}_t$. Write
$\mathbf g=B^{-1}\bigl(\mathbf r(y)-\mathbf r(y')\bigr)$, so that
$\boldsymbol\phi_t=A^t\boldsymbol\phi_0+\sum_{s<t}A^s\mathbf g$. Since
$A\one\pi^\top=\one\pi^\top$ and $(I-A)D=I-\one\pi^\top$
(Lemma~\ref{lem:Dprops}(d)),
\[
  (I-A^t)D=\sum_{s<t}A^s(I-A)D=\sum_{s<t}A^s\bigl(I-\one\pi^\top\bigr)
  =\sum_{s<t}A^s-t\,\one\pi^\top ,
\]
that is, $\sum_{s<t}A^s=t\,\one\pi^\top+(I-A^t)D$ for every $t$. Substituting,
and using $\pi^\top D=0$ to write $A^tD=(A^t-\one\pi^\top)D$,
\[
  \boldsymbol\phi_t=\bigl(\pi^\top\boldsymbol\phi_0+t\,\pi^\top\mathbf g\bigr)\one
  +D\mathbf g+\boldsymbol\varepsilon_t ,
  \qquad
  \boldsymbol\varepsilon_t=\bigl(A^t-\one\pi^\top\bigr)\bigl(\boldsymbol\phi_0-D\mathbf g\bigr),
\]
exactly, at every $t$. The first term is the first underbrace of \eqref{eq:case3-logodds}, since $\pi^\top\mathbf g=\bar r(y)-\bar r(y')$, and $D\mathbf g$ is the agent-dependent offset. For the rate, $A^t-\one\pi^\top=N^t$ for $t\geq1$,
with $N=A-\one\pi^\top$, and Lemma~\ref{lem:Dprops}(b) bounds $\|N^t\|$ by
$Ct^{m-1}|\lambda_2(A)|^t$ for $t\geq m$, so
$\|\boldsymbol\varepsilon_t\|_\infty\leq Ct^{m-1}|\lambda_2(A)|^t\,
\|\boldsymbol\phi_0-D\mathbf g\|_\infty$, which is the statement of the
theorem. The closed form also gives a bound with no constant. For any vector
$\mathbf z$, Lemma~\ref{lem:dobrushin} applied $t$ times gives
$\osc(A^t\mathbf z)\leq\tau_1(A)^t\osc(\mathbf z)$, and
$\pi^\top A^t\mathbf z=\pi^\top\mathbf z$ is a convex combination of the entries of
$A^t\mathbf z$, so every entry of $(A^t-\one\pi^\top)\mathbf z$ lies within
$\tau_1(A)^t\osc(\mathbf z)$ of zero. With $\mathbf z=\boldsymbol\phi_0-D\mathbf g$,
$|\varepsilon_{v,t}|\leq\tau_1(A)^t\osc\bigl(\boldsymbol\phi_0-D\mathbf g\bigr)$.
\proofend

\paragraph{Proof of Corollary~\ref{cor:case3-ties}.}

\emph{Mass outside $M$ vanishes.} Take $y\in M$ and $y''\notin M$. Then $\bar r(y)-\bar r(y'')>0$, so by \eqref{eq:case3-logodds} the log-odds of $y$ against $y''$ grow linearly in $t$ at that rate, and the odds $q_{v,t}(y'')/q_{v,t}(y)$ decay exponentially; summing over
the finitely many $y''\notin M$ gives $\sum_{y''\notin M}q_{v,t}(y'')\to0$.

\emph{The limit on $M$.} Let $y,y'\in M$, so that $\bar r(y)=\bar r(y')$ and
the growing term of \eqref{eq:case3-logodds} is zero at every $t$. What
remains is $\pi^\top\boldsymbol\phi_0+[D\,B^{-1}(\mathbf r(y)-\mathbf
r(y'))]_v+\varepsilon_{v,t}$, and $\varepsilon_{v,t}\to0$, which gives the
stated limiting log-odds. Since the mass outside $M$ vanishes and the log-odds
within $M$ converge, $q_{v,t}$ converges to the distribution on $M$ with those
log-odds.

\emph{The Gibbs form.} Both terms on the right are differences of a single
function of $y$: writing $g_v(y)=\pi^\top\log\mathbf q_0(y)+[D\,B^{-1}\mathbf
r(y)]_v$, where $\log\mathbf q_0(y)=(\log q_{u,0}(y))_{u\in V}$, the limiting
log-odds are $g_v(y)-g_v(y')$. A distribution on $M$ whose log-odds are
differences of $g_v$ is $q^*_v\propto e^{g_v}$ restricted to $M$, which is the
stated form with $\bar q_0\propto\prod_uq_{u,0}^{\pi_u}$.
\proofend

\paragraph{Agents that never update.}
\label{sec:fp-boundary}
Some agents may hold a belief the others observe without ever revising it: a
frozen reference model, or an annotator pool whose standards are fixed in
advance. This part sets out the structure it imposes. Partition $V=\Alg\cup\Frz$ into
updating agents running \eqref{eq:general} and \emph{frozen} agents with
frozen beliefs $q_{v,t}=q_{v,0}$, and partition $A$ conformally: $\AUU$ holds the weights that updating agents
place on one another and $\AUF$ those they place on frozen agents, so that
$(\AUU\mathbf f_{\Alg})_v+(\AUF\mathbf f_{\Frz})_v$ is the average agent
$v\in\Alg$ forms. The updating block obeys
\begin{equation}\label{eq:partial-block}
  \tilde{\mathbf f}_{\Alg,t+1}(y)=\AUU\tilde{\mathbf f}_{\Alg,t}(y)
  +\AUF\mathbf f_{\Frz}(y)+B_{\Alg}^{-1}\mathbf R_{\Alg,t}(y),
  \qquad\mathbf f_{\Frz}(y)=(\log q_{v,0}(y))_{v\in\Frz}.
\end{equation}

Say that agent $v$ \emph{reaches} agent $u$ if it attends to $u$ directly or through a chain of agents, that is, if there are $v=w_0,w_1,\dots,w_m=u$ with $a_{w_iw_{i+1}}>0$ for each $i$; in the graph $G$ of Section~\ref{sec:setup} this is a directed path from $u$ to $v$.  When $I-\AUU$ is invertible, set
\[
  Z=(I-\AUU)^{-1},\qquad H=Z\AUF ,
\]
the \emph{fundamental} and \emph{harmonic} matrices; these appear in every
statement about boundaries below. Both read as statements about the walk that
steps from an agent $v$ to one of the agents it observes, moving to $u$ with
probability $a_{vu}$,
and stops on reaching a frozen agent.

\begin{lemma}[Absorbing boundary]
\label{lem:substoch}
$\AUU$ is substochastic. Say that the graph has a \emph{reachable boundary} if every updating agent reaches some frozen agent, in the sense defined above. If it
does, then:
\begin{enumerate}[label=(\alph*),leftmargin=*]
  \item $\rho(\AUU)<1$, so $\AUU^t\to0$ geometrically and
        $Z=\sum_{s\geq0}\AUU^s$ exists with non-negative entries;
  \item for the walk started at $v$, $z_{vu}$ is the expected number of visits to
        $u$ before it is absorbed, and $h_{vb}$ the probability that it is
        absorbed at $b$;
  \item $H\one_{\Frz}=\one_{\Alg}$, so each row of $H$ is a probability
        distribution over the frozen agents.
\end{enumerate}
\end{lemma}

\begin{proof}
\emph{Step 1: $\AUU$ is substochastic.} For $v\in\Alg$,
\[
  \sum_{u\in\Alg}a_{vu}=1-\sum_{b\in\Frz}a_{vb}\;\leq\;1 .
\]

\emph{Step 2: the row sums of a power fall below one.} Write
$r_t(v)=\sum_{u\in\Alg}(\AUU^t)_{vu}$ for $v\in\Alg$. Since $\AUU$ is
non-negative with row sums at most one,
$r_{t+1}(v)=\sum_w(\AUU^t)_{vw}\sum_ua_{wu}\leq r_t(v)\leq1$, so each $r_t(v)$
is non-increasing in $t$. For $v\in\Alg$ let $d(v)$ be the length of a
shortest chain $v=w_0,w_1,\dots,w_d$ with $a_{w_iw_{i+1}}>0$ and $w_d\in\Frz$;
its updating agents are distinct, so $d(v)\leq|\Alg|$. We show
$r_{d(v)}(v)<1$ by induction on $d(v)$. If $d(v)=1$, then $a_{vb}>0$ for some
$b\in\Frz$, and $r_1(v)=1-\sum_{b'\in\Frz}a_{vb'}<1$. If $d(v)=d>1$, the next
agent $w$ of a shortest chain is updating, with $a_{vw}>0$ and $d(w)=d-1$, so
\[
  r_d(v)=\sum_{u\in\Alg}a_{vu}\,r_{d-1}(u)
  \;\leq\;\sum_{u\neq w}a_{vu}+a_{vw}\,r_{d-1}(w)
  \;<\;\sum_{u\in\Alg}a_{vu}\;\leq\;1 ,
\]
using $r_{d-1}\leq1$ and the induction hypothesis $r_{d-1}(w)<1$. By
monotonicity $r_{|\Alg|}(v)\leq r_{d(v)}(v)<1$ for every $v$, so
$c=\max_{v\in\Alg}r_{|\Alg|}(v)<1$.

\emph{Step 3: geometric decay and the fundamental matrix.} If every row of a
non-negative matrix $M$ sums to at most $a$ and every row of a non-negative $N$
to at most $b$, every row of $MN$ sums to at most $ab$. Writing $t=q|\Alg|+s$
with $0\leq s<|\Alg|$, this gives
$\|\AUU^t\|_\infty\leq c^q\leq c^{\,t/|\Alg|-1}$, so $\AUU^t\to0$ geometrically
and $\sum_{s\geq0}\AUU^s$ converges. Since
$(I-\AUU)\sum_{s<T}\AUU^s=I-\AUU^T\to I$, its sum is $Z=(I-\AUU)^{-1}$, and
$Z\geq0$ entrywise because each summand is. Finally
$\rho(\AUU)\leq\|\AUU^{|\Alg|}\|_\infty^{1/|\Alg|}\leq c^{1/|\Alg|}<1$.

\emph{Step 4: the rows of $H$ sum to one.} Row-stochasticity of $A$ reads
$\AUU\one_{\Alg}+\AUF\one_{\Frz}=\one_{\Alg}$, that is
$(I-\AUU)\one_{\Alg}=\AUF\one_{\Frz}$, whence
\[
  H\one_{\Frz}=Z\AUF\one_{\Frz}=Z(I-\AUU)\one_{\Alg}=\one_{\Alg}.
\]

\emph{Step 5: the probabilistic reading.} These are the standard identifications
for an absorbing chain~\citep{kemeny1960,norris1997}: $Z$ is the fundamental matrix of the walk absorbed on $\Frz$, so $z_{vu}$ is
the expected number of visits to $u$ before absorption, and $H$ is its matrix of first-absorption
probabilities~\cite[Ch.~1]{norris1997}.

\end{proof}

Since $H\one_{\Frz}=\one_{\Alg}$, each updating agent's total exposure to the boundary is one, and the graph determines how that exposure is distributed among the frozen agents. Non-updating agents play the role of the stubborn agents of DeGroot-type opinion dynamics~\citep{degroot1974,friedkin1990}; Section~\ref{sec:related} compares the limit with that of \citet{acemoglu2013}.

\paragraph{Proof of Theorem~\ref{thm:case3-partial}.}

By Lemma~\ref{lem:substoch}, $\rho(\AUU)<1$. The frozen agents do not move,
so the forcing
$\mathbf g=\AUF\mathbf f_{\Frz}(y)+B_{\Alg}^{-1}\mathbf r_{\Alg}(y)$ in
\eqref{eq:partial-block} is constant in $t$, and iterating that equation gives
\[
  \tilde{\mathbf f}_{\Alg,t}
  =\AUU^t\tilde{\mathbf f}_{\Alg,0}+\sum_{s<t}\AUU^s\mathbf g .
\]
Because $\rho(\AUU)<1$, the first term vanishes, at the rate $\rho(\AUU)^t$ up to a factor polynomial in $t$; with no frozen agent the corresponding term would instead tend to $\one\pi^\top\tilde{\mathbf f}_0$, as in Theorem~\ref{thm:case3}, so with a boundary the updating agents' initialisation leaves no trace. The second term is a partial sum of a convergent geometric series, whose tail decays at the same rate, and tends to
$Z\mathbf g=H\mathbf f_{\Frz}(y)+ZB_{\Alg}^{-1}\mathbf r_{\Alg}(y)$, using
$Z=(I-\AUU)^{-1}$ and $H=Z\AUF$ from Lemma~\ref{lem:substoch}. Differencing at
$y$ and $y'$ cancels the normalisers and gives \eqref{eq:case3-partial}. All
entries of $Z$ and $H$ are finite, so the limiting log-odds are bounded and
$q^*_v$ is an interior distribution.
\proofend

\subsection{Section~\ref{sec:case2}: fixed reference, aggregated challenger}
\label{app:caseII}

This appendix proves the results of Section~\ref{sec:case2}, then describes the
behaviour below the temperature threshold and gives three supporting results. The proofs use the following standard estimates.

The constant $\tfrac12$ below is the sharp Lipschitz constant of the softmax,
known to hold in every $\ell_p$ norm~\citep{nair2025}; the form stated here,
with the oscillation on the right, is the one the temperature thresholds in this paper require, and the proof is included for completeness.

\begin{lemma}[Softmax Lipschitz bound]
\label{lem:softmax}
Let $\sigma$ be the softmax on $\mathbb{R}^\Y$. For any $f,f'\in\mathbb{R}^\Y$,
\[
  \|\sigma(f)-\sigma(f')\|_1\;\leq\;\tfrac12\osc(f-f'),
\]
and the constant $\tfrac12$ cannot be improved.
\end{lemma}

\begin{proof}
Write
\[
  \mathrm d\sigma(f)[h]=\lim_{\epsilon\to0}\frac{\sigma(f+\epsilon h)-\sigma(f)}{\epsilon}
\]
for the directional derivative of $\sigma$ at $f$ along
$h\in\mathbb{R}^\Y$, a vector indexed by $\Y$ like $\sigma$ itself.
Differentiating $\sigma(f)_y=e^{f_y}/\sum_{y'}e^{f_{y'}}$ by the quotient rule
gives its $y$-th entry as $\sigma(f)_y(h_y-\E_{\sigma(f)}h)$: increasing $f$
along $h$ raises the mass at $y$ in proportion to how far $h_y$ exceeds its
average under $\sigma(f)$. With $p=\sigma(f)$,
\[
  \|\mathrm d\sigma(f)[h]\|_1=\E_p\bigl|h-\E_ph\bigr| .
\]
We use the following elementary bound. Let $X$ be a random variable taking
values in an interval $[a,b]$ and write $\mu=\E X$. The function
$x\mapsto|x-\mu|$ is convex, so on $[a,b]$ it lies below the chord joining its
values at the endpoints, and taking expectations,
\[
  \E|X-\mu|
  \;\leq\;\frac{b-\mu}{b-a}\,(\mu-a)+\frac{\mu-a}{b-a}\,(b-\mu)
  \;=\;\frac{2(\mu-a)(b-\mu)}{b-a}
  \;\leq\;\frac{b-a}{2} ,
\]
the last step because $(\mu-a)(b-\mu)$ is a concave quadratic in $\mu$
vanishing at $\mu=a$ and $\mu=b$, so it is largest at the midpoint
$\mu=\tfrac{a+b}{2}$, where it equals $\bigl(\tfrac{b-a}{2}\bigr)^2$. So
$\E|X-\E X|\leq\tfrac12(b-a)$ for any such $X$.

Apply this with $X=h$ under $p$, which takes values in
$[\min_yh_y,\max_yh_y]$, an interval of length $\osc(h)$. This gives
$\|\mathrm d\sigma(f)[h]\|_1\leq\tfrac12\osc(h)$ for every $f$ and $h$. The path $s\mapsto f'+s(f-f')$ runs from $f'$ to $f$ with constant velocity
$f-f'$, so
\[
  \sigma(f)-\sigma(f')=\int_0^1\mathrm d\sigma\bigl(f'+s(f-f')\bigr)[f-f']\,\mathrm ds ,
\]
and taking $\|\cdot\|_1$ inside the integral and applying the bound above with
$h=f-f'$ at each $s$ gives the inequality of the lemma.

It remains to show that the constant $\tfrac12$ cannot be improved. Take two
responses, $f'=(0,0)$ and
$f=(\epsilon,0)$, so that $\osc(f-f')=\epsilon$. Then $\sigma(f')$ is uniform
and $\sigma(f)=\bigl(\tfrac{e^\epsilon}{1+e^\epsilon},\tfrac{1}{1+e^\epsilon}\bigr)$,
whence
\[
  \|\sigma(f)-\sigma(f')\|_1
  =2\Bigl(\frac{e^\epsilon}{1+e^\epsilon}-\frac12\Bigr)
  =\tanh(\epsilon/2) ,
\]
and $\tanh(\epsilon/2)/\bigl(\tfrac12\epsilon\bigr)\to1$ as $\epsilon\to0$. So
no constant smaller than $\tfrac12$ works. The two inequalities in the
argument above are tight in the same limit: the distribution is uniform, so
$\mu$ is the midpoint of $[a,b]$, where the quadratic attains its maximum, and the mass lies entirely at the two endpoints, where $x\mapsto|x-\mu|$
meets its chord.
\end{proof}

\paragraph{Proof of Proposition~\ref{prop:case2-fp}.}

\emph{(a)} The map $\{q_v\}\mapsto\{q'_v\}$ given by \eqref{eq:case2} is
continuous, being the composition of the softmax with the linear map
$\mathbf q\mapsto(R_v(\cdot;\check q_v))_v$; the domain $\Delta(\Y)^n$ is compact
and convex, so Brouwer's fixed-point theorem \citep[Ch.~6]{border1985} applies.

\emph{(b)} Let $q^*,\tilde q^*$ be fixed points and write
$d_v=\|q^*_v-\tilde q^*_v\|_1$. Subtracting the fixed-point equations, the
anchor $\mu_v$ cancels, so the two beliefs at agent $v$ are softmaxes of
log-vectors differing only by the reward term, and
\[
  \osc_y\log\frac{q^*_v(y)}{\tilde q^*_v(y)}
  =\osc_y\Bigl(\tfrac{1}{\beta_v}\bigl[R_v(y;\check q^*_v)-R_v(y;\check{\tilde q}^*_v)\bigr]\Bigr)
  \leq\frac{2}{\beta_v}\sup_y\bigl|R_v(y;\check q^*_v)-R_v(y;\check{\tilde q}^*_v)\bigr| ,
\]
using $\osc(x)\leq2\sup|x|$. By Lemma~\ref{lem:signal-lip} and
$\sum_ua_{vu}=1$,
\[
  \bigl|R_v(y;\check q^*_v)-R_v(y;\check{\tilde q}^*_v)\bigr|
  \leq\tfrac12\bigl\|\check q^*_v-\check{\tilde q}^*_v\bigr\|_1
  \leq\tfrac12\sum_ua_{vu}d_u .
\]
Combining the two displays, that oscillation is at most
$\tfrac{1}{\beta_v}\sum_ua_{vu}d_u$. Write $q^*_v=\sigma(g)$ and $\tilde q^*_v=\sigma(\tilde g)$ for the two
log-vectors of which they are the softmaxes. Since a softmax changes its
argument only by an additive constant,
$\log\bigl(q^*_v(y)/\tilde q^*_v(y)\bigr)=g_y-\tilde g_y$ up to a constant, so
the oscillation just bounded is exactly $\osc(g-\tilde g)$.
Lemma~\ref{lem:softmax} applied to $g$ and $\tilde g$ therefore gives
\[
  d_v\;\leq\;\frac{1}{2\beta_v}\sum_ua_{vu}d_u ,
  \qquad\text{that is}\qquad
  \mathbf d\;\leq\;\tfrac12B^{-1}A\,\mathbf d
\]
entrywise, where $M=\tfrac12B^{-1}A$ has non-negative entries. A non-negative
matrix preserves the entrywise order, so applying $M$ to $\mathbf d\leq
M\mathbf d$ gives $M\mathbf d\leq M^2\mathbf d$, and hence by induction
$\mathbf d\leq M^k\mathbf d$ for every $k$. If $\rho(M)<1$ then $M^k\to0$, so
the right-hand side tends to zero and $\mathbf d\leq0$; as $\mathbf d\geq0$ by
definition, $\mathbf d=0$. The condition $\rho(M)<1$ is
$\rho(B^{-1}A)<2$.

Bounding each convex combination by its largest term gives a form in which no
feature of the graph appears: with $\Delta_\infty=\max_vd_v$, the display
above yields $\Delta_\infty\leq\Delta_\infty/(2\beta_{\min})$, that is
$\Delta_\infty\bigl(1-1/(2\beta_{\min})\bigr)\leq0$, and the bracket is
positive once $\beta_{\min}>\tfrac12$, so $\Delta_\infty\leq0$ and hence
$\Delta_\infty=0$.

\emph{(c)} Suppose first that an interior consensus fixed point
$q^*_v\equiv q^*$ exists. Its challenger is
$\check q^*_v=\sum_ua_{vu}q^*=q^*$ at every agent, since the rows of $A$ sum to
one. Interiority makes $q^*(y)>0$ for every $y$, so logarithms may be taken in
\eqref{eq:case2}, and the fixed-point equation reads
\[
  \log q^*(y)=\log\mu_v(y)+\tfrac{1}{\beta_v}R_v\bigl(y;q^*\bigr)-\Phi_v
\]
for every $v$, with $\Phi_v$ the $y$-independent log-normaliser. The left-hand side does not depend on $v$, so neither does the right-hand
side; exponentiating returns the stated proportionality. Differencing at two responses removes $\Phi_v$ and gives the log-odds form, whose left-hand side $\log\bigl(q^*(y)/q^*(y')\bigr)$ is kept; conversely, the log-odds form at every pair fixes $\log\mu_v+R_v(\cdot\,;q^*)/\beta_v$ up to a constant as $\log q^*$, which is the proportionality, so the two forms are equivalent. Conversely, if some interior $q^*$ satisfies it, then the
right-hand side is the same function of $y$ at every agent, and normalising it
returns $q^*$ itself, so the profile $q^*_v\equiv q^*$ is a fixed point. This
is the only constraint the equations impose, which is why the anchors and kernels are constrained jointly.

\emph{(d)} Let $\mathbf q^1,\mathbf q^2$ be fixed points and write
$\Delta q_v=q^1_v-q^2_v$, so that each $\Delta q_v$ sums to zero. Taking
logarithms in \eqref{eq:case2} at each fixed point, subtracting and multiplying
by $\beta_v$, the anchors cancel and
\[
  \beta_v\bigl(\log q^1_v-\log q^2_v\bigr)
  =P\check q^1_v-P\check q^2_v+c_v\one
  =P\!\!\sum_u a_{vu}\Delta q_u+c_v\one ,
\]
with $c_v$ a constant. Pair each side with $\Delta q_v$; the term $c_v\one$ drops
out because $\Delta q_v$ sums to zero. Multiplying by $\omega_v$ and summing over the
agents,
\begin{equation}\label{eq:case2-uniq}
  \sum_v\omega_v\beta_v\bigl\langle\log q^1_v-\log q^2_v,\,\Delta q_v\bigr\rangle
  \;=\;\sum_{v,u}\omega_va_{vu}\bigl\langle P\Delta q_u,\,\Delta q_v\bigr\rangle .
\end{equation}
Each inner product on the left is of the form $\langle\log a-\log b,\,a-b\rangle$
with $a,b$ positive, which is non-negative because $\log$ is increasing and zero
only when $a=b$. Its weight $\omega_v\beta_v$ is positive, so the left-hand side is
strictly positive unless $\mathbf q^1=\mathbf q^2$.

The right-hand side vanishes. A preference kernel satisfies
$P+P^\top=\one\one^\top$, so for any $d,d'$ summing to zero,
\[
  \langle Pd',d\rangle+\langle Pd,d'\rangle
  =d^\top(P+P^\top)d'
  =\Bigl(\sum_yd(y)\Bigr)\Bigl(\sum_yd'(y)\Bigr)=0 .
\]
Taking $d=d'=\Delta q_v$ kills every diagonal term. The off-diagonal terms pair
by reversibility: since $\omega_ua_{uv}=\omega_va_{vu}$, the terms for $(v,u)$ and $(u,v)$
add to
$\omega_va_{vu}\bigl(\langle P\Delta q_u,\Delta q_v\rangle+\langle P\Delta q_v,\Delta q_u\rangle\bigr)=0$.
Hence the right-hand side of \eqref{eq:case2-uniq} is zero, the left is
therefore zero, and $\mathbf q^1=\mathbf q^2$. No condition on the temperatures
enters.
\proofend

\paragraph{Proof of Theorem~\ref{thm:case2-lottery}.}

Throughout, $S_v=P_v-\tfrac12\one\one^\top$ is the skew part of $P_v$. The
following bound is used in Steps~2 and~4: for any vector $d$ with
$\sum_yd(y)=0$,
\begin{equation}\label{eq:skew-bound}
  \max_y\bigl|(P_vd)_y\bigr|=\max_y\bigl|(S_vd)_y\bigr|\leq\tfrac12\|d\|_1 ,
\end{equation}
the equality because $\one\one^\top d=0$ and the inequality because every entry
of $S_v$ lies in $[-\tfrac12,\tfrac12]$.

\emph{Step 1: each belief nearly maximises its own quality signal.} Fix $v$
and hold its challenger at $\check q^*_v$. An agent at a fixed point does not
maximise $\langle p,P_v\check q^*_v\rangle$ outright, since the penalty holds
it near its anchor; Step~1 bounds the shortfall by $\beta_vM_v$. By
\eqref{eq:general} the belief $q^*_v$ maximises
\[
  G_v(p)=\bigl\langle p,\;P_v\check q^*_v\bigr\rangle-\beta_v\KL(p\,\|\,\mu_v)
\]
over $\Delta(\Y)$: the first term is linear in $p$, the second strictly
convex, so the stationary point is the unique maximiser, and that stationary
point is the Gibbs policy $\mu_v\exp(P_v\check q^*_v/\beta_v)$ normalised, which
is $q^*_v$ by the fixed-point equation. Hence $G_v(q^*_v)\geq G_v(p)$ for
every $p\in\Delta(\Y)$, that is
\[
  \bigl\langle p,P_v\check q^*_v\bigr\rangle
  -\bigl\langle q^*_v,P_v\check q^*_v\bigr\rangle
  \;\leq\;\beta_v\bigl[\KL(p\,\|\,\mu_v)-\KL(q^*_v\,\|\,\mu_v)\bigr]
  \;\leq\;\beta_v\KL(p\,\|\,\mu_v),
\]
the last step because $\KL\geq0$. Now, since $p(y)\leq1$ and
$\mu_v(y)\geq\min_{y'}\mu_v(y')$ at every $y$,
\[
  \KL(p\,\|\,\mu_v)=\sum_yp(y)\log\frac{p(y)}{\mu_v(y)}
  \;\leq\;\sum_yp(y)\log\frac{1}{\mu_v(y)}\;\leq\;M_v ,
\]
with equality when $p$ is the point mass at the least likely response. Hence
\begin{equation}\label{eq:case2-abr}
  \bigl\langle p,P_v\check q^*_v\bigr\rangle
  \;\leq\;\bigl\langle q^*_v,P_v\check q^*_v\bigr\rangle+\beta_vM_v .
\end{equation}

\emph{Step 2: the self-term is close to one half.} Because
$P_v+P_v^\top=\one\one^\top$, for any distributions $q,q'$,
\[
  \bigl\langle q,P_vq'\bigr\rangle+\bigl\langle q',P_vq\bigr\rangle
  =q^\top\bigl(P_v+P_v^\top\bigr)q'=1 ,
\]
so $\langle q,P_vq'\rangle=\tfrac12+\langle q,S_vq'\rangle$. Taking
$q=q'=q^*_v$ and using $\langle q,S_vq\rangle=0$ for skew $S_v$ gives
$\langle q^*_v,P_vq^*_v\rangle=\tfrac12$, and therefore
\[
  \bigl\langle q^*_v,P_v\check q^*_v\bigr\rangle
  =\tfrac12+\bigl\langle q^*_v,S_v(\check q^*_v-q^*_v)\bigr\rangle
  \;\leq\;\tfrac12+\tfrac12\bigl\|\check q^*_v-q^*_v\bigr\|_1 ,
\]
by \eqref{eq:skew-bound} and $\|q^*_v\|_1=1$.

\emph{Step 3: the challengers are close to the collective policy.} Since
$\check q^*_v-\bar q_w=\sum_ua_{vu}(q^*_u-\bar q_w)$ is a convex combination,
\[
  \bigl\|\check q^*_v-\bar q_w\bigr\|_1\leq\max_u\bigl\|q^*_u-\bar q_w\bigr\|_1=\mathcal D_w ,
\]
and by the triangle inequality
$\|\check q^*_v-q^*_v\|_1\leq\|\check q^*_v-\bar q_w\|_1+\|\bar q_w-q^*_v\|_1\leq2\mathcal D_w$.

\emph{Step 4: averaging against $w$.} Multiply \eqref{eq:case2-abr} by $w_v$, sum over $v$, and insert Step~2:
\begin{equation}\label{eq:case2-avg}
  \sum_vw_v\bigl\langle p,P_v\check q^*_v\bigr\rangle
  \;\leq\;\tfrac12+\tfrac12\sum_vw_v\bigl\|\check q^*_v-q^*_v\bigr\|_1
  +\sum_vw_v\beta_vM_v
  \;\leq\;\tfrac12+\mathcal D_w+\sum_vw_v\beta_vM_v ,
\end{equation}
using Step~3 for the middle term. For the left-hand side, write $\check q^*_v=\bar q_w+(\check q^*_v-\bar q_w)$ and use $\bar P_w=\sum_vw_vP_v$:
\[
  \sum_vw_v\bigl\langle p,P_v\check q^*_v\bigr\rangle
  =\bigl\langle p,\bar P_w\bar q_w\bigr\rangle
  +\sum_vw_v\bigl\langle p,P_v(\check q^*_v-\bar q_w)\bigr\rangle .
\]
Each $\check q^*_v-\bar q_w$ sums to zero, so by \eqref{eq:skew-bound} and
Step~3 the second sum is at least
$-\tfrac12\sum_vw_v\|\check q^*_v-\bar q_w\|_1\geq-\tfrac12\mathcal D_w$.
Substituting into \eqref{eq:case2-avg},
\[
  \bigl\langle p,\bar P_w\bar q_w\bigr\rangle
  \;\leq\;\tfrac12+\tfrac32\mathcal D_w+\sum_vw_v\beta_vM_v
\]
for every $p\in\Delta(\Y)$. The kernel $\bar P_w$ is an average of preference kernels, so
$\bar P_w(y,y')+\bar P_w(y',y)=1$, and therefore
$\langle q,\bar P_w\bar q_w\rangle=\E_{y'\sim q,\,y\sim\bar q_w}[\bar P_w(y',y)]
=1-\E_{y\sim\bar q_w,\,y'\sim q}[\bar P_w(y,y')]$ for every $q\in\Delta(\Y)$. Taking
$p=q$ gives \eqref{eq:case2-lottery}.

\emph{The homogeneous case.} If $P_v\equiv P$, $\mu_v\equiv\mu$ and
$\beta_v\equiv\beta$, then by Proposition~\ref{prop:case2-fp}(c) the consensus
profile $q^*_v\equiv q^*$ is a fixed point, $\mathcal D_w=0$ and
$\bar q_w=q^*$ for every $w$, so $\varepsilon\leq\beta M$. As $\beta\to0$ the right-hand side vanishes and
$\bar q_w$ converges to the set of maximal lotteries of $P$, which is
non-empty and compact.
\proofend

\paragraph{Proof of Corollary~\ref{cor:case2-homog}.}

Throughout $P_v\equiv P$, $\mu_v\equiv\mu$, $\beta_v\equiv\beta$, and we write
$R(y;q)=\sum_{y'}q(y')P(y,y')$.

\emph{(a) Consensus fixed points are exactly the regularised Nash policies.} If
$q_v\equiv q$ then $\check q_v=\sum_ua_{vu}q=q$ for every $v$, since the weights
sum to one, so the fixed-point equation \eqref{eq:case2} reads
$q\propto\mu\,e^{R(\cdot;q)/\beta}$ at every agent simultaneously, which is the
single-agent equation \eqref{eq:nash}; the graph has dropped out. Conversely any
solution of the single-agent equation gives a consensus fixed point. For
uniqueness, let $p$ and $p'$ both solve it. Taking logarithms,
\[
  \beta\bigl(\log p-\log\mu\bigr)=R(\cdot;p)-c\one,
  \qquad
  \beta\bigl(\log p'-\log\mu\bigr)=R(\cdot;p')-c'\one
\]
for constants $c,c'$. Subtracting and pairing with $p-p'$,
\[
  \beta\bigl\langle \log p-\log p',\,p-p'\bigr\rangle
  =\bigl\langle R(\cdot;p)-R(\cdot;p'),\,p-p'\bigr\rangle
  -(c-c')\textstyle\sum_y(p-p')(y).
\]
The last term vanishes because both are probability vectors. Writing $d=p-p'$
and $P=\tfrac12\one\one^\top+S$ with $S$ antisymmetric, as $P(y,y')+P(y',y)=1$ requires, the first term on the right is
$d^\top Pd=\tfrac12\bigl(\sum_yd(y)\bigr)^2+d^\top Sd=0$, the two terms
vanishing because $\sum_yd(y)=0$ and because $d^\top Sd=0$ for antisymmetric
$S$. Hence $\langle\log p-\log p',p-p'\rangle=0$. But $x\mapsto x\log x$ is
strictly convex, so $\langle\log p-\log p',p-p'\rangle>0$ whenever $p\neq p'$.
Therefore $p=p'$. Existence for every $\beta>0$ is
Proposition~\ref{prop:case2-fp}(a) applied to the single agent, or equivalently
Brouwer's fixed-point theorem~\citep[Ch.~6]{border1985} applied to
$q\mapsto\mu e^{R(\cdot;q)/\beta}/Z$, which maps the
compact convex simplex into itself continuously.

\emph{(b) The approximation bound.} Fix the fixed point $p^*$ and consider
$\Phi(q)=\E_q[R(\cdot;p^*)]-\beta\KL(q\|\mu)$ as a function of $q$ alone, the
challenger being held at $p^*$. The first term is linear in $q$ and
$\KL(\cdot\,\|\mu)$ is strictly convex, so $\Phi$ is strictly concave and its
stationary point is its unique maximiser. That stationary point is
$q\propto\mu e^{R(\cdot;p^*)/\beta}$, which by~(a) is $p^*$ itself. So
$\Phi(p^*)\geq\Phi(q)$ for every $q\in\Delta(\Y)$, and rearranging that
inequality,
\[
  \E_q\bigl[R(\cdot;p^*)\bigr]-\E_{p^*}\bigl[R(\cdot;p^*)\bigr]
  \;\leq\;\beta\bigl[\KL(q\|\mu)-\KL(p^*\|\mu)\bigr]
  \;\leq\;\beta\,\KL(q\|\mu),
\]
the second step because $\KL(p^*\|\mu)\geq0$. Skew-symmetry gives
$\E_{p^*}[R(\cdot;p^*)]=\tfrac12$, and
$\KL(q\|\mu)=\sum_yq(y)\log\bigl(q(y)/\mu(y)\bigr)\leq\log(1/\mu_{\min})$ since
$q(y)/\mu(y)\leq1/\mu_{\min}$. Therefore
$\E_q[R(\cdot;p^*)]\leq\tfrac12+\beta\log(1/\mu_{\min})$ for every $q$, which is
the definition of an $\varepsilon$-maximal lottery with
$\varepsilon=\beta\log(1/\mu_{\min})$.

\emph{(c) Above the threshold.} With $\beta>\tfrac12$,
Proposition~\ref{prop:case2-geom} makes the update map a contraction of
modulus $\gamma=1/(2\beta)<1$ in
$d(\mathbf q,\mathbf q')=\max_v\|q_v-q'_v\|_1$. A contraction on a complete
metric space has exactly one fixed point and every trajectory converges to it at
the rate $\gamma^t$; by (a) that fixed point is the consensus profile at the
regularised Nash policy. Below the threshold the contraction argument is
unavailable, and the Supplementary Material shows convergence can in
fact fail.
\proofend

\paragraph{Proof of Proposition~\ref{prop:case2-geom}.}

The estimate is the one used in Proposition~\ref{prop:case2-fp}(b), here applied to any two profiles. Subtracting the
updates, the reference $\mu_v$ cancels; Lemma~\ref{lem:softmax} bounds
$\|T_v(\mathbf q)-T_v(\mathbf q')\|_1$, where $T_v(\mathbf q)$ denotes the
right-hand side of \eqref{eq:case2} as a function of the profile, by half the
oscillation of the reward difference, and Lemma~\ref{lem:signal-lip} bounds
that oscillation as before, giving the componentwise bound
\[
  e_v(T\mathbf q,T\mathbf q')\;\leq\;\frac{1}{2\beta_v}\sum_ua_{vu}\,e_u(\mathbf q,\mathbf q'),
  \qquad e_v(\mathbf q,\mathbf q')=\|q_v-q'_v\|_1 ,
\]
that is
$\mathbf e(T\mathbf q,T\mathbf q')\leq\tfrac12B^{-1}A\,\mathbf e(\mathbf q,\mathbf q')$
entrywise. Since $\rho(\tfrac12B^{-1}A)<1$ the geometric series
$(I-\tfrac12B^{-1}A)^{-1}=\sum_{s\geq0}(\tfrac12B^{-1}A)^s$ converges with non-negative
entries, so $\mathbf z=(I-\tfrac12B^{-1}A)^{-1}\one$ satisfies $z_v\geq1$ and
$\tfrac12B^{-1}A\mathbf z=\mathbf z-\one$. Writing $d_{\mathbf z}=\max_ve_v/z_v$ and
using $\mathbf e\leq d_{\mathbf z}\mathbf z$ entrywise,
\[
  \mathbf e(T\mathbf q,T\mathbf q')\;\leq\;d_{\mathbf z}\,\tfrac12B^{-1}A\mathbf z
  \;=\;d_{\mathbf z}(\mathbf z-\one),
\]
so
$d_{\mathbf z}(T\mathbf q,T\mathbf q')\leq\gamma\,d_{\mathbf z}(\mathbf q,\mathbf q')$
with $\gamma=\max_v\{1-1/z_v\}=1-1/\|\mathbf z\|_\infty$, as stated. Each
$z_v$ is finite, since the series converges, so $1-1/z_v<1$ for every $v$; as $V$ is finite the maximum is
attained and $\gamma<1$.

The hypotheses of Banach's fixed point theorem are therefore met. The space is
$\Delta(\Y)^n$ with the metric $d_{\mathbf z}$, which is a metric because the
$z_v$ are finite and positive and $\|\cdot\|_1$ is one; it is complete,
being a closed and bounded subset of a finite-dimensional space with a metric
equivalent to the Euclidean one, again because the $z_v$ lie in a finite
positive range. The map $T$ sends $\Delta(\Y)^n$ into itself, each $T_v$ being
a normalised distribution, and the display above makes it a contraction with
modulus $\gamma<1$. Banach's contraction mapping theorem therefore gives a unique fixed point and
convergence to it from every initialisation at the rate $\gamma^t$.

Finally, suppose $\beta_v>\tfrac12$ at every $v$. Then
$\tfrac12B^{-1}A\one\leq\max_v\tfrac{1}{2\beta_v}\one$, so $\mathbf z=\one$
may be taken in place of the weighted vector above, and $d_{\mathbf z}$
reduces to the unweighted metric $\max_v\|q_v-q'_v\|_1$ with modulus
$1/(2\beta_{\min})$.
\proofend

\paragraph{Proof of Proposition~\ref{prop:case2-confine}.}

Fix $t\geq0$, $v$ and two responses $y,y'$. From \eqref{eq:case2},
\[
  \frac{q_{v,t+1}(y)}{q_{v,t+1}(y')}
  =\frac{\mu_v(y)}{\mu_v(y')}
   \exp\Bigl(\tfrac{1}{\beta_v}\bigl[R_v(y;\check q_{v,t})-R_v(y';\check q_{v,t})\bigr]\Bigr)
  \;\geq\;\frac{\mu_v(y)}{\mu_v(y')}\,e^{-1/\beta_v},
\]
because $R_v(\cdot;q)=\E_{y'\sim q}[P_v(\cdot,y')]$ takes values in $[0,1]$, so
the bracket is at least $-1$. Interchanging $y$ and $y'$ in that inequality and
taking reciprocals gives the upper bound
\[
  \frac{q_{v,t+1}(y')}{q_{v,t+1}(y)}\;\leq\;\frac{\mu_v(y')}{\mu_v(y)}\,e^{1/\beta_v} .
\]
Summing it over $y'$ and using $\sum_{y'}q_{v,t+1}(y')=1$ and
$\sum_{y'}\mu_v(y')=1$,
\[
  \frac{1}{q_{v,t+1}(y)}=\sum_{y'}\frac{q_{v,t+1}(y')}{q_{v,t+1}(y)}
  \;\leq\;\sum_{y'}\frac{\mu_v(y')}{\mu_v(y)}\,e^{1/\beta_v}
  \;=\;\frac{e^{1/\beta_v}}{\mu_v(y)},
\]
which on rearranging is \eqref{eq:case2-confine}. The bound holds for every
$t\geq1$, and the set of profiles satisfying it is closed, bounded and
contained in the interior of the simplex, hence a compact subset of it that
does not depend on $t$.
\proofend

\paragraph{Stability of the fixed points of \eqref{eq:case2}.}
\label{app:case2-stability}
Write $T$ for the map taking a profile to the right-hand side of
\eqref{eq:case2}, so that the fixed points are the solutions of
$T(\mathbf q)=\mathbf q$. We compute its derivative and read off when a fixed
point is locally asymptotically stable, as
Section~\ref{sec:case2-convergence} asserts.

\emph{The derivative.} Agent $v$'s component is
$T_v(\mathbf q)=\mathrm{softmax}\bigl(\log\mu_v+P_v\check q_v/\beta_v\bigr)$
with $\check q_v=\sum_ua_{vu}q_u$, a composition of three maps: the linear
map $\mathbf q\mapsto\check q_v$, with derivative $a_{vu}I$ in the $u$th
block; multiplication by $P_v/\beta_v$; and the softmax, whose derivative at
$q^*_v$ is $J_v=\mathrm{diag}(q^*_v)-q^*_vq^{*\top}_v$. By the chain rule the
Jacobian of $T$ at $\mathbf q^*$ has blocks
\[
  \bigl[\mathrm dT\bigr]_{vu}=\frac{a_{vu}}{\beta_v}\,J_vP_v .
\]
The log-normaliser
contributes a multiple of $\one$, which $J_v$ annihilates, so it does not
appear. The fixed point is locally asymptotically stable when
$\rho(\mathrm dT)<1$ and unstable when $\rho(\mathrm dT)>1$, the update being
analytic.

\emph{The two regimes.} The blocks differ only through $J_v$, so the
behaviour at low temperature is decided by how fast $J_v$ vanishes relative
to $\beta_v$. Consider first the case in which each belief concentrates on a single
response, as happens when the game the agents play in the limit
$\beta_v\to0$ has a strict pure equilibrium. The off-response mass is then
exponentially small in $1/\beta_v$, so $J_v$ vanishes faster than $\beta_v$. Let $y_v$ be the response agent $v$ plays at that
equilibrium, with margin $\delta>0$. By the argument of
Proposition~\ref{prop:local-strict}, $1-q^*_v(y_v)\leq(|\Y|-1)Me^{-\delta/2\beta_v}$,
where $M$ bounds the spread of the anchors. The contraction estimate in the proof of Proposition~\ref{prop:local-strict}
then applies at $\mathbf q^*$: for a perturbation $(h_u)$ with $\|h_u\|_1\leq r$
for every $u$, it gives $\|\mathrm dT_v\,h\|_1\leq L(\beta_v)\,r$, using that the
rows of $A$ sum to one. The spectral radius is at most this induced norm, so
\[
  \rho\bigl(\mathrm dT\bigr)\;\leq\;\max_vL(\beta_v)
  \;=\;\max_v\frac{4(|\Y|-1)M}{\beta_v}\,e^{-\delta/2\beta_v}.
\] The right-hand side tends to zero as
the temperatures fall, since the exponential dominates $1/\beta_v$, so the
fixed point is locally asymptotically stable at every low enough temperature.

Consider now the case in which some belief remains spread over several
responses in the limit. A mixed maximal lottery requires this, and a
lottery supported on two responses places half its mass on each, so it arises
only for a kernel degenerate on that pair; a genuinely mixed limit therefore
has at least three responses in its support. Suppose then that $q^*_v$
remains in the interior as $\beta_v\to0$, with $\min_yq^*_v(y)\geq c>0$ along
the sequence. Then $J_v$ is bounded below on the
sum-zero subspace: for $d$ with $\sum_yd(y)=0$,
\[
  \langle d,J_vd\rangle=\mathrm{Var}_{q^*_v}(d)
  =\sum_yq^*_v(y)\Bigl(d(y)-\textstyle\sum_{y'}q^*_v(y')d(y')\Bigr)^2
  \;\geq\;\frac{c}{|\Y|}\,\|d\|_2^2 ,
\]
the inequality because each $q^*_v(y)\geq c$ and the variance of $d$ under any
distribution is at least $\|d\|_2^2/|\Y|$ when $d$ sums to zero. So $J_v$,
restricted to that subspace, has smallest eigenvalue at least $c/|\Y|$, and
in particular does not degenerate as $\beta_v\to0$. Unless $J_vP_v$ vanishes
on the sum-zero subspace, which happens only for a degenerate kernel,
$\rho(J_vP_v)$ is therefore bounded below along the sequence and
$\rho(\mathrm dT)\geq\rho(J_vP_v)/\beta_v\to\infty$. The fixed point is then
unstable once the temperature is small enough.

\proofend

\subsection{Section~\ref{sec:case4}: both roles aggregated}
\label{app:caseIII}

\paragraph{Proof of Proposition~\ref{prop:case4-fp}.}

\emph{(a)} This follows from~(b), proved below: every atom profile is a fixed
point. Brouwer's theorem does not apply directly here, because the reference
$\bar q_v\propto\prod_uq_u^{a_{vu}}$ is not defined on all of $\Delta(\Y)^n$: it
vanishes at every response when the agents $v$ observes have disjoint
supports.

\emph{(c)} A consensus profile has a common support by construction, every
agent holding $q^*$. Fix $y,y'\in S$, so that both log-odds are finite. The
log-odds form of the fixed-point equation is
\[
  \log\frac{q^*_v(y)}{q^*_v(y')}=\sum_ua_{vu}\log\frac{q^*_u(y)}{q^*_u(y')}
  +\frac{1}{\beta_v}\bigl[R_v(y;\check q^*_v)-R_v(y';\check q^*_v)\bigr].
\]
If every agent holds $q^*$, then each agent's challenger is the arithmetic
mean $\check q^*_v=\sum_ua_{vu}q^*=q^*$, the weights summing to one. The same
weights make the first term on the right equal to the left-hand side, so the
two cancel, leaving $R_v(y;q^*)=R_v(y';q^*)$. As $y,y'\in S$ were arbitrary this
is \eqref{eq:unanimity}.

Conversely, suppose each $R_v(\cdot\,;q^*)$ is constant on $S$. On $S$ the
update multiplies $q^*$ by a factor independent of the response, which
normalisation removes, so it reproduces $q^*$ there; off $S$ the reference
$\bar q^*_v$ vanishes, so the update assigns zero mass whatever the signal.
Hence $q^*$ is a fixed point.

\emph{(b)} Take $q^*=\delta_y$, so $S=\{y\}$. The only pair in
\eqref{eq:unanimity} is $y=y$, which holds for every kernel, temperature and
attention matrix, so (c) makes the profile a fixed point.
\proofend

\paragraph{Proof of Proposition~\ref{prop:case4-spread}.}

\emph{(a)} Since $R_v$ takes values in $[0,1]$ and $\beta_v>0$, the factor
$\exp(R_v(y;\check q^*_v)/\beta_v)$ lies in $[1,e^{1/\beta_v}]$ and never
vanishes, so $q^*_v(y)=0$ if and only if the reference $\bar q^*_v(y)=0$.
That reference is a weighted geometric mean, so it vanishes at $y$ exactly
when $q^*_u(y)=0$ for some $u$ with $a_{vu}>0$. Hence an agent assigns $y$ no mass
exactly when some agent it observes does. Under Assumption~\ref{ass:graph}
there is $m$ with $A^m>0$, so following that implication along a path of length
$m$ shows that if any agent assigns $y$ no mass then every agent does. The
supports therefore coincide.

\emph{(b)} Fix $y\in S$. Every agent has $q^*_v(y)>0$, so the logarithms are
finite and the log form of \eqref{eq:case4} reads
$(I-A)\mathbf f^*(y)=B^{-1}\mathbf R^*(y)-\boldsymbol\Phi$, with
$\boldsymbol\Phi$ the vector of $y$-independent normalisers. Applying
$\pi^\top$ annihilates the left-hand side, since $\pi^\top A=\pi^\top$, and
gives $\sum_v(\pi_v/\beta_v)R^*_v(y)=\pi^\top\boldsymbol\Phi$, a constant
independent of $y$: this is \eqref{eq:case4-solv}. For $y\notin S$ the
derivation does not apply, since $f^*_v(y)=-\infty$ for every $v$; there the
fixed-point relation reduces to $\bar q^*_v(y)=0$, which holds whatever the
value of $R^*_v(y)$.

\emph{(c)} Fix $y\in S$. By~(b) the right-hand side of the log form satisfies
the solvability condition of Theorem~\ref{thm:fredholm}, so that theorem gives
$\mathbf f^*(y)=D\bigl[B^{-1}\mathbf R^*(y)-\boldsymbol\Phi\bigr]+c(y)\one$ for
some scalar $c(y)$.
Since $\pi^\top D=0$ we get $\sum_u\pi_uf^*_u(y)=c(y)$, and subtracting,
$f^*_v(y)-\sum_u\pi_uf^*_u(y)=[DB^{-1}\mathbf R^*(y)]_v-[D\boldsymbol\Phi]_v$,
the second term independent of $y$; this is \eqref{eq:case4-spread}.
Exponentiating gives \eqref{eq:case4-spread-gibbs} on $S$, and off $S$ both
sides vanish, since $p^*(y)\propto\prod_uq^*_u(y)^{\pi_u}=0$ there.
\proofend

\paragraph{The disagreement measure.}
The disagreement $\mathcal D_t$ of Theorem~\ref{thm:case4-directed} is
measured in the unnormalised log-beliefs because \eqref{eq:logspace-vec} is
affine in them, whereas
renormalising each round destroys that structure. As a consequence, $\mathcal D_t$ sees the per-agent shifts $c_{v,t}$ of
Remark~\ref{rem:normalisation}, which do not cancel in an oscillation over
agents. So $\mathcal D_t=0$ implies consensus but not conversely.

The results need the shift-free quantity
$\osc_y\bigl(\tilde f_{u,t}(y)-\tilde f_{v,t}(y)\bigr)$. There the shifts
contribute the single constant $c_{u,t}-c_{v,t}$, the same at every $y$, and an
oscillation is unchanged by adding a constant, so the quantity depends only on
the normalised beliefs; by Lemma~\ref{lem:softmax} it bounds
$\|q_{u,t}-q_{v,t}\|_1$. The two are linked by
\begin{equation}\label{eq:D-controls}
  \osc_y\bigl(\tilde f_{u,t}(y)-\tilde f_{v,t}(y)\bigr)\;\leq\;2\,\mathcal D_t
  \qquad\text{for every pair }u,v ,
\end{equation}
since each of $\max_y$ and $-\min_y$ of the difference is at most
$\mathcal D_t$.

\paragraph{Proof of Theorem~\ref{thm:case4-directed}.}

Fix $t$ and abbreviate $f_v=\tilde f_{v,t}$ for the unnormalised log-beliefs
at that round, $f_{v,\mathrm{new}}=\tilde f_{v,t+1}$ for those at the next,
and $\mathcal D=\mathcal D_t$. By \eqref{eq:logspace-vec},
$f_{v,\mathrm{new}}(y)=\sum_ua_{vu}f_u(y)+R_v(y;\check q_v)/\beta_v$, with no
normalising term to account for.

\emph{The averaging term.} At each fixed $y$, Lemma~\ref{lem:dobrushin} applied
to the vector $(f_u(y))_u$ gives
$\osc_v\bigl(\sum_ua_{vu}f_u(y)\bigr)\leq\tau_1(A)\osc_u(f_u(y))\leq\tau_1(A)\mathcal D$.

\emph{The signal term.} For any two agents $u,v$ and any $y$, write
\[
  \frac{R_v(y;\check q_v)}{\beta_v}-\frac{R_u(y;\check q_u)}{\beta_u}
  =\frac{R_v(y;\check q_v)-R_u(y;\check q_u)}{\beta_v}
  +R_u(y;\check q_u)\Bigl(\frac1{\beta_v}-\frac1{\beta_u}\Bigr).
\]
The second term is at most $\Delta_\beta$ in absolute value, since
$R_u(y;\cdot)\in[0,1]$; it is the contribution of unequal temperatures and does
not vanish when the agents share a kernel. The first is at most
$1/\beta_{\min}\bigl[|R_v(y;\check q_v)-R_v(y;\check q_u)|
+|R_v(y;\check q_u)-R_u(y;\check q_u)|\bigr]$, the second bracket being at most
$\Delta_P$ by definition. For the first, $R_v$ is linear in its second argument, so it is at most $\tfrac12\|\check q_v-\check q_u\|_1$ by
Lemma~\ref{lem:signal-lip}, and in this case the challengers are themselves averages, which contracts their spread before the kernel acts on
it:
\begin{equation}\label{eq:challenger-contraction}
  \|\check q_v-\check q_u\|_1\;\leq\;\tau_1(A)\max_{u',v'}\|q_{v'}-q_{u'}\|_1 .
\end{equation}
To see \eqref{eq:challenger-contraction}, note that the row difference
$a_{v\cdot}-a_{u\cdot}$ has total mass zero, so it can be written as
$s(\mathbf p^+-\mathbf p^-)$ with $\mathbf p^\pm$ probability vectors and
$s=\tfrac12\|a_{v\cdot}-a_{u\cdot}\|_1\leq\tau_1(A)$ by
Definition~\ref{def:resolvent}. Then
\[
  \check q_v-\check q_u=\sum_w(a_{vw}-a_{uw})q_w
  =s\Bigl(\sum_wp^+_wq_w-\sum_wp^-_wq_w\Bigr),
\]
a multiple $s$ of the difference of two convex combinations of the $q_w$;
such a difference has $\ell^1$ norm at most $\max_{u',v'}\|q_{v'}-q_{u'}\|_1$,
which gives \eqref{eq:challenger-contraction}. Now for every agent pair and
every $y$ we have $|f_{v'}(y)-f_{u'}(y)|\leq\mathcal D$, because both lie in the
range whose width is $\osc_v(f_v(y))$; hence
$\osc_y\bigl(f_{v'}-f_{u'}\bigr)\leq2\mathcal D$, and Lemma~\ref{lem:softmax}, which carries a factor $\tfrac12$, gives
$\|q_{v'}-q_{u'}\|_1\leq\mathcal D$. Collecting, the first term is at most $\bigl(\tfrac12\tau_1(A)\mathcal D+\Delta_P\bigr)/\beta_{\min}$, the $\tfrac12$ coming from Lemma~\ref{lem:signal-lip}, so the signal term contributes at most $\bigl(\tfrac12\tau_1(A)\mathcal D+\Delta_P\bigr)/\beta_{\min}+\Delta_\beta$ to $\osc_v$, an oscillation over the agents being the largest difference between two of them.

Adding the two contributions gives \eqref{eq:case4-cond}, and iterating that
affine recursion yields the displayed bound.

If the agents share both a kernel and a temperature then $\Delta_P=0$ and
$\Delta_\beta=0$, so the recursion reads $\mathcal D_{t+1}\leq\gamma\mathcal D_t$
and $\mathcal D_t\leq\gamma^t\mathcal D_0\to0$ under
\eqref{eq:case4-threshold}. A fixed point has
$\mathcal D_{t+1}=\mathcal D_t=\mathcal D$, so $\mathcal D\leq\gamma\mathcal D$
with $\gamma<1$ forces $\mathcal D=0$: the profile is a consensus one, which
Proposition~\ref{prop:case4-fp}(c) identifies as a maximal lottery of the
shared kernel. A shared kernel alone leaves $\Delta_\beta$ in the recursion,
and the disagreement then contracts only to within
$\Delta_\beta/(1-\gamma)$.
\proofend

\paragraph{Proof of Theorem~\ref{thm:cesaro}.}

\emph{Step 1 (the aggregate is exponential weights).} Applying $\pi^\top$ to
\eqref{eq:logspace-vec} and using $\pi^\top A=\pi^\top$,
\[
  F_{t+1}=F_t+\sum_v\frac{\pi_v}{\beta_v}P_v\check q_{v,t}
  =F_t+\eta\,u_t,
  \qquad
  u_t=\frac1\eta\sum_v\frac{\pi_v}{\beta_v}P_v\check q_{v,t}\in[0,1]^\Y ,
\]
the range following because each $(P_v\check q_{v,t})(y)\in[0,1]$ and the
weights $(\pi_v/\beta_v)/\eta$ form a probability vector. Hence
$p_t=\mathrm{softmax}(F_t)$ is the exponential-weights iterate with learning
rate $\eta$ on the gain sequence $(u_t)$.

\emph{Step 2 (regret).} Let $\Phi_t=\log\sum_ye^{F_t(y)}$, so that
$p_t=e^{F_t-\Phi_t}$. Since $u_t(y)\in[0,1]$, Hoeffding's lemma
\citep{hoeffding1963} gives
\[
  \Phi_{t+1}-\Phi_t=\log\sum_yp_t(y)\,e^{\eta u_t(y)}
  \;\leq\;\eta\langle p_t,u_t\rangle+\frac{\eta^2}{8} ,
\]
and summing over $t<T$,
$\Phi_T\leq\Phi_0+\eta\sum_{t<T}\langle p_t,u_t\rangle+\eta^2T/8$. Fix
$q\in\Delta(\Y)$. Then
$\langle q,F_T\rangle=\langle q,F_0\rangle+\eta\sum_{t<T}\langle q,u_t\rangle$;
$\langle q,F_T\rangle\leq\Phi_T$, since $F_T(y)\leq\Phi_T$ for every $y$; and
$\Phi_0-\langle q,F_0\rangle=\sum_yq(y)\log\bigl(1/p_0(y)\bigr)\leq M_0$.
Combining the three,
\[
  \sum_{t<T}\langle q,u_t\rangle-\sum_{t<T}\langle p_t,u_t\rangle
  \;\leq\;\frac{M_0}{\eta}+\frac{\eta T}{8} .
\]
This is the standard regret bound for exponential weights
\citep[Thm.~2.2]{cesabianchi2006}, versions of which go back to
\citet{littlestone1994} and \citet{freund1999}, here started from $p_0$ instead of the uniform distribution.

\emph{Step 3 (identifying the game).} Write
$u_t=\bar P\,p_t+e_t$ with
$e_t=\eta^{-1}\sum_v(\pi_v/\beta_v)P_v(\check q_{v,t}-p_t)$, so that
$\|e_t\|_\infty\leq\tfrac12\Delta_t$ by the definition of
$\Delta_t$. Since each $P_v$ is a preference kernel and the weights are non-negative and
sum to one, $\bar P$ is one too, so
$\bar P(y,y')+\bar P(y',y)=1$ for all $y,y'$. Hence for any distribution $p$,
$2\langle p,\bar Pp\rangle=\sum_{y,y'}p(y)p(y')[\bar P(y,y')+\bar P(y',y)]=1$,
and in particular $\langle p_t,\bar Pp_t\rangle=\tfrac12$. Substituting $u_t=\bar Pp_t+e_t$ into Step~2, the term
$\langle p_t,u_t\rangle$ becomes $\tfrac12+\langle p_t,e_t\rangle$, and
$|\langle q-p_t,e_t\rangle|\leq\|q-p_t\|_1\|e_t\|_\infty\leq2\|e_t\|_\infty
\leq\Delta_t$, so
\[
  \sum_{t<T}\langle q,\bar Pp_t\rangle
  \;\leq\;\frac T2+\frac{M_0}{\eta}+\frac{\eta T}{8}
  +\sum_{t<T}\Delta_t .
\]
Dividing by $T$ and using linearity of
$q\mapsto\langle q,\bar P\bar p_T\rangle$ gives
$\langle q,\bar P\bar p_T\rangle\leq\tfrac12+\varepsilon_T$ for every
$q$. Since $\bar P$ is a preference kernel,
$\langle\bar p_T,\bar Pq\rangle
=1-\langle q,\bar P\bar p_T\rangle\geq\tfrac12-\varepsilon_T$, which is
\eqref{eq:cesaro}.

\emph{Step 4 (the two regimes).} If $\bar\Delta_T\to0$ then the disagreement
term vanishes and the horizon term tends to zero at fixed $\eta$, leaving
$\limsup_T\varepsilon_T\leq\eta/8$. Taking instead $\eta=\eta_T=1/\sqrt T$
makes the horizon term $M_0/\sqrt T$ and the step-size term
$1/(8\sqrt T)$, so $\varepsilon_T=O(M_0/\sqrt T)+\bar\Delta_T$ and
$\bar p_T$ approaches the maximal lotteries of $\bar P$ whenever
$\bar\Delta_T\to0$.
\proofend

\paragraph{Proof of Proposition~\ref{prop:defect}.}

\emph{Step 1: $\Delta_t\leq2\mathcal D_t$.} Split
$\|\check q_{v,t}-p_t\|_1\leq\|\check q_{v,t}-q_{v,t}\|_1+\|q_{v,t}-p_t\|_1$.
For any two agents, $\osc_y(\tilde f_{u,t}-\tilde f_{v,t})\leq2\mathcal D_t$,
since that difference lies in $[-\mathcal D_t,\mathcal D_t]$ at every
response, so Lemma~\ref{lem:softmax} and its factor $\tfrac12$ give
$\|q_{u,t}-q_{v,t}\|_1\leq\mathcal D_t$. The first term is a convex
combination of such distances, $\check q_{v,t}-q_{v,t}=\sum_ua_{vu}(q_{u,t}-q_{v,t})$,
and so is at most $\mathcal D_t$. For the second,
$p_t=\mathrm{softmax}(\sum_u\pi_u\tilde f_{u,t})$, and the same lemma gives
\[
  \|q_{v,t}-p_t\|_1
  \leq\tfrac12\osc_y\Bigl(\sum_u\pi_u(\tilde f_{v,t}-\tilde f_{u,t})\Bigr)
  \leq\tfrac12\max_u\osc_y(\tilde f_{v,t}-\tilde f_{u,t})
  \leq\mathcal D_t .
\]
Averaging over $v$ against $\pi$ gives $\Delta_t\leq2\mathcal D_t$.

\emph{Step 2: averaging over the horizon.} Theorem~\ref{thm:case4-directed}
gives
$\mathcal D_t\leq\gamma^{\,t}\mathcal D_0
+\bigl((1/\beta_{\min})\Delta_P+\Delta_\beta\bigr)/(1-\gamma)$ whenever $\gamma<1$.
Averaging over $t<T$ and using
$\tfrac1T\sum_{t<T}\gamma^{\,t}\leq1/\bigl((1-\gamma)T\bigr)$,
\[
  \frac1T\sum_{t<T}\mathcal D_t
  \;\leq\;\frac{\mathcal D_0}{(1-\gamma)T}
  +\frac{1}{1-\gamma}\Bigl(\frac{1}{\beta_{\min}}\Delta_P+\Delta_\beta\Bigr) ,
\]
and Step~1 doubles this to bound $\bar\Delta_T$, which is the stated
inequality. The two limiting claims follow: with a shared kernel and
temperature the second term vanishes, leaving $O(1/T)$; otherwise the first
term vanishes as $T\to\infty$.
\proofend

\subsection{Details of the language-model experiment of Section~\ref{sec:case3-experiment}}
\label{app:experiment}
This appendix gives the details of the experiment of Section~\ref{sec:case3-experiment}. Three language models are the agents, each aligned towards a different attribute that human annotators have scored, and each holding a belief over the two candidate responses to a prompt. They are connected on a hub-and-spoke graph, with the attention weights and temperatures of the two configurations given in Section~\ref{sec:case3-experiment}, and run the dynamics of case~(I): at every round each model forms the geometric mean of its neighbours'
beliefs, makes one pass over the prompts maximising the KL-regularised
objective \eqref{eq:general} against that reference, and publishes the
result. Theorem~\ref{thm:case3} predicts that the collective
policy concentrates on whichever response maximises the influence-weighted
reward $\bar r$, and the experiment asks whether it does when the agents optimise parametric policies and do not attain their maximisers exactly.
The paragraphs below specify the beliefs, the data, the rewards, the policies, the reference and the update, then two controls applied to the design and the reason the update is not preference-based.

\paragraph{Beliefs.} An agent's model assigns a probability to any response it
might produce. Its belief here is that distribution restricted to the two
rated responses and renormalised, and those two responses are fixed
throughout. With only two of them the expectation and the divergence in
\eqref{eq:general} are both exact, so no sampling enters.

\paragraph{Data.} HelpSteer2~\citep{wang2024helpsteer2} stores two responses to
each prompt in consecutive records, each rated $0$ to $4$ by human annotators on
helpfulness, correctness, coherence, complexity and verbosity. The agents are helpfulness, at the hub, and complexity and verbosity, at the spokes. Correctness is excluded because, as \citet{wang2024helpsteer2} report, it correlates with helpfulness at $0.94$, and would add nothing to the aggregate. Prompts on which all three attributes rate the two responses equally are
dropped, since no agent has a preference. From the remainder, $400$ are
selected so that among the prompts where the influence-weighted and unweighted
rules disagree, the $\bar r$-maximiser is the longer response in half the
cases and the shorter in the other half; the rest are filled with
length-balanced prompts on which the two rules agree. The balance holds by construction, so no adjustment is made afterwards.

\paragraph{Rewards.} $r_v(y)$ is the Likert score of response $y$ on attribute $v$, standardised for each attribute separately over the evaluation set, to mean zero and unit variance, so that no attribute counts for more merely because of the scale it is scored on. No reward model is fitted, so the prediction
$\arg\max_y\sum_v(\pi_v/\beta_v)r_v(y)$ is computed from human annotations
alone and is fixed before any policy is trained.

\paragraph{Policies and training.} Each agent is Qwen2.5-0.5B-Instruct with a
rank-$32$ low-rank adapter on the attention projections, all three initialised
from the same base model. An agent's belief over the two responses is the
softmax of their mean per-token log-probabilities; the mean rather than the sum
is needed because summed log-probabilities differ by tens of nats between
responses of different lengths, which fixes the belief at a point mass. Each round, agent $v$ takes one pass over the $400$ prompts, maximising
$\E_q[r_v]-\beta_v\KL(q\|\bar q_{v,t})$ by Adam at learning rate $5\times10^{-5}$.

\paragraph{The reference.} The neighbourhood geometric mean is, in log space,
a weighted average of the other agents' current log-probabilities,
$\log\bar q_{v,t}=\sum_ua_{vu}\log q_{u,t}$, recomputed every round. This is the one
departure from a standard alignment pipeline, where the reference is a frozen
checkpoint. It is also the operation used to combine separately trained models
at generation time, so the aggregated reference is not merely an analytical
device.

\paragraph{The update.} Each agent ascends the gradient of
$\E_q[r_v]-\beta_v\KL(q\,\|\,\bar q_{v,t})$ directly. Because $\Y$ holds two
fixed responses the expectation is computed in closed form, so the gradient is exact. Proximal policy optimisation maximises the same
objective, but its importance ratios, clipping and value function exist to
control the variance of a sampled expectation and have no role when the expectation is exact; they would leave the same gradient with more moving
parts.

\paragraph{Two controls.} First, verbosity is a length measure: the verbosity margin between the two responses correlates with their difference in length at $0.67$. Because a parametric policy
shares weights across prompts, all three adapters acquire a preference for
longer responses, which would let a length-driven policy score above chance
for the wrong reason: in an unbalanced version the drift common to the three
agents correlated at $0.53$ with the length difference, although the hub's
own reward margin correlated with it at $0.02$. The prompts are therefore
selected so that the $\bar r$-maximiser is the longer response in half the
discriminating cases and the shorter in the other half, which leaves that
correlation at $-0.11$ and the margin uncorrelated with length. Second, the
objective depends only on the difference of the two log-probabilities and not
their level, so it asks an agent to prefer one response and never to make either likely; the level is free, and in these runs it
falls. The theorem is a claim about the ratio, and the runs measure the ratio, and a deployed run would pin the
level with a divergence against a frozen reference over the whole
distribution.

\paragraph{Why not a preference-based update.} Direct preference optimisation
minimises the negative log-sigmoid of a margin. That is decreasing in the
margin, so it has no stationary point: each agent drives its own margin up
without bound and the disagreement grows without settling. The objective above has as its optimum the Boltzmann maximiser $\bar q_{v,t}e^{r_v/\beta_v}$ of \eqref{eq:case3}, so optimising it implements that update. With the preference-based update, the collective policy followed $\bar r$ no better than the unweighted rule.

\clearpage
\setcounter{section}{0}
\renewcommand{\thesection}{S\arabic{section}}
\renewcommand{\thesubsection}{S\arabic{section}.\arabic{subsection}}
\renewcommand{\thesubsubsection}{S\arabic{section}.\arabic{subsection}.\arabic{subsubsection}}
\setcounter{theorem}{0}
\renewcommand{\thetheorem}{S\arabic{section}.\arabic{theorem}}
\setcounter{equation}{0}
\renewcommand{\theequation}{S\arabic{equation}}
\setcounter{figure}{0}
\renewcommand{\thefigure}{S\arabic{figure}}
\renewcommand{\theHsection}{S\arabic{section}}
\renewcommand{\theHfigure}{S\arabic{figure}}
\renewcommand{\theHequation}{S\arabic{equation}}

\begin{center}
\Large\bfseries Supplementary Material
\end{center}
\medskip

\label{app:supplementary}

This supplementary material collects results and illustrations that extend the
body without being needed to follow it. The first section describes what the model represents. The next four follow
Sections~\ref{sec:setup} to~\ref{sec:case4} of the body in turn, Section~S$k$
holding the material for Section~$k$, as Appendix~A.$k$ holds its proofs; and the last treats graphs outside Assumption~\ref{ass:graph}.

\section{What the model represents}
\label{supp:model}
\label{app:modelling}
\label{app:deployments}

Each round is a round of training. Agent $v$ holds its reference
$\mathcal R_{v,t}$ and challenger $\mathcal C_{v,t}$ fixed, fine-tunes against
them until
\[
  \E_p\bigl[R_v(\cdot;\mathcal{C}_{v,t})\bigr]
  -\beta_v\KL\bigl(p\,\|\,\mathcal{R}_{v,t}\bigr)
\]
is optimised, then publishes and observes its neighbours. One application is a
complete local training run in a federated deployment, or one retraining
against the current teacher in a distillation pipeline, with $\beta_v$ fixing how far the run may move from the anchor.

The same expression supports a second interpretation. Mirror descent on the simplex maximises, at each step, a linear gain less the inverse step size times the divergence from the current point, which is the objective above with gain $R_v(\cdot;\mathcal C_{v,t})$, current point $\mathcal R_{v,t}$ and step size $1/\beta_v$. On this reading the maximiser is not approached over several
steps but \emph{is} one mirror-descent step, available in closed form because
a policy here may be any distribution over the finite set $\Y$; the same step
is the natural policy gradient under a softmax
parametrisation~\citep{kakade2001,agarwal2021,lan2023,cen2022}. The index $t$
then counts optimiser steps rather than rounds of exchange.

Either way the operational content is the same: the four dynamics differ only in what the trainer is given each round, as reference a fixed checkpoint in cases~(0) and~(II) or the pooled neighbourhood model in cases~(I) and~(III), and as challenger a fixed distribution in cases~(0) and~(I) or the neighbours' mixture in cases~(II) and~(III). A deployed agent ascends the same objective in the
parameters of a policy family, by PPO~\citep{schulman2017} as in the RL stage
of instruction tuning~\citep{ouyang2022} or by GRPO~\citep{shao2024}, both of which include the penalty $\beta_v\KL(p\|\mathcal{R}_{v,t})$ towards a reference
model; DPO~\citep{rafailov2023} removes the inner loop by reparameterisation, and its optimum coincides with the maximiser of \eqref{eq:general} when the challenger is fixed, the preferences follow the Bradley--Terry model of a reward, and the policy class is unconstrained.

The two aggregates of Section~\ref{sec:aggregates} are equally standard: the
arithmetic mean of the challengers and the geometric mean of the references
are linear and logarithmic pooling, the latter implemented as a weighted combination of logits and used to combine separately trained models at generation time~\citep{liu2021dexperts,liu2024proxy}.

Asymmetric attention is likewise what deployments produce. Any scheme
weighting agents by confidence, perplexity, or position in a hierarchy gives
an asymmetric $A$, and a distillation cascade is directed by construction.
Soft-label exchange on a complete graph and the doubly stochastic gossip
matrices of decentralised training weight participants equally, so the results
specialise there to a uniform weighting; the server-hub star of ensemble distillation does not. On the hub-and-spoke graph of Section~\ref{sec:running}, \eqref{eq:hub-pi} gives the hub $\pi_h=a/(a+b)$, so a hub that each spoke attends to as much as the hub attends to all the spokes together, $a=b$, holds half the influence however many spokes there are. A large pretrained
model enters as an agent whose own training is not modelled: the frozen source
of a distillation, or the fixed grader supplying the signal.

Two further idealisations separate this from practice, beyond the two that Section~\ref{sec:case3-experiment} tests: $R_v$ is estimated from a finite sample, where we take it as known, and an agent may observe only responses drawn from its neighbours, where we take their beliefs themselves.

\section{Section~\ref{sec:setup}: the model}
\label{supp:sec2}

This section collects supporting material for Section~\ref{sec:setup}: what
degenerate preference kernels do, and the structure shared by the three cases.

\subsection{Degenerate preference kernels}
\label{supp:degenerate}

An agent's quality signal $R_v(y;q)$ depends on the challenger $q$, and that
dependence is the only nonlinearity in the dynamics. Some kernels do not have it: the differences of the signal between responses do not depend on $q$, and the update is then affine in the log-beliefs. Such kernels are called
\emph{degenerate}, and the lemma below gives three equivalent ways to
recognise one.

That dependence on the challenger is measured by a bilinear form. Write $\Y_0$ for the subspace of $\mathbb{R}^\Y$ whose entries sum to zero; the
difference of any two distributions on $\Y$ lies in it. For $d,d'\in\Y_0$ define $B_P(d,d')=d^\top Pd'$, which measures that dependence exactly. Write $R(y;q)=\sum_{y'}P(y,y')q(y')$
for the signal, let $d=e_y-e_{\tilde y}$ compare two responses and let $d'$
be a change in the challenger. Then
\[
  \bigl[R(y;q+d')-R(\tilde y;q+d')\bigr]-\bigl[R(y;q)-R(\tilde y;q)\bigr]
  \;=\;\sum_{y'}\bigl[P(y,y')-P(\tilde y,y')\bigr]d'(y')
  \;=\;B_P(d,d') ,
\]
so moving the challenger by $d'$ shifts the signal's ranking of $y$ against
$\tilde y$ by $B_P(d,d')$, and the kernel is degenerate precisely when $B_P$
vanishes on $\Y_0$.

The form is antisymmetric, meaning $B_P(d,d')=-B_P(d',d)$, so in particular
$B_P(d,d)=0$: no change in the challenger alters the signal along its own direction. To see where it comes from, define
$S=P-\tfrac12\one\one^\top$: subtracting $\tfrac12$ from each of $P(y,y')$ and
$P(y',y)$ turns their sum of one into a sum of zero, so $S(y,y')=-S(y',y)$,
and a preference kernel is the constant matrix $\tfrac12\one\one^\top$ plus an
antisymmetric one. The constant part contributes nothing when both arguments
sum to zero, leaving $B_P(d,d')=d^\top Sd'$.

\begin{lemma}[Degenerate kernels]
\label{lem:degenerate}
For a preference kernel $P$ the following are equivalent, and a kernel
satisfying them is called \emph{degenerate}:
\begin{enumerate}[label=(\roman*),leftmargin=*]
  \item $B_P$ vanishes identically on $\Y_0$;
  \item the quality signal ignores the opposition, in that $R(y;q)-R(y';q)$
        takes the same value for every distribution $q$;
  \item fixed scores decide comparisons, in that $P(y,y')=\tfrac12+c(y)-c(y')$
        for some $c:\Y\to\mathbb{R}$.
\end{enumerate}
\end{lemma}

\begin{proof}
The proof begins with the antisymmetry of $B_P$ stated above. For
$d,d'\in\Y_0$, relabelling the summation
indices in the second term gives
\[
  B_P(d',d)=\sum_{y,y'}d'(y)d(y')P(y,y')=\sum_{y,y'}d(y)d'(y')P(y',y) ,
\]
so
\[
  B_P(d,d')+B_P(d',d)=\sum_{y,y'}d(y)d'(y')\bigl[P(y,y')+P(y',y)\bigr]
  =\Bigl(\sum_yd(y)\Bigr)\Bigl(\sum_{y'}d'(y')\Bigr)=0 ,
\]
using $P(y,y')+P(y',y)=1$ and then $\sum_yd(y)=\sum_{y'}d'(y')=0$.

Turning to the equivalence, write $P=\tfrac12J+S$ with $J$ the all-ones matrix; $P+P^\top=J$ is
equivalent to $S^\top=-S$. For $d,d'\in\Y_0$ we have $d^\top Jd'=0$, so
$B_P=B_S$ on $\Y_0$.

(i)$\Rightarrow$(iii): the inner product of a vector with $\one$ is the sum of
its entries, so $\Y_0=\{x:\langle x,\one\rangle=0\}$ is the orthogonal
complement of $\one$ in $\mathbb{R}^\Y$, a subspace of dimension $|\Y|-1$.
Subtracting from $x$ its mean entry projects onto it, which is
$\Pi=I-\tfrac{1}{|\Y|}J$. This $\Pi$ is symmetric, sends every $x$ into
$\Y_0$, and fixes $\Y_0$ pointwise. Hence
$B_S\equiv0$ on $\Y_0\times\Y_0$ is the same as $\Pi S\Pi=0$: if the form
vanishes there then for arbitrary $x,x'$,
\[
  x^\top\Pi S\Pi x'=(\Pi x)^\top S(\Pi x')=B_S(\Pi x,\Pi x')=0 ,
\]
and a bilinear form vanishing on all of $\mathbb{R}^\Y\times\mathbb{R}^\Y$ has
zero matrix; conversely, if $\Pi S\Pi=0$ then
$B_S(d,d')=(\Pi d)^\top S(\Pi d')=d^\top\Pi S\Pi d'=0$ for $d,d'\in\Y_0$.
Expanding the product,
\[
  0=\Pi S\Pi=S-\tfrac{1}{|\Y|}JS-\tfrac{1}{|\Y|}SJ+\tfrac{1}{|\Y|^2}JSJ ,
\]
so $S=\tfrac{1}{|\Y|}(JS+SJ)-\tfrac{1}{|\Y|^2}JSJ$. Every term on the right
has all its rows equal or all its columns equal, since $J=\one\one^\top$, so
$S=\one u^\top+v\one^\top$ for some $u,v\in\mathbb{R}^\Y$: the matrix is a
row-constant part plus a column-constant one. Antisymmetry, $S^\top=-S$,
reads $u\one^\top+\one v^\top=-\one u^\top-v\one^\top$, which reads $(u+v)(y)+(u+v)(y')=0$ for all $y,y'$; taking $y'=y$ gives $v=-u$. Hence
$S=\one u^\top-u\one^\top$, that is $S(y,y')=u(y')-u(y)$, and setting $c=-u$
gives $P(y,y')=\tfrac12+c(y)-c(y')$, which is (iii).

(iii)$\Rightarrow$(ii): if $P(y,y')=\tfrac12+c(y)-c(y')$ then
$R(y;q)=\tfrac12+c(y)-\E_q[c]$, so $R(y;q)-R(y';q)=c(y)-c(y')$ for every $q$.

(ii)$\Rightarrow$(i): suppose $R(y;q)-R(\tilde y;q)=:\Delta(y,\tilde y)$ does
not depend on $q$. Take any $d'\in\Y_0$, and write it as $d'=d'_+-d'_-$ with $d'_+$ and $d'_-$
its positive and negative parts, both non-negative. Since $d'$ sums to zero
the two have a common total $m=\sum_yd'_+(y)=\sum_yd'_-(y)$. If $m=0$ then
$d'=0$ and there is nothing to prove, so assume $m>0$ and replace $d'$ by
$d'/m$, which is legitimate because $B_P$ is linear in its second argument.
Now $d'_+$ and $d'_-$ are distributions and $d'=q_1-q_2$ with $q_1=d'_+$ and
$q_2=d'_-$. Then for any two responses $y,\tilde y$,
\begin{align*}
  \sum_{y'}\bigl[P(y,y')-P(\tilde y,y')\bigr]d'(y')
  &=\bigl[R(y;q_1)-R(\tilde y;q_1)\bigr]-\bigl[R(y;q_2)-R(\tilde y;q_2)\bigr]\\
  &=\Delta(y,\tilde y)-\Delta(y,\tilde y)=0 ,
\end{align*}
so the quantity $g(y):=\sum_{y'}P(y,y')d'(y')$ takes the same value at every
$y$. Writing $g\equiv\kappa$ for that common value,
\[
  B_P(d,d')=\sum_yd(y)\,g(y)=\kappa\sum_yd(y)=0
\]
for every $d\in\Y_0$, since such $d$ sum to zero. As $d$ and $d'$ were
arbitrary in $\Y_0$, the form vanishes identically.

\end{proof}

Every kernel on two responses is degenerate. With $\Y=\{y_0,y_1\}$ the space
$\Y_0$ is spanned by the single vector $d=(1,-1)$, and $B_P(d,d)=0$ by
antisymmetry, so (i) holds; equivalently, $c(y_0)=P(y_0,y_1)-\tfrac12$ and
$c(y_1)=0$ give (iii). This is why two responses behave exceptionally: the difference of the signals at the two responses is then constant along the whole trajectory. Degeneracy therefore matters only for the two dynamics whose quality signal is
endogenous; the other two are affine for every $\Y$ already.

\subsection{Shared structure}
\label{supp:shared}

This subsection collects three pieces of supporting material, none of which
is needed to read the statements in the body. The first reads off what each
term of the solution of Theorem~\ref{thm:fredholm} does. The second
interpolates between a fixed and an aggregated reference, showing that the
theorem's solvability condition belongs to the fully aggregated endpoint
rather than to aggregation as such. The third gives a form of $\tau_1$ that is
easier to evaluate, with its values for the graphs used as examples. The two
bounds that every contraction estimate in Sections~\ref{sec:case2}
and~\ref{sec:case4} runs through, on how much one round of averaging closes a
spread and how far an agent's quality signal moves with the distribution it is
measured against, are Lemmas~\ref{lem:dobrushin} and~\ref{lem:signal-lip} of
Appendix~\ref{app:shared}.

\paragraph{Reading the solution of Theorem~\ref{thm:fredholm}.}

Theorem~\ref{thm:fredholm}(b) gives the solutions of \eqref{eq:fp-compact} as
\[
  \mathbf f^*(y)=D\,B^{-1}\mathbf R^*(y)-D\boldsymbol\Phi^*+c(y)\,\one ,
\]
with $c(y)$ a scalar common to all agents. The last two terms do not distinguish the agents: $c(y)$ is the same at every agent, and
$-D\boldsymbol\Phi^*$ does not depend on $y$, so it shifts agent $v$'s
log-belief by one constant at every response and disappears under
normalisation. The agents are separated by $D\,B^{-1}\mathbf R^*$ alone, and
agent $v$'s position relative to the common level is
$\sum_u d_{vu}R^*_u/\beta_u$, where $d_{vu}$ measures the excess attention
$v$'s repeated averages pay $u$ over $u$'s long-run share $\pi_u$.

Two properties of $D$ from Lemma~\ref{lem:Dprops}(d) limit this. Since $D\one=0$, if every agent's scaled signal $R^*_u/\beta_u$ is the same then every offset vanishes, and the agents hold the same belief. Since $\pi^\top D=0$, the offsets have zero
$\pi$-weighted total, so an agent above the common level is matched by others
below it in proportion to influence; their unweighted sum need not vanish.
So $\pi$ fixes the level and $D$ fixes the spread around it. The entries $d_{vu}$ have both signs, since each row of $D$ sums to zero: an agent whose repeated averages under-weight $u$ relative to $\pi_u$ is pushed away from $u$'s signal, not towards it.

\paragraph{Interpolating the reference.}

The condition \eqref{eq:solvability} is a feature of a fully aggregated reference only, as the intermediate case shows.

\begin{proposition}[Interpolating the reference]
\label{prop:interpolate}
Assume that Assumption~\ref{ass:graph} holds, fix $\theta\in[0,1)$, and let each agent
anchor to the weighted geometric mean
$\mathcal R_{v,t}\propto\mu_v^{1-\theta}\,\bar q_{v,t}^{\,\theta}$ of its own
reference and its neighbours', with the challenger fixed, so that
$R_{v,t}=r_v$. Then the fixed-point equation reads $(I-\theta A)\,\mathbf f^*(y)=(1-\theta)\log\boldsymbol\mu(y)+B^{-1}\mathbf r(y)+\boldsymbol\kappa$, with $\boldsymbol\kappa$ a vector that does not depend on $y$, and:
\begin{enumerate}[label=(\alph*),leftmargin=*]
\item $I-\theta A$ is invertible, so this equation is solvable for every choice
  of rewards and there is no condition of the form
      \eqref{eq:solvability}. The fixed point is unique and reached from every
      start, geometrically at rate $\theta$.
\item Writing $W=(1-\theta)(I-\theta A)^{-1}$, the matrix $W$ is
      row-stochastic and the fixed point is the Gibbs policy
      \begin{equation}\label{eq:interp-gibbs}
        q^*_v(y)\;\propto\;\tilde\mu_v(y)\,
        \exp\Bigl(\bigl[(I-\theta A)^{-1}B^{-1}\mathbf r(y)\bigr]_v\Bigr),
        \qquad
        \tilde\mu_v\;\propto\;\prod_u\mu_u^{\,w_{vu}} ,
      \end{equation}
      whose reference is a weighted geometric mean of all the agents' anchors.
\item Fix two responses $y,y'$ and write
      $\phi^*_v=\log\bigl(q^*_v(y)/q^*_v(y')\bigr)$ for agent $v$'s log-odds
      at the fixed point. With
      $D_\theta=\sum_{s\geq0}\theta^s(A^s-\one\pi^\top)$, these split exactly
      into a part common to the network and a part particular to each agent:
      \[
        \phi^*_v
        =\underbrace{\pi^\top\boldsymbol\ell
        +\tfrac{1}{1-\theta}\bigl(\bar r(y)-\bar r(y')\bigr)}_{\text{the same at every agent}}
        +\underbrace{\bigl[D_\theta\bigl((1-\theta)\boldsymbol\ell
        +B^{-1}(\mathbf r(y)-\mathbf r(y'))\bigr)\bigr]_v}_{\text{differs between agents}} ,
      \]
      with $\ell_v=\log\bigl(\mu_v(y)/\mu_v(y')\bigr)$.
\item $D_\theta\to D$ as $\theta\to1^-$. Hence if $\bar r$ has a strict
      maximiser $y^*$ then $q^*_v\to\delta_{y^*}$ for every agent, and if
      $y,y'$ both maximise $\bar r$ then
      $\phi^*_v\to\pi^\top\boldsymbol\ell+[D\,B^{-1}(\mathbf r(y)-\mathbf r(y'))]_v$.
\end{enumerate}
\end{proposition}

\begin{proof}
\emph{(a)} Taking logarithms in the update gives
$\tilde{\mathbf f}_{t+1}=\theta A\tilde{\mathbf f}_t+\mathbf g$ with
$\mathbf g=(1-\theta)\log\boldsymbol\mu+B^{-1}\mathbf r$. Since $A$ is
row-stochastic, $\rho(A)=1$ and $\rho(\theta A)=\theta<1$, so $I-\theta A$ is
invertible with $(I-\theta A)^{-1}=\sum_{s\geq0}\theta^sA^s$ and the unique
fixed point is $(I-\theta A)^{-1}\mathbf g$. As
$\|\theta A\|_\infty=\theta$, the map is a contraction of modulus $\theta$ in
the supremum norm, so the iterates converge to it from every start.

\emph{(b)} Each $A^s$ has non-negative entries, so
$(I-\theta A)^{-1}\geq0$ entrywise; and $A\one=\one$ gives
$(I-\theta A)^{-1}\one=\sum_s\theta^s\one=\one/(1-\theta)$. Hence
$W=(1-\theta)(I-\theta A)^{-1}$ is non-negative with rows summing to one.
Splitting $\mathbf g$,
$\mathbf f^*=W\log\boldsymbol\mu+(I-\theta A)^{-1}B^{-1}\mathbf r$, and
exponentiating gives \eqref{eq:interp-gibbs}; the first factor is
$\prod_u\mu_u^{w_{vu}}$, a weighted geometric mean because the $w_{vu}$ are
non-negative and sum to one over $u$.

\emph{(c)} Differencing the fixed point of~(a) at $y$ and $y'$ gives
$\boldsymbol\phi^*=(I-\theta A)^{-1}\mathbf h$ with
$\mathbf h=(1-\theta)\boldsymbol\ell+B^{-1}(\mathbf r(y)-\mathbf r(y'))$.
Summing $A^s=\one\pi^\top+(A^s-\one\pi^\top)$ over $s$ with weights $\theta^s$
splits the resolvent,
\[
  (I-\theta A)^{-1}=\sum_{s\geq0}\theta^sA^s
  =\tfrac{1}{1-\theta}\one\pi^\top+D_\theta ,
\]
the first term being rank one and the second $D_\theta$ as defined. Applying
this to $\mathbf h$, the rank-one part contributes
$\tfrac{1}{1-\theta}(\pi^\top\mathbf h)\one$, which is the same at every agent,
and $\pi^\top\mathbf h=(1-\theta)\pi^\top\boldsymbol\ell+\bar r(y)-\bar r(y')$
because $\pi^\top B^{-1}(\mathbf r(y)-\mathbf r(y'))=\bar r(y)-\bar r(y')$ by
the definition of $\bar r$. The factor $(1-\theta)$ cancels on the first
summand, leaving the common term stated; the second part contributes
$[D_\theta\mathbf h]_v$, which is the offset.

\emph{(d)} By Lemma~\ref{lem:Dprops}(b), $\|A^s-\one\pi^\top\|_\infty\leq Cs^{m-1}|\lambda_2(A)|^s$ for $s\geq m$. The series defining $D_\theta$ is therefore dominated, apart from its first $m$ terms, by $\sum_sCs^{m-1}|\lambda_2(A)|^s<\infty$, uniformly in $\theta\in[0,1]$, so it converges uniformly and $\theta\mapsto D_\theta$ is
continuous on $[0,1]$ with $D_1=D$. The offset in~(c) is thus bounded as
$\theta\to1$ and tends to $[D\,B^{-1}(\mathbf r(y)-\mathbf r(y'))]_v$, while the
common term is $\tfrac{1}{1-\theta}(\bar r(y)-\bar r(y'))+O(1)$. If $y^*$
strictly maximises $\bar r$ then $\phi^*_v(y^*,y)\to+\infty$ for every $v$ and
every $y\neq y^*$, so $q^*_v\to\delta_{y^*}$; if $y$ and $y'$ both maximise
$\bar r$ the divergent part vanishes and the stated limit remains.

\end{proof}

\noindent The proposition locates the solvability condition of
Theorem~\ref{thm:fredholm}. For every $\theta<1$ there is a unique interior
fixed point and no such condition, by part~(a); only at $\theta=1$, where the
reference is entirely the neighbours' and none of it the agent's own anchor,
does one appear. The reason is visible in the resolvent. As $\theta\to1$ the
matrix $(I-\theta A)^{-1}=\tfrac{1}{1-\theta}\one\pi^\top+D_\theta$ has a bounded part $D_\theta\to D$ and a part diverging like $1/(1-\theta)$, and that divergence is confined to the rank-one term $\one\pi^\top$. At $\theta=1$ the
equation can therefore be solved only on the complement of that direction,
which is exactly what \eqref{eq:solvability} asks. Part~(d) then shows that Theorem~\ref{thm:case3}
and Corollary~\ref{cor:case3-ties} describe that endpoint: the concentration of
case~(I) is a family of interior fixed points receding to the boundary, with
$1/(1-\theta)$ in the role that the round $t$ plays there. Part~(b) places the
Gibbs forms of this paper on one scale, the reference running from $\mu_v$ at $\theta=0$ to the $\pi$-weighted geometric mean $\prod_u\mu_u^{\pi_u}$ as $\theta\to1$. Parts~(a) and~(b) continue to hold with agent-dependent weights
$\Theta=\mathrm{diag}(\theta_v)$, with $\rho(\Theta A)\leq\max_v\theta_v$ and
$W=(I-\Theta A)^{-1}(I-\Theta)$ again row-stochastic; parts~(c) and~(d) need a
common $\theta$, since $\pi^\top(I-\Theta A)^{-1}$ is proportional to
$\pi^\top$ only then. The construction interpolates the \emph{reference} only:
with an aggregated challenger $\mathbf R^*$ depends on the profile, the
equation ceases to be affine, and none of this applies.

\paragraph{Computing the ergodic coefficient.}
\label{app:tau1}

The ergodic coefficient of Definition~\ref{def:resolvent} has an equivalent
form that is often easier to evaluate. Two probability vectors satisfy
$\tfrac12\sum_u|p_u-q_u|=1-\sum_u\min\{p_u,q_u\}$, so
\begin{equation}\label{eq:tau1-overlap}
  \tau_1(A)=1-\min_{v,v'}\sum_u\min\bigl\{a_{vu},a_{v'u}\bigr\} ,
\end{equation}
one minus the smallest overlap between the weightings two agents use. In this
form $\tau_1(A)=1$ exactly when some two agents place their weight on
disjoint sets, and $\tau_1(A)=0$ exactly when all agents weight alike.

The table gives $\tau_1(A)$ for the graphs used as examples in this paper.
The columns for $\tau_1(A^2)$ and $\tau_1(A^3)$ show whether several rounds of averaging contract when one round does not; the thresholds stated in terms of $\tau_1(A)$, such as that of Theorem~\ref{thm:case4-directed}, give no conclusion when it equals one.

\begin{center}
\small
\setlength{\tabcolsep}{7pt}
\renewcommand{\arraystretch}{1.2}
\begin{tabular}{@{}lrrrr@{}}
\toprule
\emph{Graph} & $\tau_1(A)$ & $\tau_1(A^2)$ & $\tau_1(A^3)$ & $\|D\|_\infty$\\
\midrule
Complete graph, uniform weights & $0$ & $0$ & $0$ & $2(n-1)/n$\\
Running example, $k=6$, $a=0.5$, $b=0.15$ & $0.50$ & $0.25$ & $0.13$ & $3.73$\\
Introduction example, $k=2$, $a=0.2$, $b=0.6$ & $0.80$ & $0.64$ & $0.51$ & $5.31$\\
Ring of $10$, weight $\tfrac13$ on self and each neighbour
 & $1$ & $1$ & $0.85$ & $9.75$\\
Star without self-weights
 & $1$ & $1$ & $1$ & ---\\
\bottomrule
\end{tabular}
\end{center}

\noindent
On the uniform complete graph one round of averaging removes all disagreement,
so the condition loses its connectivity factor entirely. On the hub-and-spoke
graph of Section~\ref{sec:running} with $k\geq2$ leaves, \eqref{eq:tau1-overlap}
gives the closed form
\[
  \tau_1(A)=\max\Bigl\{\,1-a,\;
  1-\min\{1-b,\,a\}-\min\{b/k,\,1-a\}\Bigr\},
\]
since two spokes overlap only in the weight $a$ each places on the hub, while a
spoke and the hub overlap at the hub and at that spoke and nowhere else. The
first term is the spoke-to-spoke comparison, which is why $k\geq2$ is needed;
for $k=1$ the coefficient is $|1-a-b|$. In the regime $b/k\leq1-a$ and
$a\leq1-b$ the first term decides and $\tau_1=1-a$, the weight a spoke keeps on
itself, which is why the introduction's instance mixes more slowly than the
running example. The last two rows show that $\tau_1=1$ is common: on the ring
it falls below one at $m=3$, once neighbourhoods overlap, while the star in which each client adopts the server's model outright and the server averages the clients, keeping no weight on itself, is periodic, so
$\tau_1(A^m)=1$ for every $m$ and Assumption~\ref{ass:graph} fails; letting the agents retain any weight on themselves restores primitivity.

The last column gives the norm $\|D\|_\infty$ that bounds, in
Proposition~\ref{prop:case4-spread}(c), how far each agent's belief can lie
from the collective policy. It is never small: $(I-A)D=I-\one\pi^\top$, whose rows have $\ell_1$-norm $2(1-\pi_v)$, and $\|I-A\|_\infty\leq2$, so $\|D\|_\infty\geq\max_v(1-\pi_v)\geq1-1/n$. On the complete graph $D=I-\one\pi^\top$ and the bound is attained to within a factor of two; beyond it,
$\|D\|_\infty$ grows with how slowly the graph mixes, from $3.73$ on the
running example to $9.75$ on the ring. Lemma~\ref{lem:dobrushin} gives $\|D\|_\infty\leq2/(1-\tau_1(A))$. This is close for the running example, where it gives $4$, but uninformative for the ring, where $\tau_1(A)=1$
although $\|D\|_\infty$ is finite. On the star $A$ is periodic, so the series
defining $D$ does not converge.

\section{Section~\ref{sec:case3}: aggregated reference, fixed challenger}
\label{supp:case3}

This section collects supporting material for Section~\ref{sec:case3}: the proofs of the two consequences stated after Theorem~\ref{thm:case3} and two examples of that theorem, the reconciliation of the theorem with Theorem~\ref{thm:fredholm}, the reason a frozen boundary removes the concentration with an example of Theorem~\ref{thm:case3-partial}, and two examples of ties in $\bar r$.

\begin{corollary}[Relative positions settle]
\label{cor:case3-agree}
For any two agents, $\phi_{v,t}-\phi_{u,t}$ converges to
$[D\,B^{-1}(\mathbf r(y)-\mathbf r(y'))]_v-[D\,B^{-1}(\mathbf r(y)-\mathbf r(y'))]_u$
at the rate $|\lambda_2(A)|^t$, up to a factor polynomial in $t$. The agents therefore need not become equal, but come to differ by a fixed amount, and that amount is zero for every pair of agents and every pair of responses if and only if $r_v/\beta_v$ does not depend on $v$ up to an additive constant.
\end{corollary}

\paragraph{Proof of Corollary~\ref{cor:case3-agree}.}
\label{app:case3-corollaries}

Write \eqref{eq:case3-logodds} for two agents $u$ and $v$ and subtract. The
growing term $t(\bar r(y)-\bar r(y'))$ does not depend on the agent and
cancels, as does $\pi^\top\boldsymbol\phi_0$, leaving the difference of the offsets plus $\varepsilon_{v,t}-\varepsilon_{u,t}$, which vanishes at the rate
$|\lambda_2(A)|^t$, up to a factor polynomial in $t$, by the proof of Theorem~\ref{thm:case3}, which gives the stated limit. For the second claim, the offsets agree across all agents exactly when
$D\,B^{-1}(\mathbf r(y)-\mathbf r(y'))$ is a constant vector. By
Lemma~\ref{lem:Dprops}(d) that constant must be zero, since
$\pi^\top D=0$ and $\pi$ sums to one. So the condition is
$D\,B^{-1}(\mathbf r(y)-\mathbf r(y'))=0$, that is
$B^{-1}(\mathbf r(y)-\mathbf r(y'))\in\ker D$. That kernel is exactly
$\mathrm{span}(\one)$: it contains $\one$ because $D\one=0$, and if $Dx=0$
then applying $I-A$ and using $(I-A)D=I-\one\pi^\top$ gives
$x=(\pi^\top x)\one$. Hence the offsets vanish for every pair of responses
precisely when $(r_v(y)-r_v(y'))/\beta_v$ is the same at every agent for every
pair, which is to say $r_v/\beta_v$ is independent of $v$ up to an additive
constant, immaterial to the beliefs.
\proofend

\begin{corollary}[Concentration]
\label{cor:case3-conc}
Write
$\bar r(y)=\sum_v(\pi_v/\beta_v)r_v(y)$ for the influence-weighted reward and
suppose it has a unique maximiser $y^*$, with margin
\[
  \Delta_{\pi,\beta}=\bar r(y^*)-\max_{y\neq y^*}\bar r(y)>0 .
\]
Then every agent's belief converges to the point mass at $y^*$, and for
$y\neq y^*$ the odds of $y$ against $y^*$ are
$O\bigl(e^{-t\Delta_{\pi,\beta}}\bigr)$.
\end{corollary}

\paragraph{Proof of Corollary~\ref{cor:case3-conc}.}

Fix an agent $v$ and a response $y\neq y^*$, and apply
\eqref{eq:case3-logodds} with the pair $(y,y^*)$. Its growing term is
$t(\bar r(y)-\bar r(y^*))\leq-t\Delta_{\pi,\beta}$, while the other three
terms are bounded in $t$: $\pi^\top\boldsymbol\phi_0$ and the offset
$[D\,B^{-1}(\mathbf r(y)-\mathbf r(y^*))]_v$ do not depend on $t$, and
$\varepsilon_{v,t}\to0$. Writing $M_v$ for a bound on their sum,
\[
  \frac{q_{v,t}(y)}{q_{v,t}(y^*)}=e^{\phi_{v,t}(y,y^*)}
  \;\leq\;e^{M_v}e^{-t\Delta_{\pi,\beta}} ,
\]
which is the stated rate. Summing over the $|\Y|-1$ responses $y\neq y^*$
gives $1-q_{v,t}(y^*)\leq q_{v,t}(y^*)\sum_{y\neq y^*}e^{M_v}e^{-t\Delta_{\pi,\beta}}
\to0$, so $q_{v,t}\to\delta_{y^*}$. The bound holds for every agent, with only
the constant $M_v$ depending on $v$.
\proofend

Two examples illustrate Theorem~\ref{thm:case3}: the first shows what the
graph contributes to an agent's weight $\pi_v/\beta_v$, the second what its
temperature contributes.

\begin{example}[The spokes' weight does not grow with their number]
\label{ex:case3-hub}
In case~(I) with two responses, Theorem~\ref{thm:case3} reduces to a single sign condition, which shows what a group of agents can and cannot do to change
the outcome. Take $\Y=\{y_0,y_1\}$ and write $\Delta r_v=r_v(y_0)-r_v(y_1)$ for
how much agent $v$ prefers $y_0$. The drift in \eqref{eq:case3-logodds} is
then the single number $S=\sum_u(\pi_u/\beta_u)\Delta r_u$, the same at every
agent, and every belief converges to the point mass on $y_0$ if $S>0$ and on
$y_1$ if $S<0$. On the hub-and-spoke graph of Section~\ref{sec:running},
substituting \eqref{eq:hub-pi} and multiplying by $a+b>0$, which does not
change the sign, $y_0$ wins exactly when
\[
  \frac{a\,\Delta r_h}{\beta_h}
  \;+\;\frac{b}{k}\sum_{\ell}\frac{\Delta r_\ell}{\beta_\ell}\;>\;0 ,
\]
the hub's preference weighed against the spokes'. The spokes enter only
through the average $\tfrac1k\sum_\ell\Delta r_\ell/\beta_\ell$, scaled by the total attention $b$ the hub pays them and not by how many they are. Adding spokes refines that average without increasing its weight. The spokes can instead gain by being attended to more, which raises $b$ relative to $a$, or by optimising harder, which lowers $\beta_\ell$.
\end{example}

\begin{figure}[htb]
\centering
\includegraphics[width=0.62\textwidth]{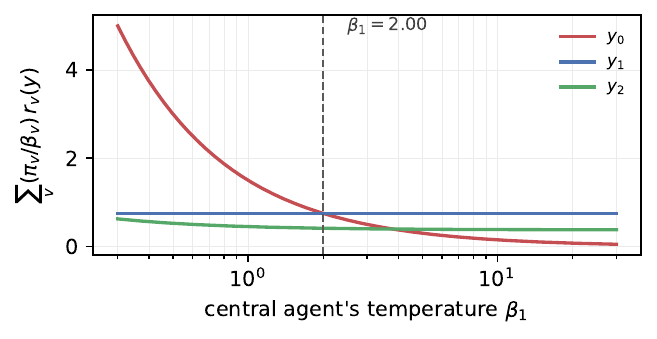}
\caption{The influence-weighted reward $\bar r(y)$ of each response as the
central agent's temperature $\beta_1$ varies, the others held at $1$; see
Example~\ref{ex:weights}.}
\label{fig:weights}
\end{figure}

\begin{example}[One agent changes the answer by changing its temperature]
\label{ex:weights}
On the instance of Section~\ref{sec:intro-example} in case~(I), the central
agent is the least influential of the three, $\pi_1=\tfrac28$ against
$\tfrac38$ for each of the others, yet at equal temperatures $\beta_v=1$ its
large reward puts the maximiser of the influence-weighted reward
$\bar r(y)=\sum_v(\pi_v/\beta_v)r_v(y)$ at $y_0$, and every agent concentrates
there by Theorem~\ref{thm:case3}. Raising $\beta_1$ alone reverses this:
$\bar r(y_0)=1.5/\beta_1$ and $\bar r(y_1)=0.75$ cross at $\beta_1=2$, where the
central agent's weight $\pi_1/\beta_1$ has halved, and beyond that every agent
concentrates on $y_1$. Only the central agent's temperature has changed; the preferences are the same. Figure~\ref{fig:weights} tracks $\bar r$.
\end{example}

\paragraph{Reconciling the trajectory with the fixed point.}
\label{app:reconcile}

Theorem~\ref{thm:case3} and Theorem~\ref{thm:fredholm} appear to differ. The first has every agent concentrating on the maximiser of $\bar r$,
a limit the dynamics reach from any start. The second shows that the fixed-point equation \eqref{eq:fp-compact} is solvable only under the
condition \eqref{eq:solvability}, which for this case asks $\bar r$ to be
constant. A limit that is reached from every start would be expected to be a
fixed point, so the two statements must be reconciled. The reconciliation rests on $\bar r$, since the growing term of
\eqref{eq:case3-logodds} is $t(\bar r(y)-\bar r(y'))$. If
$\bar r$ is not constant, the log-odds of $y$ over $y'$ grow without bound
whenever $\bar r(y)>\bar r(y')$, so the maximisers of $\bar r$ defeat every other response and the agents concentrate on them. If $\bar r$ is constant the growing term
vanishes, the log-odds settle at
$\pi^\top\boldsymbol\phi_0+[D\,B^{-1}(\mathbf r(y)-\mathbf r(y'))]_v$,
and this is the fixed point of Theorem~\ref{thm:fredholm}, the first term being
the difference $c(y)-c(y')$ of the free constants in
\eqref{eq:resolvent-sol}. Constancy of $\bar r$ is exactly the solvability
condition \eqref{eq:solvability}, so an interior fixed point exists only under
a tie and is then one of a family, the initialisation deciding which member is reached, so the two statements are consistent. What Theorem~\ref{thm:fredholm}
denies for non-constant $\bar r$ is an \emph{interior} limit, not a limit: the
concentrated profile is a fixed point, since a geometric mean of atoms is that
atom and reweighting a point mass leaves it unchanged, but it lies on the
boundary, where the log-odds are infinite and \eqref{eq:fp-compact} does not
apply. 

Both regimes appear in a single instance if the rewards are set so that
$\bar r$ ties between two responses but not a third. Against the third the
growing term is present and the agents concentrate; between the tied two it
is absent and they settle in the interior, at the limit
Corollary~\ref{cor:case3-ties} gives. Example~\ref{ex:ties} exhibits this, and shows that the offsets can break the tie in opposite directions at different
agents, so that a tie in $\bar r$ need not leave the network agreeing.

\paragraph{Why a boundary removes the concentration.}
\label{app:partial}

Theorems~\ref{thm:case3} and~\ref{thm:case3-partial} describe the same
dynamics and reach opposite conclusions: with every agent updating and
$\bar r$ having a unique maximiser the beliefs concentrate on it, and with at least one agent frozen, and every updating agent reaching one, they converge to an interior profile. Neither statement explains why so small a change in the graph should change the outcome so completely. The whole of the difference is whether the weights of the updating agents sum
to one. With every agent updating, $\AUU=A$ is row-stochastic, so
$(I-A)\one=0$ and $I-A$ is singular. The forcing $B^{-1}\mathbf r$ has a
component along the corresponding left null vector $\pi$ which the iteration
therefore cannot absorb into a finite limit, and it accumulates instead: the log-odds of $y$ against $y'$ grow linearly at rate $\bar r(y)-\bar r(y')$ and the beliefs leave every compact subset of the interior. Nor is there an interior
fixed point to converge to instead, since by Theorem~\ref{thm:fredholm} one
exists only when $\pi^\top B^{-1}\mathbf r(y)$ does not depend on $y$. With a
non-empty boundary, $\AUU$ is substochastic and a reachable boundary makes
$\rho(\AUU)<1$ by Lemma~\ref{lem:substoch}. Then $I-\AUU$ is invertible, no
direction is left unabsorbed, and the same iteration converges to the interior
profile \eqref{eq:case3-partial}. A boundary does not slow the concentration
down; it removes the mechanism that produced it.

One qualification applies to Theorem~\ref{thm:case3-partial}. Its limit has
positive mass on every response, as claimed, but its tilt is $ZB^{-1}\mathbf r$, and $Z$ grows as the updating agents attend less to the
boundary, so the masses left on disfavoured responses can be exponentially
small. A frozen agent guarantees that the network stops, and determines where; it
does not guarantee that the beliefs it stops at are spread out.

\begin{figure}[htb]
\centering
\includegraphics[width=0.62\textwidth]{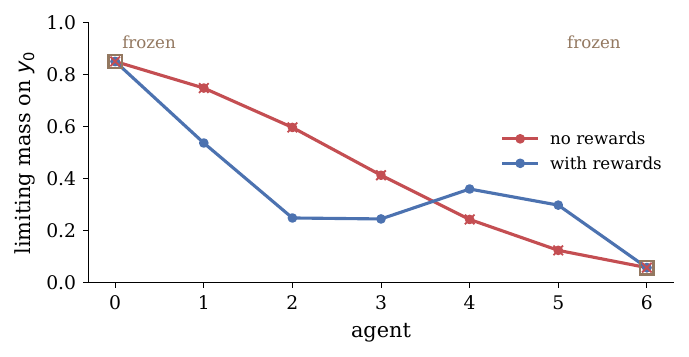}
\caption{A line with both ends frozen; see Example~\ref{ex:frozen}. Each
updating agent's limit, with rewards and without, against the interpolation
between the frozen beliefs that the absorption weights $h_{vb}$ of
Theorem~\ref{thm:case3-partial} predict, marked by crosses.}
\label{fig:frozen}
\end{figure}

\begin{example}[A line with both ends frozen]
\label{ex:frozen}
Take seven agents in a line in case~(I) with three responses, the left end frozen at $(0.85,0.09,0.06)$ and the right end at its mirror image $(0.06,0.09,0.85)$, each updating agent placing $0.45$ on each neighbour and $0.10$ on itself, starting uniform and with temperature~$1$. Without rewards every updating agent converges to the weighted
geometric mean of the two ends, the weights on one end being
$(h_{vb})_{v\in\Alg}=\bigl(\tfrac56,\tfrac23,\tfrac12,\tfrac13,\tfrac16\bigr)$
in order along the line and those on the other end one minus these. With rewards, drawn independently at each updating agent with standard deviation $0.6$, each limit is tilted further, by the term $ZB^{-1}\mathbf r$ of Theorem~\ref{thm:case3-partial}, the rewards accumulated along the walk
before it is absorbed. In both cases the limits are interior and differ from
agent to agent, whereas with every agent updating, rewards whose
influence-weighted sum has a unique maximiser make all of them concentrate on
that response. Figure~\ref{fig:frozen} shows the limits with and without
rewards against the interpolation $\prod_b q_{b,0}^{h_{vb}}$ that the weights
predict.
\end{example}

\paragraph{Ties in $\bar r$.} The remaining two examples concern ties, the case of Corollary~\ref{cor:case3-ties}.

\begin{figure}[t]
\centering
\includegraphics[width=0.62\textwidth]{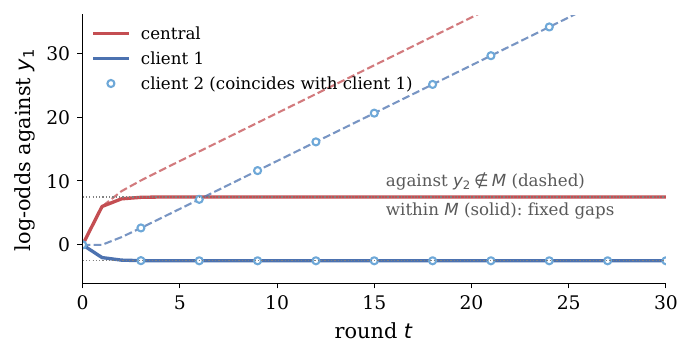}
\caption{A tie between two of three responses; see Example~\ref{ex:ties}.
Solid curves are log-odds within $M$, dashed against $y_2\notin M$. The two clients' curves coincide, by symmetry, and client~2's are marked by circles.}
\label{fig:ties}
\end{figure}

\begin{example}[A tie between two of three responses]
\label{ex:ties}
Take the graph of Section~\ref{sec:intro-example} with temperatures $\beta_v=1$, starting uniform, and rewards $(6,0,0)$ at the central agent and $(0,2,0)$ at each client, so that $\bar r(y_0)=\bar r(y_1)=1.5>\bar r(y_2)=0$, making $M=\{y_0,y_1\}$. Against
$y_2\notin M$ the log-odds grow without bound, so that response is
extinguished. Within $M$ they settle at $+7.5$, $-2.5$ and $-2.5$, the offset differences $[D\,B^{-1}\mathbf r]_v(y_0)-[D\,B^{-1}\mathbf r]_v(y_1)$ of Corollary~\ref{cor:case3-ties}. The
tie is therefore broken agent by agent, and in opposite directions for the central agent and the clients. Figure~\ref{fig:ties} shows the trajectories.
\end{example}

\paragraph{What decides a tie: starting beliefs or rewards.}

Corollary~\ref{cor:case3-ties} gives the limit on a tied set $M$ as
$\pi^\top\boldsymbol\phi_0$, the same at every agent, plus the agent's own
offset. The two terms have different sources and the corollary does not separate their contributions, so separating them is informative. The example
below holds $\bar r$, the graph and the starting beliefs fixed and varies only
whether the agents' rewards agree. With identical agents the offsets vanish and the limit is the $\pi$-weighted initial imbalance alone; with
differing agents that have the same $\bar r$, the offsets are non-zero and the agents end apart, though their $\pi$-weighted average is unchanged.
The second case also shows that an agent whose own rewards do not separate the
tied responses need not end between its neighbours: it can take one
neighbour's position outright.

\begin{example}[Two ways a tie can end]
\label{ex:ties-triangle}
Take three agents on a directed triangle, each keeping half its weight and
giving half to the next, so that $\pi=(\tfrac13,\tfrac13,\tfrac13)$, and three
responses, with $\beta_v\equiv1$ and rewards chosen so that
$\bar r(y_0)=\bar r(y_1)=1$ and $\bar r(y_2)=0$, making $M=\{y_0,y_1\}$. The
agents start from unequal beliefs, with log-odds of $y_0$ against $y_1$ equal to $\boldsymbol\phi_0=(0.6,-0.3,0.2)$, so that $\pi^\top\boldsymbol\phi_0=\tfrac16$.
In both cases below the limiting log-odds on $M$ are a fixed point of the
recursion $\boldsymbol\phi\mapsto A\boldsymbol\phi+B^{-1}(\mathbf r(y_0)-\mathbf
r(y_1))$, which is \eqref{eq:case3} restricted to the tied pair.
\begin{itemize}[leftmargin=*,itemsep=1pt]
  \item \emph{identical agents}, $r_v=(1,1,0)$ for every $v$. All three settle at log-odds $\pi^\top\boldsymbol\phi_0=\tfrac16$ of $y_0$
  against $y_1$.
        The offsets vanish, so the agents agree exactly; the limit is not
        uniform on $M$, and what it retains is precisely the initial
        imbalance.
  \item \emph{differing agents}, $r_1=(1.4,0.6,0)$, $r_2=(0.6,1.4,0)$ and $r_3=(1,1,0)$, which still give
  $\bar r=(1,1,0)$. Now the limits differ: $+0.7$, $-0.9$ and $+0.7$ for the same pair. The agents disagree permanently within $M$.
\end{itemize}
In both cases the mass leaves $y_2$: its log-odds against $y_0$ fall linearly at
rate $\bar r(y_0)-\bar r(y_2)=1$ per round, so the limit lies on the edge of the simplex spanned by $M$, with the two log-odds involving $y_2$ divergent and the
one within $M$ finite. The numbers above report that surviving coordinate. In both cases its $\pi$-weighted average is $\pi^\top\boldsymbol\phi_0=\tfrac16$, as Corollary~\ref{cor:case3-ties}
requires: since $\pi^\top D=0$, the offsets redistribute the log-odds among
the agents but cannot move that average. The positive sign is therefore
inherited from the initial beliefs, the rewards being tied. Agent~3 illustrates the redistribution. It attends only to itself and to
agent~1, and its own rewards do not separate $y_0$ from $y_1$, so the third
row of the fixed-point equation reads
$\phi_3=\tfrac12\phi_1+\tfrac12\phi_3$ and forces $\phi_3=\phi_1$: it takes agent~1's position outright. Agents~1
and~3 therefore agree while agent~2 does not, so the disagreement is between two camps.

\end{example}

\section{Section~\ref{sec:case2}: fixed reference, aggregated challenger}

This section collects supporting material for Section~\ref{sec:case2}. The first part describes the limit where
the fixed-point equation admits a closed form, by expanding about the agents'
own anchors when they are cautious. The next three treat the behaviour below the threshold, where uniqueness can fail and the iterates can
circulate. The last two concern the reversible case, in which the threshold disappears altogether, the identity the argument rests on being stated first.

\paragraph{The expansion about the anchors.}
\label{app:case2-expansion}

The fixed-point equation \eqref{eq:case2} has no closed-form solution in
general. When the agents are cautious it has an approximate one. An agent that
optimises slowly moves little from its anchor each round, so its limit lies
near $\mu_v$, and the leading correction evaluates each neighbour at
\emph{its} anchor rather than at its limit. The result below makes that
precise, bounds the error, and records how far through the graph the first few
iterates can see.

\begin{proposition}[Local expansion of the limit]
\label{prop:case2-expansion}
Assume $\beta_v>\tfrac12$ for all $v$, so that by
Proposition~\ref{prop:case2-geom} the map $T$ of \eqref{eq:case2} is a
contraction of modulus $\gamma=1/(2\beta_{\min})<1$ in
$d(\mathbf q,\mathbf q')=\max_v\|q_v-q'_v\|_1$. Write $\boldsymbol\mu=(\mu_v)_{v\in V}$ for the
profile of fixed references of Section~\ref{sec:setup}. Then:
\begin{enumerate}[label=(\alph*),leftmargin=*]
  \item \emph{(Local dependence.)} $\mathbf q^*=\lim_kT^k(\boldsymbol\mu)$ with
        $d\bigl(T^k(\boldsymbol\mu),\mathbf q^*\bigr)\leq
        \dfrac{\gamma^k}{2(1-\gamma)\beta_{\min}}$, and $T^k(\boldsymbol\mu)_v$
        depends only on $\mu_u$ and $P_u$ for those $u$ from which $v$ is
        reachable in $G$ by a directed path of length at most $k$, together with
        $\mu_v$ and $P_v$.
  \item \emph{(Leading term.)} Define
        \begin{equation}\label{eq:case2-expansion}
          q^{(1)}_v(y)\;\propto\;\mu_v(y)\,
          \exp\Bigl(\frac{1}{\beta_v}\sum_ua_{vu}R_v\bigl(y;\mu_u\bigr)\Bigr).
        \end{equation}
        Then $q^{(1)}=T(\boldsymbol\mu)$ and
        \begin{equation}\label{eq:case2-expansion-err}
          \max_v\bigl\|q^*_v-q^{(1)}_v\bigr\|_1
          \;\leq\;\frac{\gamma}{2(1-\gamma)\beta_{\min}}
          \;=\;\frac{1}{4(1-\gamma)\beta_{\min}^2}
          \;=\;O\bigl(1/\beta_{\min}^2\bigr).
        \end{equation}
\end{enumerate}
\end{proposition}

\begin{proof}
\emph{Step 1: one round moves little.} Both
$T_v(\mathbf q)=\mathrm{softmax}(\log\mu_v+R_v(\cdot;\check q_v)/\beta_v)$ and
$\mu_v=\mathrm{softmax}(\log\mu_v)$ are softmaxes of log-vectors differing by
$R_v(\cdot;\check q_v)/\beta_v$, so Lemma~\ref{lem:softmax} gives
$\|T_v(\mathbf q)-\mu_v\|_1\leq\osc_y\bigl(R_v(y;\check q_v)\bigr)/(2\beta_v)$.
Since $R_v(y;q)=\E_{y'\sim q}[P_v(y,y')]\in[0,1]$ for every $y$ and $q$, that
oscillation is at most $1$, so
$\|T_v(\mathbf q)-\mu_v\|_1\leq1/(2\beta_v)$ for every profile $\mathbf q$ and
hence $d(\boldsymbol\mu,T(\boldsymbol\mu))\leq1/(2\beta_{\min})$.

\emph{Step 2: distance from the anchor to the limit.} By the triangle
inequality and the contraction property,
\[
  d(\boldsymbol\mu,\mathbf q^*)
  \;\leq\;\sum_{j\geq0}d\bigl(T^j(\boldsymbol\mu),T^{j+1}(\boldsymbol\mu)\bigr)
  \;\leq\;d\bigl(\boldsymbol\mu,T(\boldsymbol\mu)\bigr)\sum_{j\geq0}\gamma^j
  \;=\;\frac{d\bigl(\boldsymbol\mu,T(\boldsymbol\mu)\bigr)}{1-\gamma} ,
\]
which with Step 1 gives
$d(\boldsymbol\mu,\mathbf q^*)\leq1/\bigl(2(1-\gamma)\beta_{\min}\bigr)$.

\emph{Step 3: part (a).} The profile $\mathbf q^*$ is the unique fixed point
by Proposition~\ref{prop:case2-fp}(b), so $T^k(\mathbf q^*)=\mathbf q^*$, and
applying the contraction property $k$ times,
\[
  d\bigl(T^k(\boldsymbol\mu),\mathbf q^*\bigr)
  \;=\;d\bigl(T^k(\boldsymbol\mu),T^k(\mathbf q^*)\bigr)
  \;\leq\;\gamma^k\,d(\boldsymbol\mu,\mathbf q^*) ,
\]
and Step 2 bounds the last factor.
For the dependence claim, $T_v$ depends only on $\mu_v$, $P_v$ and the entries $q_u$ with $a_{vu}>0$; by induction, $T^k(\boldsymbol\mu)_v$ therefore depends on $\mu_u$ and $P_u$ only for $u$ joined to $v$ by a directed path of length at most $k$, since
$a_{vu}>0$ exactly when $(u,v)\in E$.

\emph{Step 4: part (b).} The map $q\mapsto R_v(y;q)$ is linear, so
$R_v(y;\check\mu_v)=\sum_ua_{vu}R_v(y;\mu_u)$ and therefore $q^{(1)}=T(\boldsymbol\mu)$
by \eqref{eq:case2}. Taking $k=1$ in Step 3 gives
\[
  d\bigl(q^{(1)},\mathbf q^*\bigr)
  \;\leq\;\gamma\,d(\boldsymbol\mu,\mathbf q^*)
  \;\leq\;\frac{\gamma}{2(1-\gamma)\,\beta_{\min}} ,
\]
Substituting $\gamma=1/(2\beta_{\min})$ turns the right-hand side into
$1/\bigl(4(1-\gamma)\beta_{\min}^2\bigr)$, which is
\eqref{eq:case2-expansion-err}; as $\beta_{\min}>\tfrac12$ the factor
$(1-\gamma)^{-1}$ is finite, and it is bounded on any range
$\beta_{\min}\geq\tfrac12+\delta$, so the bound is $O(1/\beta_{\min}^2)$
there.
\end{proof}

The error in \eqref{eq:case2-expansion-err} contains two powers of $\beta_{\min}$ because two approximations of the same order compound. A single
round moves an agent's log-policy by at most $1/\beta_v$, so its limit lies
within $O(1/\beta_{\min})$ of its own anchor; the expansion then evaluates each neighbour at that neighbour's anchor instead of its limit, an error
of the same order, which the division by $\beta_v$ in the exponent reduces by
a further factor. Doubling every temperature therefore halves how far the
agents move and quarters the error in describing it.

The bound is informative only for cautious agents. As $\beta_{\min}$
approaches $\tfrac12$ the modulus $\gamma$ approaches one and
$(1-\gamma)^{-1}$ diverges, so the right-hand side eventually exceeds $2$,
which is the largest the left-hand side can be. A second limitation is on what the leading term \eqref{eq:case2-expansion}
can express. The only place the graph enters it is
$\sum_ua_{vu}R_v(\cdot\,;\mu_u)$, a weighted average of $v$'s signal evaluated
at its neighbours' anchors. If those anchors are all equal, to $\mu$ say, that
average is $R_v(\cdot\,;\mu)$ whatever the weights $a_{vu}$ are, and the
expansion reduces to the single-agent formula with no trace of the network.
Any dependence of the expansion on the network therefore comes from the agents differing, which is Corollary~\ref{cor:case2-homog} seen at this order.

\paragraph{Below the threshold.}

The convergence results of Section~\ref{sec:case2} assume $\beta_v>\tfrac12$ at every agent $v$, which makes the update map a contraction. The assumption cannot simply be dropped: the instance below has a
unique fixed point from which every other trajectory moves away, so at low
enough temperature the iterates converge from no starting belief but that one.
The next two parts then give what can still be said below the threshold.

\begin{proposition}[Convergence can fail]
\label{prop:case2-below}
Take one agent with a self-loop, three responses, a uniform prior, and the
cyclic kernel on three responses, $P=\tfrac12+\tfrac{\kappa}{2}C$ with
$\kappa\in(0,1]$, where $y_0$ beats $y_1$ beats $y_2$ beats $y_0$ and
$C_{yy'}\in\{0,\pm1\}$
 indicates which of $y,y'$ wins. The unique fixed point is the uniform distribution, and it is
repelling whenever
\begin{equation}\label{eq:case2-unstable}
    \beta\;<\;\frac{\kappa}{2\sqrt3} .
\end{equation}
The iterates then converge for no starting belief other than the fixed point
itself.
\end{proposition}

\begin{proof}
Here $n=1$, $a_{vv}=1$, $\Y=\{1,2,3\}$ and $\mu$ is uniform, so
$\check q_t=q_t$ and the update is $T(q)=\mathrm{softmax}\bigl(\log\mu+P q/\beta\bigr)$.
Write $C$ for the antisymmetric matrix with $C_{yy'}=1$ when $y$ beats $y'$ in
the cycle $y_0\succ y_1\succ y_2\succ y_0$, $-1$ when it loses and $0$ when
$y=y'$, so that
$P=\tfrac12\one\one^\top+\tfrac{\kappa}{2}C$ and $P$ takes values in $[0,1]$ for
$\kappa\in(0,1]$. Each row of $C$ contains one $+1$, one $-1$ and one $0$, so
$C\one=0$, and $C^\top=-C$ gives $\one^\top C=0$; hence $C$ maps $\Y_0$ into
itself.

\emph{The fixed point.} At $q=\one/3$ we get $Cq=0$, so $Pq=\tfrac12\one$ is
constant across responses and $T(q)=\mathrm{softmax}(\log\mu)=\mu=\one/3$. Thus
the uniform distribution is a fixed point, and it is the only one because the
single-agent equation $q\propto\mu\exp(R(\cdot;q)/\beta)$ has a unique solution
at every $\beta>0$, as
noted after \eqref{eq:nash}.

\emph{The linearisation.} The derivative of the softmax at a point where it
returns $q$ is $S_q=\mathrm{diag}(q)-qq^\top$, and at $q=\one/3$ this is
$S=\tfrac13 I-\tfrac19\one\one^\top=\tfrac13\Pi$, where
$\Pi=I-\tfrac13\one\one^\top$ projects onto $\Y_0$. The map is $T(q)=\sigma\bigl(\log\mu+Pq/\beta\bigr)$, a composition of the
affine map $q\mapsto\log\mu+Pq/\beta$, whose derivative is $P/\beta$, with the
softmax, whose derivative at the fixed point is $S$. By the chain rule
$\mathrm dT=S\,P/\beta$. Writing $P=\tfrac12\one\one^\top+\tfrac{\kappa}{2}C$,
the constant part contributes nothing, since
$S\one\one^\top=\tfrac13\Pi\one\one^\top=0$ because $\Pi\one=0$, and so
\[
  \mathrm dT=\frac{1}{\beta}\cdot\frac{\kappa}{2}\,S\,C
  =\frac{\kappa}{6\beta}\,\Pi C .
\] Since $C$ preserves $\Y_0$ and $\Pi$ acts as the identity
there, the restriction of $\mathrm dT$ to $\Y_0$ is $\tfrac{\kappa}{6\beta}C$.
The eigenvalues of $C$ are $0$ and $\pm i\sqrt3$, the first belonging to $\one$
and the other two to $\Y_0$, so the two eigenvalues of $\mathrm dT$ on $\Y_0$
both have modulus $\kappa\sqrt3/(6\beta)$. Only $\Y_0$ is relevant, the iterates
being probability vectors. Condition \eqref{eq:case2-unstable} is exactly the
statement that this modulus exceeds one.

\emph{Repulsion.} Suppose \eqref{eq:case2-unstable} holds and set
$\lambda=\kappa\sqrt3/(6\beta)>1$. Every eigenvalue of the restriction
$L=\mathrm dT|_{\Y_0}$ has modulus $\lambda$, so $L$ is invertible with
$\rho(L^{-1})=\lambda^{-1}<1$; choosing a norm on $\Y_0$ in which
$\|L^{-1}\|<1$, we get $\|Lx\|\geq c\|x\|$ for all $x\in\Y_0$ with
$c=\|L^{-1}\|^{-1}>1$. Since $T$ is $C^1$, there are $\varepsilon>0$ and
$c'\in(1,c)$ with
$\|T(q)-\tfrac13\one\|\geq c'\|q-\tfrac13\one\|$ whenever
$\|q-\tfrac13\one\|\leq\varepsilon$. Hence an orbit at distance $r$ from the fixed point, with
$0<r\leq\varepsilon$, is at distance at least $c'r$ after one step, at least
$c'^{\,k}r$ after $k$ steps while it remains within $\varepsilon$, and since
$c'>1$ this exceeds $\varepsilon$ for $k$ large; so the orbit leaves that
neighbourhood in finitely many steps. An orbit converging to $\tfrac13\one$ would have to remain
in that ball from some time on, so no orbit converges to $\tfrac13\one$ except the constant one. As
$\tfrac13\one$ is the only fixed point, no other orbit converges at all.
\end{proof}

The proposition therefore shows that the temperature condition of
Section~\ref{sec:case2} is not an artefact of the proof technique: for $\beta<\kappa/(2\sqrt3)$, which
is below $\tfrac12$ for every admissible $\kappa$, the fixed point repels in
every direction and the dynamics genuinely fail to converge. The iterates cycle, the kernel being
cyclic and the agent at low temperature moving almost all its mass onto
whichever response beats the one it currently favours.

One open question is where the boundary lies. On this
instance convergence holds for $\beta>\tfrac12$ by
Proposition~\ref{prop:case2-geom} and fails for $\beta<\kappa/(2\sqrt3)$ by
the proposition above, and the two bounds do not meet: for $\kappa=1$ they
leave the interval $(0.289,0.5)$ uncharacterised. The other is uniqueness: Proposition~\ref{prop:case2-fp}(b) gives a unique fixed point whenever
$\rho(B^{-1}A)<2$, and the instance above shows that below such a condition
the iterates may fail to converge while the fixed point is still unique, so
non-convergence is no evidence of multiplicity. The instance below shows that several fixed points are possible when the
kernels differ, so Proposition~\ref{prop:case2-fp}(d), which rules this out
when they do not, is not vacuous.

\paragraph{Contraction near a strict equilibrium.}
\label{app:local-strict}

The threshold of Proposition~\ref{prop:case2-geom} is a condition for
\eqref{eq:case2} to contract \emph{globally}. It contracts near a strict
equilibrium whatever the temperature, and the lower the temperature the more
strongly. By the limiting game is meant the game the agents play as
$\beta_v\to0$, in which each maximises $R_v(\cdot;\check q_v)$ outright and
best replies are pure.

Write $\check\delta_v=\sum_ua_{vu}\delta_{y_u}$ for the challenger an agent
faces when every other holds the pure response assigned to it. A pure profile
$(y_1,\dots,y_n)$ is a \emph{strict equilibrium with margin $\delta$} if
\[
  R_v(y_v;\check\delta_v)\;\geq\;R_v(y;\check\delta_v)+\delta
  \qquad\text{for every }v\text{ and every }y\neq y_v .
\]
Write $M=\max_v\max_{y,y'}\mu_v(y)/\mu_v(y')$ for the spread of the anchors,
a finite constant since each $\mu_v$ is interior.

\begin{proposition}[Contraction near a strict equilibrium]
\label{prop:local-strict}
Let $(y_v)$ be a strict equilibrium with margin $\delta>0$ and let the agents
hold arbitrary temperatures $\beta_1,\dots,\beta_n>0$. Write
\[
    L(\beta)=\frac{4(|\Y|-1)\,M}{\beta}\,e^{-\delta/2\beta},
  \qquad
    L_{\max}=\max_vL(\beta_v) .
\]
There are $\varepsilon>0$ and $\beta_0>0$, depending only on $\delta$, $M$ and $|\Y|$, such that whenever $\beta_v<\beta_0$ for
every $v$, the set
$N=\{\mathbf q:\ q_v(y_v)\geq1-\varepsilon\ \forall v\}$ satisfies
$T(N)\subseteq N$ and $T$ is a contraction on $N$ with modulus
$L_{\max}<1$. Consequently \eqref{eq:case2} has a unique fixed point in $N$ and converges
to it geometrically from every start in $N$.
\end{proposition}

In $L(\beta)$ the exponential is the mass an agent at temperature $\beta$ leaves
on responses other than $y_v$, and the factor $1/\beta$ comes from dividing
the quality signal by the temperature. The modulus tends to zero as
$\beta\to0$, so the contraction is \emph{sharper} at lower temperatures. Aggressive optimisation removes the global contraction of Proposition~\ref{prop:case2-geom} while keeping the local one. The
restriction $\beta_v<\beta_0$ is needed for the invariance of $N$ rather than
for $L_{\max}<1$, and fails once an agent is cautious enough that its update is
no longer close to a point mass.

\begin{proof}
\emph{Invariance.} Let $\mathbf q\in N$. Each $q_u$ puts mass at least
$1-\varepsilon$ on $y_u$, so $\|q_u-\delta_{y_u}\|_1\leq2\varepsilon$, and
since $\check q_v$ and $\check\delta_v$ are the same convex combination of
these, $\|\check q_v-\check\delta_v\|_1\leq2\varepsilon$. By
Lemma~\ref{lem:signal-lip},
$|R_v(y;\check q_v)-R_v(y;\check\delta_v)|\leq\varepsilon$ for every $y$,
so each of the two terms in the margin moves by at most $\varepsilon$ and
$y_v$ still beats every alternative by at least $\delta-2\varepsilon$, which
is at least $\delta/2$ once $\varepsilon\leq\delta/4$. For the update
$T(\mathbf q)_v\propto\mu_v\exp\bigl(R_v(\cdot;\check q_v)/\beta_v\bigr)$ this
gives, for $y\neq y_v$,
\[
  \frac{T(\mathbf q)_v(y)}{T(\mathbf q)_v(y_v)}
  =\frac{\mu_v(y)}{\mu_v(y_v)}
   \exp\Bigl(\tfrac{1}{\beta_v}\bigl[R_v(y;\check q_v)-R_v(y_v;\check q_v)\bigr]\Bigr)
  \;\leq\;M\,e^{-\delta/2\beta_v} ,
\]
and summing over the $|\Y|-1$ such $y$,
\[
  1-T(\mathbf q)_v(y_v)\;\leq\;T(\mathbf q)_v(y_v)\,(|\Y|-1)\,M\,e^{-\delta/2\beta_v}
  \;\leq\;(|\Y|-1)\,M\,e^{-\delta/2\beta_v},
\]
which is below $\varepsilon$ once $\beta_v<\beta_0$; so $T(N)\subseteq N$. The
estimate is made agent by agent, so the temperatures need not agree.

\emph{Contraction.} Differentiating,
$\mathrm dT_v=\tfrac1{\beta_v}D_vP_vA_{v\cdot}$ with
$D_v=\mathrm{diag}(q_v)-q_vq_v^\top$ the softmax differential. Take a
perturbation $(h_u)$ of the profile with $\|h_u\|_1\leq r$ for every $u$ and
put $z=P_v\sum_ua_{vu}h_u$. The rows of $A$ sum to one, so
$\|\sum_ua_{vu}h_u\|_1\leq r$, and $P_v$ has entries in $[0,1]$, so
$\|z\|_\infty\leq r$ and $\osc(z)\leq2r$. Now
$(D_vz)_y=q_v(y)\bigl(z_y-\E_{q_v}z\bigr)$, so
$\|D_vz\|_1=\E_{q_v}|z-\E_{q_v}z|\leq2\,\E_{q_v}|z-z_{y_v}|$, and since
$z-z_{y_v}$ vanishes at $y_v$ and is at most $\osc(z)$ elsewhere,
$\E_{q_v}|z-z_{y_v}|\leq\bigl(1-q_v(y_v)\bigr)\osc(z)$. On $N$ after one step
$1-q_v(y_v)\leq(|\Y|-1)Me^{-\delta/2\beta_v}$ by the first part, so
\[
  \|D_vz\|_1\;\leq\;4(|\Y|-1)M\,e^{-\delta/2\beta_v}\,r ,
\]
and dividing by $\beta_v$ gives $\|\mathrm dT_v\|\leq L(\beta_v)$. Taking the
maximum over agents bounds $\|\mathrm dT\|$ by $L_{\max}$. Each $L(\beta_v)\to0$ as $\beta_v\to0$, so shrinking $\beta_0$ if necessary
makes $L_{\max}<1$. The set $N$ is a closed subset of the product of simplices, hence complete,
and $T$ maps it into itself by the first part, so Banach's contraction
mapping theorem applies and completes the proof.
\end{proof}

Proposition~\ref{prop:case2-geom} asks for $\beta_v>\tfrac12$ at every agent
and this one for $\beta_v<\beta_0$ at every agent, so the two cover disjoint
regimes and this one applies where the other does not. Read beside
Proposition~\ref{prop:case2-below}, it suggests that what matters at low
temperature is whether the limiting game has a strict equilibrium. Where one
exists, \eqref{eq:case2} has an attracting fixed point near it, so the
dynamics converge from any start close enough, to a limit determined by the
starting profile. Where none exists, as for the cyclic kernel of
Proposition~\ref{prop:case2-below}, whose limiting game is
rock--paper--scissors, Proposition~\ref{prop:local-strict} does not apply, and on that instance the
iterates are shown not to converge at all. Whether the absence
of a strict equilibrium always precludes convergence is not established
here.

\paragraph{Several fixed points below the threshold.}
\label{app:several-fp}

Proposition~\ref{prop:local-strict} attaches an attracting fixed point to each
strict equilibrium of the limiting game, so a game with several strict
equilibria gives \eqref{eq:case2} several fixed points at low temperature.
Such a game exists on two agents and three responses. Take
$\Y=\{0,1,2\}$, a common uniform anchor and a common temperature $\beta$,
attention $a_{vv}=\varepsilon$ and $a_{vu}=1-\varepsilon$ with
$\varepsilon<\tfrac13$, and opposite cyclic kernels
$P_1=\tfrac12\one\one^\top+\tfrac25C$ and $P_2=\tfrac12\one\one^\top-\tfrac25C$
with $C$ the cyclic generator of Proposition~\ref{prop:case2-below}. As
$\beta\to0$ agent~1's best reply to $z$ is $\tau(z)$ and agent~2's to $y$ is
$\tau^{-1}(y)$, where $\tau$ is the cyclic shift, so the pure profiles
$(0,1)$, $(1,2)$ and $(2,0)$ are equilibria, each strict with margin
$\tfrac25$. By Proposition~\ref{prop:local-strict} each has a fixed point of
\eqref{eq:case2} nearby for all small $\beta$, and the three are distinct
because their limits are. These fixed points are the logit equilibria of the
game at noise level $\beta$, in the sense of \citet{mckelvey1995}. At these temperatures $\rho(B^{-1}A)=1/\beta$ far
exceeds the value $2$ that Proposition~\ref{prop:case2-fp}(b) requires, so
that threshold cannot be dropped: below it the fixed point need not be
unique. The attention matrix here is also symmetric, so symmetry alone does not restore uniqueness. Proposition~\ref{prop:case2-fp}(d) needs a shared kernel as well, and the two kernels here differ.

\paragraph{When the threshold can be dropped.}

Proposition~\ref{prop:case2-fp}(b) obtains uniqueness from
$\rho(B^{-1}A)<2$, a condition that makes the update a contraction and so
gives more than uniqueness alone requires. This section identifies what uniqueness requires. Suppose there were two distinct fixed points.
Pairing the difference of their log-beliefs with the difference of the
profiles themselves produces an identity with a definite sign on one side and
a single quantity on the other, so that uniqueness follows whenever that
quantity vanishes. The rest of the section computes it and finds when it does.

So let $\mathbf q^1$ and $\mathbf q^2$ be fixed points, and fix positive weights
$(\omega_v)_{v\in V}$. Taking logarithms in \eqref{eq:case2} at each of them,
subtracting and multiplying by $\beta_v$,
\[
  \beta_v\bigl(\log q^1_v-\log q^2_v\bigr)
  \;=\;R_v(\cdot\,;\check q^1_v)-R_v(\cdot\,;\check q^2_v)
  \;+\;\text{const} ,
\]
the constant being $\beta_v$ times the difference of the two normalisers. Now
take the inner product of each side with $\Delta q_v=q^1_v-q^2_v$. The constant
drops out, because $\Delta q_v$ sums to zero, and multiplying by $\omega_v$ and
summing over the agents leaves
\begin{equation}\label{eq:uniq-identity}
  \sum_v\omega_v\beta_v\bigl\langle\log q^1_v-\log q^2_v,\,\Delta q_v\bigr\rangle\;=\;\Xi_\omega ,
\end{equation}
where
\[
  \Xi_\omega:=\sum_v\omega_v\bigl\langle
  R_v(\cdot\,;\check q^1_v)-R_v(\cdot\,;\check q^2_v),\,\Delta q_v\bigr\rangle .
\]
Each inner product in \eqref{eq:uniq-identity} has the form
$\langle\log a-\log b,\,a-b\rangle$, which is non-negative because $\log$ is
increasing, and zero only when $a=b$, and its weight $\omega_v\beta_v$ is positive.
That sum is therefore strictly positive unless the two profiles coincide. So if
$\Xi_\omega=0$ the sum must vanish, and the two profiles coincide: the fixed point is
unique. The threshold is therefore unnecessary whenever $\Xi_\omega$ vanishes for
some choice of positive weights. The next result evaluates $\Xi_\omega$ and determines exactly when it does.

\begin{lemma}[The coupling term]
\label{lem:M3}
Assume that the agents share a kernel $P$, and let $(\omega_v)_{v\in V}$ be positive
weights. Then for any two profiles,
\begin{equation}\label{eq:M3}
  \Xi_\omega=\frac{1}{2}\sum_{v,u}\bigl(\omega_va_{vu}-\omega_ua_{uv}\bigr)
  \,B_P(\Delta q_v,\Delta q_u),
\end{equation}
with $B_P$ the form of Lemma~\ref{lem:degenerate}.
\end{lemma}

\begin{proof}
By definition,
\[
  \Xi_\omega=\sum_v\omega_v\sum_y
  \bigl[R_v(y;\check q^1_v)-R_v(y;\check q^2_v)\bigr]\,\Delta q_v(y),
  \qquad
  R_v(y;\check q^i_v)=\sum_ua_{vu}\sum_{y'}q^i_u(y')P(y,y'),
\]
so
\[
  \Xi_\omega=\sum_{v,u}\omega_va_{vu}\sum_{y,y'}\Delta q_v(y)\Delta q_u(y')P(y,y')
  =\sum_{v,u}\omega_va_{vu}B_P(\Delta q_v,\Delta q_u).
\]
Relabelling $(v,u)\to(u,v)$ in this sum and averaging with the original,
\[
  \begin{aligned}
  \Xi_\omega&=\frac{1}{2}\sum_{v,u}
  \bigl[\omega_va_{vu}B_P(\Delta q_v,\Delta q_u)+\omega_ua_{uv}B_P(\Delta q_u,\Delta q_v)\bigr]\\
  &=\frac{1}{2}\sum_{v,u}(\omega_va_{vu}-\omega_ua_{uv})B_P(\Delta q_v,\Delta q_u),
  \end{aligned}
\]
by the antisymmetry of $B_P$ established in Lemma~\ref{lem:degenerate}. That
lemma applies because each $\Delta q_v$ is a difference of two distributions
and so lies in $\Y_0$.
\end{proof}

\begin{proposition}[When the coupling term vanishes]
\label{prop:M3}
Assume that the agents share a kernel $P$, and let $(\omega_v)_{v\in V}$ be positive
weights. Then $\Xi_\omega=0$ for every pair of profiles if and only if
$\omega_va_{vu}=\omega_ua_{uv}$ for every pair $u,v$ or $P$ is degenerate.
\end{proposition}

\begin{proof}
By Lemma~\ref{lem:M3}, $\Xi_\omega$ is given by \eqref{eq:M3}. If
$\omega_va_{vu}=\omega_ua_{uv}$ for every pair, every coefficient $\omega_va_{vu}-\omega_ua_{uv}$
there vanishes, and if $P$ is degenerate then $B_P\equiv0$ on $\Y_0$ by
Lemma~\ref{lem:degenerate}(i); either way $\Xi_\omega=0$ for all profiles.

Conversely suppose that $\omega_{v_0}a_{v_0u_0}\neq \omega_{u_0}a_{u_0v_0}$ for some pair,
necessarily with $u_0\neq v_0$, and that $P$ is not degenerate. Pick, by
non-degeneracy, $d,d'\in\Y_0$ with $B_P(d,d')\neq0$. Take $\Delta q_{v_0}=\epsilon d$,
$\Delta q_{u_0}=\epsilon d'$ and $\Delta q_v=0$ otherwise. All terms of
\eqref{eq:M3} with $v=u$ vanish, since $B_P(\Delta q_v,\Delta q_v)=0$ by
antisymmetry, so only the pair $(v_0,u_0)$ and its transpose survive and
\[
  \Xi_\omega=\epsilon^2\,(\omega_{v_0}a_{v_0u_0}-\omega_{u_0}a_{u_0v_0})\,B_P(d,d')\neq0 .
\]
For $\epsilon$ small enough each $\Delta q_v$ is a difference of two
profiles, so this exhibits profiles with $\Xi_\omega\neq0$.
\end{proof}

Reversibility of the weights enters this paper only here and in
Proposition~\ref{prop:case2-fp}(d), which rests on the same computation. With
uniform weights it is symmetry, $a_{vu}=a_{uv}$. Either condition constrains each
pair of agents, whereas the double stochasticity usually imposed in
decentralised optimisation fixes only that $\pi$ be uniform.

Combining the identity with the two cases in which the coupling term vanishes
gives the uniqueness statement the section was aiming at.

\begin{corollary}[Uniqueness without a temperature condition]
\label{cor:sym-unique}
Assume that the agents share a kernel, and that either $A$ is reversible,
$\omega_va_{vu}=\omega_ua_{uv}$ for some positive weights $(\omega_v)_{v\in V}$, or the kernel is
degenerate. Then \eqref{eq:case2} has a unique fixed point for every vector of
temperatures. Since
every kernel on two responses is degenerate, this covers every binary
problem.
\end{corollary}

\begin{proof}
Under the first hypothesis $\Xi_\omega=0$ for those weights, and under the second
for any positive weights, by Proposition~\ref{prop:M3}; the identity
\eqref{eq:uniq-identity} then forces two fixed points to coincide, as shown
there. The first case is Proposition~\ref{prop:case2-fp}(d). Every kernel on two responses is degenerate by
Lemma~\ref{lem:degenerate}, which gives the binary case.
\end{proof}

Of the two hypotheses, only one is known to be necessary. The shared kernel
is: Appendix~\ref{app:several-fp} exhibits two agents
on a \emph{symmetric} graph holding opposite cyclic kernels on three
responses, with at least three fixed points for all small $\beta$, one near
each strict equilibrium of the limiting game. It is therefore heterogeneity
of the kernels, and not asymmetry of the attention, that defeats uniqueness
there. Whether reversibility is necessary is not known.
Proposition~\ref{prop:M3} shows only that without it, and with a non-degenerate kernel, every choice of positive weights leaves some pair of profiles with $\Xi_\omega\neq0$, which closes this route to uniqueness
without ruling uniqueness out, so whether the threshold is genuinely needed on a
non-reversible graph with a non-degenerate kernel remains open.

Uniqueness and convergence are separate matters, and this corollary gives
only the first. The instance of Proposition~\ref{prop:case2-below} is a single
agent with a cyclic kernel, so the corollary applies and the uniform
distribution is the unique fixed point; yet that fixed point is repelling for
every $\beta<\kappa/(2\sqrt3)$, and the iterates converge from no starting
belief but the fixed point itself. Convergence on a directed graph with a
non-degenerate kernel still needs Proposition~\ref{prop:case2-geom}, and so
the threshold.

\section{Section~\ref{sec:case4}: both roles aggregated}

This section collects supporting material for Section~\ref{sec:case4}:
interior fixed points and their stability, the conditions under which an atom
attracts, and further results for particular kernels, boundaries and graphs.

\subsection{Interior fixed points: existence and stability}
\label{supp:interior}

Whether an interior fixed point exists at all, and whether the dynamics stay
near one, are separate questions. The second is settled here by linearisation:
the update \eqref{eq:case4} is analytic, so the behaviour near a fixed point is
governed by the Jacobian there, acting on the log-beliefs modulo the constants
that normalisation removes. A spectral radius below one gives local asymptotic
stability, and an eigenvalue of modulus above one gives instability.

\begin{proposition}[Interior fixed points: existence and stability]
\label{prop:case4-interior}
Assume that Assumption~\ref{ass:graph} holds, and write
$P_v=\tfrac12\one\one^\top+S_v$, so that $S_v$ is the skew part of agent $v$'s
kernel. 
\begin{enumerate}[label=(\alph*),leftmargin=*]
\item If $S_v\one=0$ for every agent, the uniform consensus profile
  $q^*_v\equiv(1/|\Y|,\dots,1/|\Y|)$ is an interior fixed point of
  \eqref{eq:case4}, for every graph and every vector of temperatures.
\item If the agents share a kernel and a temperature, every interior
  consensus fixed point is unstable, at every temperature, unless the kernel is
  the constant kernel $\tfrac12\one\one^\top$, under which every comparison is a
  tie: the Jacobian has an
  eigenvalue of modulus greater than one, so no neighbourhood of the fixed
  point is invariant.
\item There are instances with a locally asymptotically stable interior fixed
  point, already with two agents and three responses.
\end{enumerate}
\end{proposition}

\paragraph{Proof of Proposition~\ref{prop:case4-interior}.}

Throughout, the update is \eqref{eq:case4},
\[
  q_{v,t+1}(y)\;\propto\;\bar q_{v,t}(y)\,
  \exp\Bigl(\tfrac{1}{\beta_v}R_v\bigl(y;\check q_{v,t}\bigr)\Bigr),
  \qquad
  \bar q_{v,t}\propto\prod_u q_{u,t}^{a_{vu}},
  \quad \check q_{v,t}=\sum_u a_{vu}q_{u,t},
\]
with $R_v(y;q)=\sum_{y'}q(y')P_v(y,y')$. In log-beliefs it reads
$\mathbf f_{t+1}(y)=A\mathbf f_t(y)+B^{-1}\mathbf R_t(y)-\boldsymbol\Phi_t$,
whose rest points are the solutions of
\[
  (I-A)\,\mathbf f^*(y)=B^{-1}\mathbf R^*(y)-\boldsymbol\Phi^* ,
\]
the $\boldsymbol\Phi$ being the $y$-independent log-normalisers.

\emph{Part (a).} Let $q^*_v\equiv(1/|\Y|,\dots,1/|\Y|)$. Every challenger is
then a convex combination of uniform distributions, so $\check q^*_v$ is
uniform, and
\[
  R^*_v(y)=\frac{1}{|\Y|}\sum_{y'}P_v(y,y')
  =\tfrac12+\frac{1}{|\Y|}\,(S_v\one)_y=\tfrac12
\]
by the hypothesis $S_v\one=0$, the same value at every response. The reference is also uniform, being a weighted geometric mean of uniform
distributions, so writing $T_v(\mathbf q^*)$ for the right-hand side of \eqref{eq:case4} at
this profile,
\[
  T_v(\mathbf q^*)(y)\;\propto\;\frac{1}{|\Y|}\,e^{1/(2\beta_v)} ,
\]
the same value at every $y$. A distribution proportional to a constant is
uniform, so $T_v(\mathbf q^*)=q^*_v$ at every agent and the uniform profile is
a fixed point, whatever $A$ and the temperatures. The constant
$e^{1/(2\beta_v)}$ is removed by normalisation, which is why no condition on
$\beta_v$ appears.

\emph{The Jacobian, used in parts~(b) and~(c).} Let $q^*$ be any interior
consensus fixed point and perturb the log-beliefs by $\delta f_v$; the Jacobian is a composition of three derivatives. First, $q_v=\operatorname{softmax}(f_v)$ has derivative
$\partial q_v(y)/\partial f_v(y')=q^*(y)\bigl(\delta_{yy'}-q^*(y')\bigr)$, so
$\delta q_v=J\delta f_v$ with
$J=\operatorname{diag}(q^*)-q^*q^{*\top}$, symmetric and positive
semidefinite, with $J\one=0$ and hence $\one^\top J=0$.

Second, the challenger is linear in the beliefs,
$\check q_v=\sum_u a_{vu}q_u$, so $\delta\check q_v=\sum_u a_{vu}J\delta f_u$.

Third, the signal is linear in the challenger, $R_v=P_v\check q_v$, so
$\delta R_v=P_v\sum_u a_{vu}J\delta f_u$.

Row $v$ of the log-space update reads
\[
  f_{v,t+1}(y)=\sum_ua_{vu}f_{u,t}(y)+\frac{1}{\beta_v}R_v(y;\check q_{v,t})
  -\Phi_{v,t},
\]
where
\[
  \Phi_{v,t}=\log\sum_{y'}\exp\Bigl(\sum_ua_{vu}f_{u,t}(y')
  +\tfrac{1}{\beta_v}R_v(y';\check q_{v,t})\Bigr)
\]
is the log-normaliser, one number for agent $v$ and not a function of $y$. Perturbing each term in turn, the
first is linear in $\mathbf f$ and contributes
$\sum_ua_{vu}\,\delta f_{u,t}$; the second contributes
$\tfrac{1}{\beta_v}\delta R_v=\tfrac{1}{\beta_v}P_v\sum_ua_{vu}J\delta f_{u,t}$
by the third step above; and the third contributes $-\delta\Phi_{v,t}$ at
every $y$, that is $-\delta\Phi_{v,t}\one$ as a vector over $\Y$, because
$\Phi_{v,t}$ does not depend on $y$.

The normaliser depends on the beliefs, so $\delta\Phi_{v,t}$ is not zero, and
it contributes the multiple $-\delta\Phi_{v,t}\one$ of $\one$ to
$\delta f_{v,t+1}$. Such multiples have no effect on the beliefs, now or later. A perturbation
that is $c_v\one$ at agent $v$ leaves every belief unchanged, since
$\delta q_v=J\delta f_v=c_vJ\one=0$; and it stays of that form at the next
round, since the first term above returns $\sum_ua_{vu}c_u\one$ and the second
returns $0$. The multiples of $\one$ thus span a subspace on which the
linearised dynamics act without ever affecting a belief, and the eigenvalues
on it are those of $A$, carrying no information about stability. Discarding
the normaliser restricts attention to a complement of that subspace, and there
\[
  \delta f_{v,t+1}
  =\sum_u a_{vu}\,\delta f_{u,t}+\frac{1}{\beta_v}P_v\sum_u a_{vu}J\delta f_{u,t}
  =\sum_u a_{vu}\Bigl(I+\frac{P_vJ}{\beta_v}\Bigr)\delta f_{u,t} ,
\]
so the Jacobian is a block matrix whose $(v,u)$ block is
\begin{equation}\label{eq:jacblocks}
  M_{vu}=a_{vu}\Bigl(I+\frac{P_vJ}{\beta_v}\Bigr) .
\end{equation}
Writing $P_v=\tfrac12\one\one^\top+S_v$ and using $\one^\top J=0$ gives
$P_vJ=S_vJ$, so the constant part $\tfrac12\one\one^\top$ contributes nothing
to the Jacobian and only $S_v$ appears in \eqref{eq:jacblocks}. This is the
linearised form of the fact that adding a constant to every entry of a kernel
shifts each agent's quality signal by the same amount at every response, which
normalisation removes.

\emph{Part (b).} Suppose $P_v\equiv P$ and $\beta_v\equiv\beta$. The block in
\eqref{eq:jacblocks} is then the same for every $v$, so
$M=A\otimes K$ with $K=I+SJ/\beta$, and the eigenvalues of a Kronecker product
are the pairwise products, $\lambda_i(A)\mu_j(K)$.

The eigenvalues of $K$ are $1+s_j/\beta$ with $s_j$ running over those of $SJ$.
Each $s_j$ is purely imaginary. To see this, let $J^{1/2}$ be the symmetric
positive semidefinite square root of $J$. The matrices
$SJ=(SJ^{1/2})J^{1/2}$ and $J^{1/2}SJ^{1/2}=J^{1/2}(SJ^{1/2})$ are products of
the same two factors in the two orders, so they have the same non-zero
eigenvalues. And $J^{1/2}SJ^{1/2}$ is real and skew, since
$(J^{1/2}SJ^{1/2})^\top=J^{1/2}S^\top J^{1/2}=-J^{1/2}SJ^{1/2}$, so its
eigenvalues are purely imaginary.

Hence $|\mu_j|=|1+s_j/\beta|=\sqrt{1+|s_j|^2/\beta^2}\geq1$, with equality only
when $s_j=0$. It remains to show that some $s_j$ is non-zero, that is, that
$J^{1/2}SJ^{1/2}\neq0$, since a non-zero real skew matrix has a non-zero
eigenvalue. At an interior $q^*$ the kernel of $J$ is spanned by $\one$, so $J$
and $J^{1/2}$ map $\mathbb{R}^\Y$ onto $\Y_0$, and $J^{1/2}SJ^{1/2}=0$ exactly
when $B_S$ vanishes on $\Y_0$, that is, when the kernel is degenerate in the
sense of Lemma~\ref{lem:degenerate}. A degenerate kernel has
$P(y,y')=\tfrac12+c(y)-c(y')$, so $R(y;q^*)-R(y';q^*)=c(y)-c(y')$, and
Proposition~\ref{prop:case4-fp}(c) then admits an interior consensus fixed
point only when $c$ is constant, that is, when $P=\tfrac12\one\one^\top$, the
kernel excluded in the statement. For every other kernel some $s_j\neq0$.
Since $A$ is row-stochastic, $\lambda=1$ is among its eigenvalues, so the
products $\lambda_i(A)\mu_j(K)$ include $\mu_j$ itself and
$\rho(M)\geq\max_j|\mu_j|>1$. An eigenvalue of modulus greater than one makes the fixed point unstable. The temperature enters only through $|s_j|/\beta$, so raising it
drives the modulus towards $1$ from above and never below.

\emph{Part (c).} Take two agents and $\Y=\{0,1,2\}$, with
\[
  A=\begin{pmatrix}3/5&2/5\\[2pt]1/3&2/3\end{pmatrix},
  \qquad \beta_1=\tfrac13,\quad\beta_2=3,
  \qquad P_1=\tfrac12\one\one^\top+\tfrac13C,
  \quad P_2=\tfrac12\one\one^\top-\tfrac12C,
\]
where $C$ is the cyclic generator of the Supplementary Material,
$C_{y,y+1}=1$ and $C_{y,y-1}=-1$ modulo three. Both $P_v$ have entries in
$[0,1]$ and satisfy $P_v+P_v^\top=\one\one^\top$, so they are preference
kernels, and $S_v=P_v-\tfrac12\one\one^\top$ is a multiple of $C$, so
$S_v\one=0$. The uniform profile is therefore a fixed point by~(a), and $J=\tfrac13(I-\tfrac13\one\one^\top)$ gives
$S_vJ=S_v/3$. Adding a constant to an agent's log-beliefs leaves its belief
unchanged, so only the action of $M$ on $\{w:\one^\top w=0\}$ matters. Taking
$u_1=(1,-1,0)$ and $u_2=(0,1,-1)$ as a basis of that subspace at each agent and
writing $U=[u_1\;u_2]$, the blocks of \eqref{eq:jacblocks} become
$a_{vu}(U^\top U)^{-1}U^\top(I+S_v/(3\beta_v))U$, giving
\[
  M=\begin{pmatrix}
    2/5 & 2/5 & 4/15 & 4/15\\
    -2/5 & 4/5 & -4/15 & 8/15\\
    19/54 & -1/27 & 19/27 & -2/27\\
    1/27 & 17/54 & 2/27 & 17/27
  \end{pmatrix},
\]
Local asymptotic stability holds when every eigenvalue of $M$ has modulus
below one, so it must be shown that the four roots of the characteristic
polynomial lie in the open unit disc. Clearing denominators, that polynomial
is
\[
  p(\lambda)=18225\lambda^4-46170\lambda^3+40953\lambda^2-14316\lambda+1744 .
\]

\begin{figure}[t]
\centering
\includegraphics[width=0.34\textwidth]{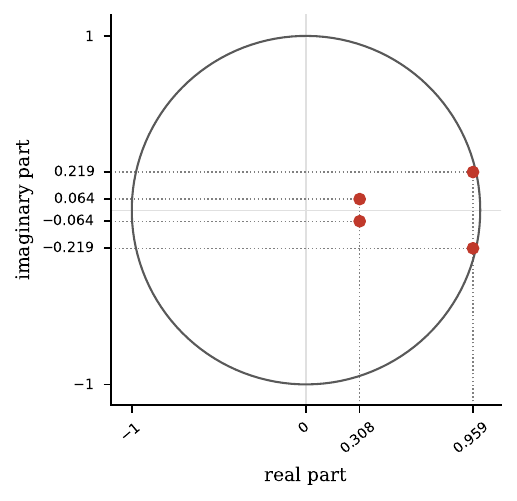}
\caption{The four eigenvalues of the Jacobian $M$ at the uniform fixed point
of the instance in the proof of Proposition~\ref{prop:case4-interior}(c),
shown with the unit circle, all strictly inside it.}
\label{fig:eigs}
\end{figure}

Its roots are $0.959\pm0.219i$ and $0.308\pm0.064i$, of moduli $0.983$ and
$0.315$, so $\rho(M)<1$ and the fixed point is locally asymptotically stable.
Figure~\ref{fig:eigs} places all four in the unit disc, with their real and
imaginary parts marked on the axes; the leading pair lies close to the
boundary, so the contraction near this fixed point is slow.

 These roots were found numerically, but the conclusion does not depend on
that. The substitution $\lambda=(1+z)/(1-z)$ maps the half-plane
$\{\operatorname{Re}z<0\}$ onto the open unit disc, so a root $\lambda$ of $p$
lies in the disc exactly when the corresponding
$z=(\lambda-1)/(\lambda+1)$ has negative real part. Substituting turns
$p(\lambda)$ into a rational function of $z$ whose denominator is $(1-z)^4$,
since $p$ has degree four; clearing it gives the polynomial
\[
  q(z)=(1-z)^4\,p\bigl((1+z)/(1-z)\bigr)
  =121408z^4+129632z^3+37908z^2+2216z+436 ,
\]
whose roots are exactly the $z$ corresponding to roots of $p$. Its leading
coefficient is $p(-1)=121408$, which is non-zero, so $q$ has degree four and
the four roots match up one for one. The question is therefore whether all
four roots of $q$ have negative real part. For a real polynomial that is decided by the signs
of finitely many determinants formed from its coefficients, the
Routh--Hurwitz criterion~\citep{gantmacher1959}; for a quartic with positive leading coefficient they are the four leading principal minors of its Hurwitz matrix, here $\Delta_1=129\,632$, $\Delta_2=4\,645\,049\,728$, $\Delta_3=2\,966\,687\,632\,384$ and $\Delta_4=436\,\Delta_3$, all positive, so every root of $q$ has negative real part and every root of $p$ lies in the disc.
\proofend

\paragraph{Fixed points at which the agents disagree.}
All three parts of Proposition~\ref{prop:case4-interior} concern consensus
fixed points. Interior fixed points at which the agents hold different
beliefs also exist, which shows that the tilts in
Proposition~\ref{prop:case4-spread}(c) do occur. Take any interior profile
$\mathbf q^*$ and write $w_v=\check q^*_v$ for agent $v$'s challenger. By
Theorem~\ref{thm:fredholm}, $\mathbf q^*$ is a fixed point exactly when each
$B^{-1}\mathbf R^*_v$ equals $[(I-A)\mathbf f^*]_v$ up to a constant, and since
$R^*_v=\tfrac12\one+S_vw_v$ this asks for a skew matrix $S_v$ with
$S_vw_v=t_v$, where
$t_v=\beta_v[(I-A)\mathbf f^*]_v-\bigl(w_v^\top\beta_v[(I-A)\mathbf f^*]_v\bigr)\one$.
The subtraction makes $w_v^\top t_v=0$, which a skew matrix requires, as
$w^\top Sw=0$ for every $w$. Then $S_v=t_vu_v^\top-u_vt_v^\top$ with
$u_v=w_v/\|w_v\|_2^2$ is skew and satisfies $S_vw_v=t_v$, and
$P_v=\tfrac12\one\one^\top+S_v$ is a kernel provided every entry of $S_v$ lies
in $[-\tfrac12,\tfrac12]$. That holds whenever the agents' log-beliefs are
close enough to one another that $\|t_v\|_\infty$ is small, so every interior
profile sufficiently close to consensus is a fixed point for some choice of
kernels, in general distinct. The construction does not show whether such fixed points are stable, or whether they arise from kernels induced by
rewards.

\subsection{Convergence to a Condorcet winner}
\label{supp:condorcet}

The agents cannot agree in the interior unless \eqref{eq:unanimity} holds,
which the count above makes exceptional, and
Theorem~\ref{thm:case4-directed} bounds how far apart they are without locating a limit, which leaves two possibilities. The agents may converge together to a single response,
in which case the last iterate shows what the network implements; or the
beliefs may fail to converge, in which case the collective rule appears only
in the time average, treated in Section~\ref{sec:cesaro}; this subsection treats the first. Proposition~\ref{prop:case4-fp}(b) leaves
$|\Y|$ atoms as candidates, and whenever the influence-weighted aggregate
$\bar P$ compares every pair strictly, at most one of them
attracts: the strict Condorcet winner of $\bar P$ if there is
one, with every other atom repelling, and none at all when the comparison
cycles. Three conditions below establish this, each weaker than the last.
Where the aggregate ties for some pair, none of the three applies.

Throughout, fix $y^*\in\Y$ and write
$\phi_{v,t}(y^*,y)=\log\bigl(q_{v,t}(y^*)/q_{v,t}(y)\bigr)$ for the log-odds
of $y^*$ against $y$ at agent $v$.

\begin{proposition}[Unanimous dominance: global convergence]
\label{prop:case4-atom}
Consider the dynamics \eqref{eq:case4}, and suppose there is $\delta>0$ with
\[
  P_v(y^*,y')-P_v(y,y')\;\geq\;\delta
  \qquad\text{for every agent }v,\text{ every }y\neq y^* \text{ and every }y' .
\]
Then for every initialisation, every $y\neq y^*$ and every $t$,
\[
  \min_v\phi_{v,t}(y^*,y)\;\geq\;\min_v\phi_{v,0}(y^*,y)
  +\frac{t\,\delta}{\beta_{\max}} ,
\]
and hence, at every agent $v$,
\[
  q_{v,t}(y)\;\leq\;\exp\Bigl(-\min_u\phi_{u,0}(y^*,y)-\frac{t\,\delta}{\beta_{\max}}\Bigr),
\]
so $q_{v,t}\to\delta_{y^*}$ at every agent, geometrically. Neither Assumption~\ref{ass:graph},
nor any other condition on the graph, nor a temperature threshold is
required.
\end{proposition}

\begin{proposition}[Aggregate dominance: global convergence]
\label{prop:case4-aggdom}
Assume that Assumption~\ref{ass:graph} holds and that there are $y^*\in\Y$ and
$\delta>0$ with
\begin{equation}\label{eq:aggdom}
  \bar P(y^*,y')\;\geq\;\bar P(y,y')+\delta
  \qquad\text{for every }y\neq y^*\text{ and every }y' ,
\end{equation}
that is, $y^*$ beats every other response by $\delta$ in every column of the
influence-weighted kernel. Assume further that \eqref{eq:case4-threshold} holds, so that $\gamma<1$, and
that $\delta>2\bar{\mathcal D}$, where $\bar{\mathcal D}$ is the residual
disagreement of Theorem~\ref{thm:case4-directed}. The margin $\delta$ must therefore exceed twice the residual, which bounds the
disagreement the agents retain in the limit and is $O(1/\beta_{\min})$ for
identical kernels, so the condition weakens as the
agents become more cautious or more alike. Then for every initialisation, every agent $v$, every $y\neq y^*$ and every
$t$,
\[
  \phi_{v,t}(y^*,y)\;\geq\;\pi^\top\boldsymbol\phi_0(y^*,y)
  +\eta\bigl(\delta-2\bar{\mathcal D}\bigr)\,t
  -\frac{2\eta\,\mathcal D_0}{1-\gamma}
  -2\bigl(\gamma^{\,t}\mathcal D_0+\bar{\mathcal D}\bigr),
\]
where $\boldsymbol\phi_0(y^*,y)$ is the vector of the agents' initial log-odds
and $\mathcal D_0$ the initial disagreement of Theorem~\ref{thm:case4-directed}.
Since $q_{v,t}(y)\leq e^{-\phi_{v,t}(y^*,y)}$, every belief in a response other
than $y^*$ decays geometrically, and $q_{v,t}\to\delta_{y^*}$ at every agent.
\end{proposition}

 The last hypothesis asks the margin to exceed the disagreement the agents
settle at. It limits how much caution is needed rather than which instances
are covered: raising every agent's temperature by a common factor leaves the margin
untouched, since it leaves $\bar P$ unchanged, while driving the disagreement to
zero, so the hypothesis holds once
the agents are cautious enough. It is also far from necessary, bounding a
defect that collapses as the agents concentrate. Checking it on a given
instance needs every agent's kernel and temperature, where the conclusions of
Sections~\ref{sec:case3} and~\ref{sec:case2} need only the graph.

\begin{proposition}[Influence-weighted Condorcet winner: attraction and
repulsion]
\label{prop:case4-condorcet}
Assume that Assumption~\ref{ass:graph} holds, let $y^*\in\Y$, and write
$\mathrm{m}(y^*,y)=\eta\bigl[\bar P(y^*,y)-\tfrac12\bigr]$ for the
influence-weighted margin of $y^*$ over $y$, with $\eta=\sum_v\pi_v/\beta_v$ as
in \eqref{eq:pbar}. If $y^*$ is a strict Condorcet winner of $\bar P$, that is
\[
  \mathrm{m}(y^*,y)>0 \qquad\text{for every }y\neq y^* ,
\]
then the atom $\delta_{y^*}$ is locally attracting, and for
initialisations in its basin
\[
  \phi_{v,t}(y^*,y)=t\,\mathrm{m}(y^*,y)+\bigl[D\,\mathbf d(y)\bigr]_v+O(1),
  \qquad d_v(y)=\frac{P_v(y^*,y)-\tfrac12}{\beta_v},
\]
the $O(1)$ being bounded uniformly in $t$ but not in the initialisation.
Explicitly, let $\mathrm m_0=\min_{y\neq y^*}\mathrm m(y^*,y)$, let
$C\geq\bigl|[D\,\mathbf d(y)]_v\bigr|$ for every $v$ and $y$, and let $L_\epsilon$
satisfy $2(|\Y|-1)e^{-L_\epsilon}\leq\beta_{\min}\mathrm m_0/2$. Every initialisation
with $\phi_{v,0}(y^*,y)\geq L\geq L_\epsilon+2C$ for every $v$ and every
$y\neq y^*$ lies in the basin: then
$\phi_{v,t}(y^*,y)\geq L-2C+t\,\mathrm m_0/2$ for every $t$, and the $O(1)$ term is
at most
\[
  X_0+\frac{K}{1-e^{-\mathrm m_0/2}},
  \qquad
  X_0=\max_{v,\,y\neq y^*}\bigl|\phi_{v,0}(y^*,y)-[D\,\mathbf d(y)]_v\bigr|,
  \quad K=\frac{2(|\Y|-1)}{\beta_{\min}}\,e^{-(L-2C)},
\]
in absolute value. If instead $\mathrm m(y^*,y_0)=-\mathrm m_1<0$ for some $y_0$,
the atom $\delta_{y^*}$ is repelling: no trajectory keeps every log-odds
$\phi_{v,t}(y^*,y)$ at least $L_\epsilon$ for all $t$, when
$2(|\Y|-1)e^{-L_\epsilon}\leq\beta_{\min}\mathrm m_1/2$.
\end{proposition}

Propositions~\ref{prop:case4-atom} and~\ref{prop:case4-condorcet} require no temperature condition and so apply below the threshold
\eqref{eq:case4-threshold} as well as above it;
Proposition~\ref{prop:case4-aggdom} needs it, since it uses the contraction
of the disagreement.

They are one requirement on three
different kernels, each weaker than the last.
Proposition~\ref{prop:case4-atom} asks $y^*$ to dominate every column of every
$P_v$; \eqref{eq:aggdom} asks the same of $\bar P$ alone;
Proposition~\ref{prop:case4-condorcet} asks only for its diagonal case,
$\bar P(y^*,y)>\tfrac12$. Averaging the first with weights
$\pi_v/\beta_v$ gives the second, and setting $y'=y$ there gives the third.
Neither implication reverses: an agent may prefer a different response provided the others outweigh it, and winning one column implies nothing about the rest. Unanimous dominance, which for
Bradley--Terry kernels means every agent's reward is maximised at $y^*$,
gives convergence from any start with the graph playing no part; the
Condorcet condition lets agents disagree provided they are outvoted, and
gives attraction only from nearby. Since
Proposition~\ref{prop:case4-condorcet} covers both signs of the margin, its hypothesis is close to necessary for an atom to attract, and
unanimous dominance adds enough to make the basin the whole simplex.

Where an atom attracts, no interior consensus profile competes with it: such
a profile would be a maximal lottery of $\bar P$, and a strict
Condorcet winner leaves $\delta_{y^*}$ as the only one. Fixed points at which
the agents hold different policies are not covered by this argument. Where
the aggregate cycles the interior remains available, and
Proposition~\ref{prop:case4-interior} describes what becomes of it; the profile
can then approach the boundary while moving between vertices indefinitely,
which is the behaviour Section~\ref{sec:cesaro} addresses through the time
average.

\paragraph{Proof of Proposition~\ref{prop:case4-atom}.}

Fix a response $y\neq y^*$ and write $\boldsymbol\phi_t$ for the vector whose
entries are
$\phi_{v,t}(y^*,y)=\log\bigl(q_{v,t}(y^*)/q_{v,t}(y)\bigr)$. In log-space the
dynamics \eqref{eq:case4} read
$\tilde{\mathbf f}_{t+1}(y)=A\tilde{\mathbf f}_t(y)+B^{-1}\mathbf R_t(y)$,
the first line of \eqref{eq:logspace-vec}: each agent averages its
neighbours' unnormalised log-beliefs and adds its own quality signal, scaled
by its temperature. Here $A$ is the attention matrix, $B=\mathrm{diag}(\beta_v)$ the temperatures
and $R_{v,t}(y)=R_v(y;\check q_{v,t})$ the quality signal agent $v$ receives
against its challenger $\check q_{v,t}=\sum_ua_{vu}q_{u,t}$. Differencing at
$y^*$ and $y$ removes the normalisers and gives
\[
  \boldsymbol\phi_{t+1}=A\boldsymbol\phi_t+\mathbf d_t,
  \qquad
  d_{v,t}=\frac{R_v\bigl(y^*;\check q_{v,t}\bigr)-R_v\bigl(y;\check q_{v,t}\bigr)}{\beta_v} ,
\]
where $\mathbf d_t$ is the vector of the $d_{v,t}$ across agents: the signal's
preference for $y^*$ over $y$ at each agent, scaled by that agent's
temperature. Since $R_v(y;q)=\sum_{y'}q(y')P_v(y,y')$ is linear in its second
argument and $\check q_{v,t}$ is a probability vector,
\[
  R_v(y^*;\check q_{v,t})-R_v(y;\check q_{v,t})
  =\sum_{y'}\check q_{v,t}(y')\bigl[P_v(y^*,y')-P_v(y,y')\bigr]\;\geq\;\delta
\]
by hypothesis, so $d_{v,t}\geq\delta/\beta_{\max}$ at every agent. As $A$ is
non-negative with rows summing to one, $(A\boldsymbol\phi_t)_v$ is a convex
combination of the $\phi_{u,t}$ and so at least $\min_u\phi_{u,t}$. Hence
\[
  \min_v\phi_{v,t+1}\;\geq\;\min_v\phi_{v,t}+\frac{\delta}{\beta_{\max}} ,
\]
and iterating gives the display in the statement.

For the convergence, $q_{v,t}(y)=q_{v,t}(y^*)\,e^{-\phi_{v,t}}\leq e^{-\phi_{v,t}}\leq e^{-\min_u\phi_{u,t}}$,
since $q_{v,t}(y^*)\leq1$, and the display bounds $\min_u\phi_{u,t}$ from below,
which gives the second display of the statement. Summing over the finitely many
$y\neq y^*$ gives $1-q_{v,t}(y^*)\to0$, so $q_{v,t}\to\delta_{y^*}$.
\proofend

\paragraph{Proof of Proposition~\ref{prop:case4-aggdom}.}

Fix $y\neq y^*$ and write $\phi_{v,t}=\phi_{v,t}(y^*,y)$, as in the proof of
Proposition~\ref{prop:case4-atom}. The same differencing of
\eqref{eq:logspace-vec} gives
\[
  \boldsymbol\phi_{t+1}=A\boldsymbol\phi_t+\mathbf d_t,
  \qquad
  d_{v,t}=\frac{R_v(y^*;\check q_{v,t})-R_v(y;\check q_{v,t})}{\beta_v} .
\]
Proposition~\ref{prop:case4-atom} assumed every agent to prefer $y^*$, which
made every $d_{v,t}$ positive. Here \eqref{eq:aggdom} constrains only the
influence-weighted kernel $\bar P$, so individual agents may prefer $y$ and
the argument must proceed through a weighted average of the $d_{v,t}$. Applying $\pi^\top$ and using
$\pi^\top A=\pi^\top$,
\[
  \pi^\top\boldsymbol\phi_{t+1}=\pi^\top\boldsymbol\phi_t+\pi^\top\mathbf d_t ,
\]
so it suffices to bound $\pi^\top\mathbf d_t$ from below, and then to show that
no agent falls far behind the average.

\emph{Step 1: a lower bound on $\pi^\top\mathbf d_t$.} Write
$c_v(y')=P_v(y^*,y')-P_v(y,y')$ for agent $v$'s preference for $y^*$ over $y$
when the challenger puts all its mass on $y'$, and decompose each challenger
as $\check q_{v,t}=p_t+e_{v,t}$, where $p_t$ is the collective policy of
Theorem~\ref{thm:cesaro} and $e_{v,t}$ is agent $v$'s departure from it. Since
$R_v(y;\check q)=\sum_{y'}\check q(y')P_v(y,y')$,
\[
  \pi^\top\mathbf d_t
  =\underbrace{\sum_{y'}p_t(y')\sum_v\frac{\pi_v}{\beta_v}c_v(y')}_{\text{common part}}
  \;+\;\underbrace{\sum_v\frac{\pi_v}{\beta_v}\sum_{y'}e_{v,t}(y')c_v(y')}_{\text{defect}} .
\]
For the common part, $\sum_v(\pi_v/\beta_v)P_v=\eta\bar P$ by \eqref{eq:pbar},
so the inner sum is $\eta\bigl[\bar P(y^*,y')-\bar P(y,y')\bigr]\geq\eta\delta$
at every $y'$ by \eqref{eq:aggdom}, and $p_t$ is a probability vector, so the
common part is at least $\eta\delta$. This is where a single distribution
shared by all agents is needed: \eqref{eq:aggdom} constrains $\bar P$, and
only a common challenger lets the weighted sum of the $P_v$ be assembled. For
the defect, $c_v$ is a difference of two numbers in $[0,1]$, so $|c_v|\leq1$
and each inner sum is at most $\|e_{v,t}\|_1$; summing, the defect is at most
$\sum_v(\pi_v/\beta_v)\|e_{v,t}\|_1=\eta\Delta_t$, the aggregation defect of
Theorem~\ref{thm:cesaro}. Hence
$\pi^\top\mathbf d_t\geq\eta(\delta-\Delta_t)$.

\emph{Step 2: summing over the rounds.} By
Proposition~\ref{prop:defect}, $\Delta_t\leq2\mathcal D_t$, and by
Theorem~\ref{thm:case4-directed} with \eqref{eq:case4-threshold},
$\mathcal D_t\leq\gamma^{\,t}\mathcal D_0+\bar{\mathcal D}$. With Step~1,
\[
  \pi^\top\boldsymbol\phi_{t+1}\;\geq\;\pi^\top\boldsymbol\phi_t+\eta\delta
  -2\eta\bigl(\gamma^{\,t}\mathcal D_0+\bar{\mathcal D}\bigr).
\]
Summing over the rounds before $t$ and using $\sum_{s<t}\gamma^s\leq1/(1-\gamma)$,
\[
  \pi^\top\boldsymbol\phi_t\;\geq\;\pi^\top\boldsymbol\phi_0
  +\eta\bigl(\delta-2\bar{\mathcal D}\bigr)\,t-\frac{2\eta\,\mathcal D_0}{1-\gamma},
\]
which grows linearly in $t$ because $\delta>2\bar{\mathcal D}$.

\emph{Step 3: no agent falls behind.} For any two agents,
\[
  \phi_{v,t}-\phi_{u,t}
  =\bigl[\tilde f_{v,t}(y^*)-\tilde f_{u,t}(y^*)\bigr]
  -\bigl[\tilde f_{v,t}(y)-\tilde f_{u,t}(y)\bigr]
  \;\leq\;\osc_y\bigl(\tilde f_{v,t}-\tilde f_{u,t}\bigr)\;\leq\;2\mathcal D_t
\]
by \eqref{eq:D-controls}, so $|\phi_{v,t}-\pi^\top\boldsymbol\phi_t|\leq2\mathcal D_t
\leq2\bigl(\gamma^{\,t}\mathcal D_0+\bar{\mathcal D}\bigr)$. Combined with Step~2
this gives the bound in the statement. Finally
$q_{v,t}(y)=q_{v,t}(y^*)\,e^{-\phi_{v,t}}\leq e^{-\phi_{v,t}}$, so every
$q_{v,t}(y)$ with $y\neq y^*$ decays geometrically; summing over the finitely
many $y\neq y^*$ gives $1-q_{v,t}(y^*)\to0$, that is $q_{v,t}\to\delta_{y^*}$.
\proofend

\paragraph{Proof of Proposition~\ref{prop:case4-condorcet}.}

The idea is that near the atom the forcing in the log-odds recursion is
almost constant, so replacing it by its limit leaves a recursion that can be
solved in closed form. Its solution
has a term growing linearly at the rate $\pi^\top\mathbf d(y)$, which is the
margin, plus a fixed offset given by the deviation matrix, plus errors that
stay bounded. A positive margin therefore drives every agent's log-odds to
infinity and keeps the trajectory in the neighbourhood where the
approximation holds; a negative one drives them the other way.

Fix $y\neq y^*$ and keep the notation of the proof of
Proposition~\ref{prop:case4-atom}: $\boldsymbol\phi_t$ is the vector of
log-odds $\phi_{v,t}(y^*,y)$, and differencing \eqref{eq:logspace-vec} at
$y^*$ and $y$ gives $\boldsymbol\phi_{t+1}=A\boldsymbol\phi_t+\mathbf d_t(y)$
with
\[
  d_{v,t}(y)=\frac{R_v(y^*;\check q_{v,t})-R_v(y;\check q_{v,t})}{\beta_v}
  =\frac{1}{\beta_v}\sum_{y'}\check q_{v,t}(y')
   \bigl[P_v(y^*,y')-P_v(y,y')\bigr] .
\]
The recursion is already affine in $\boldsymbol\phi$; it is intractable because the forcing $\mathbf d_t(y)$ depends on the beliefs. Near
the atom that dependence is weak, so the forcing is replaced by its limit and
the difference carried as an error term. As $\check q_{v,t}\to\delta_{y^*}$ the bracket tends to $P_v(y^*,y^*)-P_v(y,y^*)=P_v(y^*,y)-\tfrac12$, so
$\mathbf d_t(y)\to\mathbf d(y)$ with $d_v(y)=(P_v(y^*,y)-\tfrac12)/\beta_v$, and
\[
  \|\mathbf d_t(y)-\mathbf d(y)\|_\infty\;\leq\;
  \frac{1}{\beta_{\min}}\,
  \max_u\|q_{u,t}-\delta_{y^*}\|_1 ,
\]
by Lemma~\ref{lem:signal-lip} applied twice: $d_{v,t}(y)$ is a difference of
two signals, at $y^*$ and at $y$, and each moves by at most
$\tfrac12\|\check q_{v,t}-\delta_{y^*}\|_1$.
Now iterate the exact recursion, adding and subtracting the frozen forcing:
\[
  \boldsymbol\phi_t=A^t\boldsymbol\phi_0
    +\sum_{s<t}A^{s}\mathbf d(y)
    +\sum_{s<t}A^{t-1-s}\bigl(\mathbf d_s(y)-\mathbf d(y)\bigr).
\]
The middle sum is the one with the forcing frozen. By the identity
$\sum_{s<t}A^s=t\,\one\pi^\top+(I-A^t)D$ from the proof of
Theorem~\ref{thm:case3}, and $\pi^\top\mathbf d(y)=\mathrm m(y^*,y)$, it equals
$t\,\mathrm m(y^*,y)\one+(I-A^t)D\,\mathbf d(y)$. So, exactly,
\[
  \boldsymbol\phi_t=t\,\mathrm m(y^*,y)\one+D\,\mathbf d(y)
    +A^t\bigl(\boldsymbol\phi_0-D\,\mathbf d(y)\bigr)+\mathbf e_t(y),
  \qquad
  \mathbf e_t(y)=\sum_{s<t}A^{t-1-s}\bigl(\mathbf d_s(y)-\mathbf d(y)\bigr).
\]
Each $A^t$ is stochastic, so every entry of the third term lies within $X_0$ of
zero, and every entry of $\mathbf e_t(y)$ within
$\sum_{s<t}\|\mathbf d_s(y)-\mathbf d(y)\|_\infty$.

Suppose $\mathrm m(y^*,y)>0$ for every $y\neq y^*$, and take $\mathrm m_0$, $C$,
$L_\epsilon$ and $L$ as in the statement. While every log-odds is at least
$L_\epsilon$, every belief satisfies
$\|q_{u,t}-\delta_{y^*}\|_1\leq2\sum_{y\neq y^*}e^{-\phi_{u,t}(y^*,y)}\leq2(|\Y|-1)e^{-L_\epsilon}$,
since $q_{u,t}(y)\leq e^{-\phi_{u,t}(y^*,y)}$, and the display above gives
$\|\mathbf d_t(y)-\mathbf d(y)\|_\infty\leq2(|\Y|-1)e^{-L_\epsilon}/\beta_{\min}\leq\mathrm m_0/2$.
Compare $\boldsymbol\phi_t$ with the solution of the recursion whose forcing is
$\mathbf d(y)-\tfrac12\mathrm m_0\one$, started at $\boldsymbol\phi_0$. The
recursion is monotone, because $A$ is non-negative, so $\boldsymbol\phi_t$ stays
above that solution as long as the trajectory has stayed in the neighbourhood.
By the decomposition, and $D\one=0$, the solution is
$t\bigl(\mathrm m(y^*,y)-\tfrac12\mathrm m_0\bigr)\one+D\,\mathbf d(y)
+A^t(\boldsymbol\phi_0-D\,\mathbf d(y))$, whose entries are at least
$\tfrac12t\,\mathrm m_0-C+(L-C)$. Hence, entrywise,
\[
  \boldsymbol\phi_t\;\geq\;\Bigl(L-2C+\tfrac12\,t\,\mathrm m_0\Bigr)\one .
\]
Since $L-2C\geq L_\epsilon$, induction on $t$ shows that the trajectory never
leaves the neighbourhood, which proves the lower bound at every $t$, and with it
the attraction. At round $s$ the lower bound gives
$\|\mathbf d_s(y)-\mathbf d(y)\|_\infty\leq Ke^{-s\,\mathrm m_0/2}$, so every entry
of $\mathbf e_t(y)$ is at most $K/(1-e^{-\mathrm m_0/2})$ in absolute value; with
the bound $X_0$ on the third term this gives the stated constant.

If instead $\mathrm m(y^*,y_0)=-\mathrm m_1<0$ for some $y_0$, take $L_\epsilon$ as
in the statement. While every log-odds is at least $L_\epsilon$, the same bound
gives $\|\mathbf d_t(y_0)-\mathbf d(y_0)\|_\infty\leq\tfrac12\mathrm m_1$, so
$\pi^\top\boldsymbol\phi_{t+1}=\pi^\top\boldsymbol\phi_t+\pi^\top\mathbf d_t(y_0)\leq\pi^\top\boldsymbol\phi_t-\tfrac12\mathrm m_1$,
the log-odds being those against $y_0$. Thus $\pi^\top\boldsymbol\phi_t$ decreases
at least linearly while the trajectory stays, whereas staying requires
$\pi^\top\boldsymbol\phi_t\geq L_\epsilon$. Every trajectory therefore leaves the
neighbourhood after finitely many steps, and the atom is repelling.
\proofend

\paragraph{At most one atom attracts.}
\label{app:one-atom}

Write $\mathrm{m}(y,y')=\eta\bigl[\bar P(y,y')-\tfrac12\bigr]$ for the influence-weighted margin of $y$ over $y'$, as in Proposition~\ref{prop:case4-condorcet}. Every kernel
satisfies $P_v(y,y')+P_v(y',y)=1$, so the weighted average does too and
$\mathrm{m}$ is antisymmetric, whether or not the agents share a kernel. Were
$\delta_y$ and $\delta_{y'}$ both attracting, the proposition would give
$\mathrm{m}(y,y')\geq0$ and $\mathrm{m}(y',y)\geq0$, which antisymmetry forces to zero,
the boundary case it leaves open. Away from that case the attracting atom is
unique, and with Proposition~\ref{prop:case4-fp}(b) the atoms are fully described: all $|\Y|$ of them are fixed points, at most one attracts, and the influence-weighted comparison decides which. Fixed points in the relative interior of a face of the simplex are not covered. When that comparison is cyclic,
none is.

\paragraph{The aggregate can cycle.}
\label{app:aggregate-cycles}

A kernel derived from a reward through Bradley--Terry can have no majority
cycle, since the reward orders the responses. The influence-weighted average
$\bar P$ of such kernels is still a preference kernel but is generically not
itself of Bradley--Terry form, because an average of logistic differences is
not a logistic difference, and it can therefore contain a cycle that none of its constituents does.

For an instance, take three responses and three agents of equal influence and
temperature whose rewards are the three cyclic shifts $(2,1,0)$, $(0,2,1)$ and
$(1,0,2)$. Each agent ranks the three strictly, yet averaging the kernels
gives $\bar P(y_0,y_1)=\bar P(y_1,y_2)=\bar P(y_2,y_0)\approx0.53$, so $y_0$
beats $y_1$ beats $y_2$ beats $y_0$.
The cardinal rule of Section~\ref{sec:case3} cannot represent this, and maximal lotteries are designed for it.

\paragraph{When no atom attracts.} When $\bar P$ compares every pair strictly and has no Condorcet winner, every atom repels, and the next two examples, whose kernels are themselves cyclic, show two things that can then
happen: the beliefs may converge to the interior maximal lottery, or the
collective policy may keep moving while its time average converges.

\begin{figure}[t]
\centering
\includegraphics[width=0.62\textwidth]{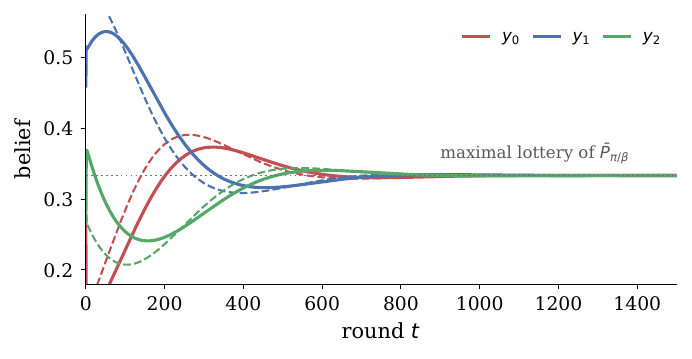}
\caption{A limit that is an interior maximal lottery; see
Example~\ref{ex:lottery}. Solid and dashed lines are the two agents.}
\label{fig:lottery}
\end{figure}

\begin{example}[An interior maximal lottery]
\label{ex:lottery}
Take three responses and two agents whose kernels are the two orientations of
the three-cycle: the first prefers $y_0$ to $y_1$, $y_1$ to $y_2$ and $y_2$ to
$y_0$, and the second prefers the reverse in each pair. The agents observe
each other and themselves, and their temperatures are $\tfrac12$ and $3$,
so by \eqref{eq:pbar} the influence-weighted kernel $\bar P$ favours the first
agent's orientation and is itself cyclic. No response then wins against the
others, and Proposition~\ref{prop:case4-condorcet} leaves every atom
repelling. The beliefs converge instead to the maximal lottery of $\bar P$,
here the uniform distribution, and do so without cycling.
Figure~\ref{fig:lottery} shows the trajectories.
\end{example}

\begin{figure}[htb]
\centering
\includegraphics[width=0.95\textwidth]{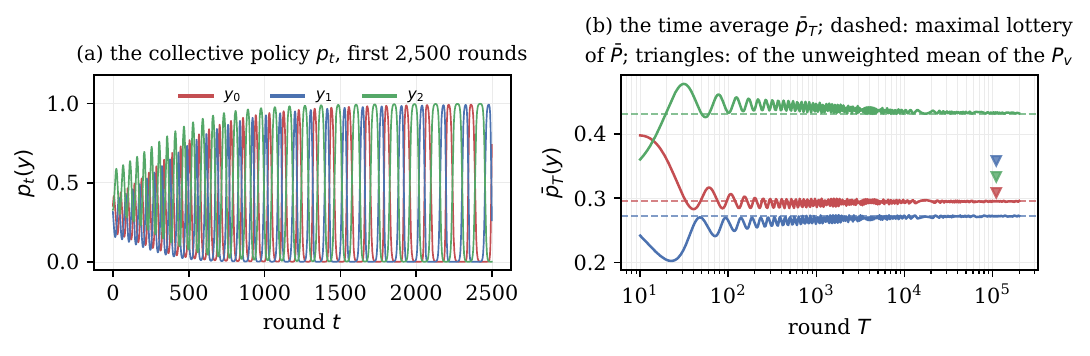}
\caption{The instance of Example~\ref{ex:maxlot}. \emph{(a)} The collective
policy over the first 2,500 rounds. \emph{(b)} Its time average, with the
maximal lottery of $\bar P$ dashed and that of the unweighted mean of the
$P_v$ marked by triangles.}
\label{fig:maxlot}
\end{figure}

\begin{example}[The average converges where the policy does not]
\label{ex:maxlot}
Take the graph of Section~\ref{sec:intro-example} in case~(III), with
temperatures $\beta=(0.4,1,2.5)$, the first at the central agent, and cyclic
kernels $P_v=\tfrac12\one\one^\top+S_v$ in which $y_0$ beats $y_1$, $y_1$ beats
$y_2$ and $y_2$ beats $y_0$, by margins $S_v(y_0,y_1)$, $S_v(y_1,y_2)$,
$S_v(y_2,y_0)$ of $(0.45,0.10,0.15)$ at the central agent and
$(0.10,0.40,0.15)$ and $(0.12,0.12,0.42)$ at the others. The
influence-weighted kernel $\bar P$ inherits the cycle, compares every pair
strictly and has no Condorcet winner, so every atom repels, by
Proposition~\ref{prop:case4-condorcet}. In the run of
Figure~\ref{fig:maxlot}(a) the collective policy does not settle: even over the last $2{,}000$ of $200{,}000$ rounds each coordinate still swings between nearly $0$ and nearly $1$. Its time
average nonetheless approaches the maximal lottery $(0.295,0.273,0.432)$ of
$\bar P$, to within $0.002$ in $\ell_1$ after $200{,}000$ rounds
(Figure~\ref{fig:maxlot}(b)). That is far closer than
Theorem~\ref{thm:cesaro} guarantees, its slack here including
$\eta/8=0.14$, so on this instance the bound is far from tight. Weighting by
influence over caution matters: the maximal lottery of the unweighted mean
$\tfrac13\sum_vP_v$ puts most mass on $y_1$, that of $\bar P$ on $y_2$.
\end{example}

On the instance of Section~\ref{sec:intro-example}, neither global result
applies. Unanimous dominance, Proposition~\ref{prop:case4-atom}, fails because
the central agent ranks $y_0$ first; aggregate dominance,
Proposition~\ref{prop:case4-aggdom}, fails because the column margin of $y_1$
in $\bar P$ is $0.041$, far below twice the residual disagreement unless the
temperature is very high. The convergence to $y_1$ in the panel of
Figure~\ref{fig:intro} with both roles aggregated is therefore covered by these
results only locally, by Proposition~\ref{prop:case4-condorcet}.

\subsection{Further results}
\label{supp:case4-further}

This subsection collects further results for Section~\ref{sec:case4}. The
first two concern particular kernels and boundaries: what degenerate kernels
do, and how a frozen boundary guarantees an interior fixed point, which
Proposition~\ref{prop:case4-fp} does not provide, while keeping interior
trajectories away from the atoms. The next two vary the graph, first to the uncoupled extreme in which
each agent faces only itself, then to the generalisation in which the two
roles need not use the same attention weights. 

\paragraph{Degenerate kernels.}
\label{app:case4-degenerate}

In the fully endogenous dynamics \eqref{eq:case4}, where both the reference
and the challenger are aggregated, the only nonlinearity is the dependence of
an agent's signal on the other agents' beliefs, and by Lemma~\ref{lem:degenerate} that is absent exactly when the
kernels are degenerate; the system is then affine and Theorem~\ref{thm:case3}
applies, with the score vectors $c_v$ in place of the rewards. Every kernel on
two responses is degenerate, so this covers every binary problem: with
$b_v=P_v(y_0,y_1)-\tfrac12$ the signal difference equals $b_v$ against every
challenger, the log-odds grow at $\sum_v(\pi_v/\beta_v)b_v$, and
\eqref{eq:case4-solv} holds exactly when that rate vanishes. With two responses
the only way to have no interior fixed point is for the network to commit to
one of them.

\paragraph{A frozen boundary guarantees existence.}
\label{sec:case4-frozen}
Without frozen agents, an interior fixed point of Section~\ref{sec:case4}
must satisfy the solvability condition \eqref{eq:case4-solv}. Because the
signals depend on the profile, that condition constrains the fixed point
rather than the data, and whether it can be met depends on the kernels:
interior fixed points exist for some and not for others. The condition is
forced by a single algebraic fact:
averaging leaves a quantity common to all agents unchanged, so
$(I-A)\one=0$, and Theorem~\ref{thm:fredholm} then requires the right-hand
side to be orthogonal to the left kernel before the equation has any solution.
Freezing some agents removes the fact and with it the condition. Restricted to the updating agents the operator is $I-\AUU$ with $\AUU$
substochastic, and $(I-\AUU)\one=\AUF\one>0$ for every agent that observes the
boundary. The constant vectors are no longer annihilated, $I-\AUU$ is
invertible, and the equation is solvable whatever the kernels and
temperatures.

\begin{proposition}[A frozen boundary guarantees existence]
\label{prop:case4-frozen}
Let $\Frz\neq\emptyset$ be frozen with interior beliefs, let every updating
agent reach $\Frz$, and let $\AUU,Z$ be as in Lemma~\ref{lem:substoch}. Then
\eqref{eq:case4} restricted to the updating agents has an interior fixed
point for every choice of kernels and temperatures, with no solvability
condition, and it satisfies
\begin{equation}\label{eq:case4-frozen}
  \mathbf f^*_{\Alg}
  =Z\bigl(\AUF\mathbf f_{\Frz}+B^{-1}\mathbf R^*\bigr) .
\end{equation}
If moreover
\[
  \rho\Bigl(\bigl(I+\tfrac12B_{\Alg}^{-1}\bigr)\AUU\Bigr)\;<\;1 ,
\]
the fixed point is unique and reached geometrically. With
$a^{\Alg}_v=\sum_{u\in\Alg}a_{vu}$ the weight agent $v$ places on the updating
agents, this holds in particular when every $a^{\Alg}_v<1$ and
\[
  \beta_v\;>\;\frac{a^{\Alg}_v}{2\bigl(1-a^{\Alg}_v\bigr)} \qquad\text{for every updating }v .
\]
The condition cannot be weakened to $\beta_v>a^{\Alg}_v$: the remark after the
proof gives an instance satisfying it with several fixed points.
\end{proposition}

\begin{proof}
\emph{The fixed-point equation.} Restricted to the updating agents, the
log-space equation reads
$\mathbf f_{\Alg}=\AUU\mathbf f_{\Alg}+\AUF\mathbf f_{\Frz}+B^{-1}\mathbf R^*$,
where $\AUF\mathbf f_{\Frz}$ collects the frozen beliefs, which do not change.
By Lemma~\ref{lem:substoch}, $\rho(\AUU)<1$, so $I-\AUU$ is invertible and
solving for $\mathbf f_{\Alg}$ gives \eqref{eq:case4-frozen}.

\emph{Why no solvability condition appears.} With every agent updating,
$(I-A)\one=0$, and Theorem~\ref{thm:fredholm} then requires the right-hand
side to be orthogonal to the left kernel. Here $\one$ is not in the kernel of
$I-\AUU$, since $(I-\AUU)\one=\AUF\one$ is positive in every coordinate
belonging to an agent that observes the boundary, so there is no singular
direction and no condition to satisfy.

\emph{Existence.} The right-hand side of
\eqref{eq:case4-frozen} is a continuous self-map of the product of simplices
over $\Alg$, so Brouwer's fixed-point theorem~\citep[Ch.~6]{border1985} gives a
fixed point.

\emph{Uniqueness and convergence.} Let $\mathbf f_t$ and $\mathbf g_t$ be two
trajectories of \eqref{eq:case4} on the updating agents, with the same frozen
beliefs, and write $m_{v,t}=\osc_y\bigl(g_{v,t}-f_{v,t}\bigr)$. The frozen terms
cancel, so
\[
  g_{v,t+1}-f_{v,t+1}
  =\sum_{u\in\Alg}a_{vu}\bigl(g_{u,t}-f_{u,t}\bigr)
  +\frac{1}{\beta_v}\Bigl(R_v\bigl(\cdot\,;\check q^{\,g}_{v,t}\bigr)-R_v\bigl(\cdot\,;\check q^{\,f}_{v,t}\bigr)\Bigr).
\]
The first term oscillates by at most $\sum_{u\in\Alg}a_{vu}m_{u,t}$. The
difference of the two challengers is
$\sum_{u\in\Alg}a_{vu}(q^g_{u,t}-q^f_{u,t})$, and by Lemma~\ref{lem:softmax} its
$\ell_1$ norm is at most $\tfrac12\sum_{u\in\Alg}a_{vu}m_{u,t}$.
By Lemma~\ref{lem:signal-lip} each signal therefore moves by at most
$\tfrac14\sum_{u\in\Alg}a_{vu}m_{u,t}$, and the second term oscillates by at
most $\frac{1}{2\beta_v}\sum_{u\in\Alg}a_{vu}m_{u,t}$. Hence, entrywise,
\[
  \mathbf m_{t+1}\;\leq\;K\,\mathbf m_t ,
  \qquad
  K=\bigl(I+\tfrac12B_{\Alg}^{-1}\bigr)\AUU .
\]
Since $K$ is non-negative with $\rho(K)<1$, the vector
$\mathbf z=(I-K)^{-1}\one=\sum_{k\geq0}K^k\one$ is finite with $\mathbf z\geq\one$,
and $K\mathbf z=\mathbf z-\one\leq\gamma\mathbf z$ with
$\gamma=1-1/\|\mathbf z\|_\infty<1$. So $m_{u,t}\leq Dz_u$ for every $u$ implies
$m_{v,t+1}\leq\gamma Dz_v$ for every $v$. Taking $\mathbf g_t$ to be a fixed
point gives $m_{v,t}\leq\gamma^tD_0z_v$, with $D_0=\max_um_{u,0}/z_u$, and hence
$\|q_{v,t}-q^*_v\|_1\leq\tfrac12\gamma^tD_0z_v$ by Lemma~\ref{lem:softmax}: the
fixed point is reached geometrically. Taking both to be fixed points, the
largest ratio $D=\max_um_u/z_u$ satisfies $D\leq\gamma D$, so $D=0$; the two
profiles of log-beliefs then differ at each agent by a constant, and give the
same beliefs. Under the explicit condition one may take $\mathbf z=\one$ and
$\gamma=\max_va^{\Alg}_v\bigl(1+\tfrac{1}{2\beta_v}\bigr)$, which is below one
exactly when $\beta_v>a^{\Alg}_v/\bigl(2(1-a^{\Alg}_v)\bigr)$ for every $v$.

\end{proof}

\begin{remark}[The condition $\beta_v>a^{\Alg}_v$ does not suffice]
\label{rem:frozen-several}
Take two updating agents that attend to each other with weight $a=0.97$ and
to one frozen agent, holding the uniform belief, with weight $0.03$. Let them
hold opposite cyclic kernels on three responses,
$P_1=\tfrac12\one\one^\top+\tfrac12C$ and $P_2=\tfrac12\one\one^\top-\tfrac12C$, with
$C$ the cyclic matrix of Proposition~\ref{prop:case2-below}, and give both the
temperature $\beta=1$. Then $\beta_v>a^{\Alg}_v$ at both agents, while
$\rho(K)=1.455$. A numerical solution of the fixed-point equations finds at
least seven interior fixed points: the uniform profile; the profile with
$q_1=(0.2146,0.2492,0.5362)$ and $q_2=(0.2492,0.2146,0.5362)$; the profile
with $q_1=(0.0379,0.5333,0.4288)$ and $q_2=(0.0379,0.4288,0.5333)$; and the
images of the last two under a cyclic relabelling of the responses. Each
satisfies the fixed-point equation of \eqref{eq:case4} to within $10^{-10}$.
Run from random initial profiles, the dynamics settle at different limits.
\end{remark}

So a boundary converts a case in which interior fixed points exist only where the weighted signal is flat, by Proposition~\ref{prop:case4-spread}(b), into one in which they always do, and \eqref{eq:case4-frozen} determines where each agent ends up: the boundary's positions carried in through the harmonic matrix
$H=Z\AUF$ of Lemma~\ref{lem:substoch}, whose row $v$ is the distribution over
frozen agents at which a walk from $v$ stops, plus the quality signal
accumulated along the way through $Z$, exactly as in Theorem~\ref{thm:case3-partial} but with
the signal now endogenous.

The collective rule is lost. The limits are agent-specific, each weighted by its own row of $H$, so a
network with a boundary aggregates nothing across itself: it delegates to the boundary. The construction that makes this case well behaved is therefore the one that
removes the object it was introduced to study. This construction appears in practice without being named as such: the
iterative alignment schemes that include fixed policies among the opponents, such as earlier checkpoints or external models, alongside the current iterate, are holding agents frozen in exactly this sense~\citep{wu2025mnpo}.

The smallest instance that shows the aggregation has two updating agents
and one frozen; the example below works it through, and there the
limit exponentiates to the Gibbs form \eqref{eq:gibbs} with the frozen belief
as the reference and a weighted sum of \emph{both} updating agents' scores as
the reward.

\begin{example}[Two agents behind a boundary]
\label{ex:frozen-hub}
The smallest instance that shows the aggregation has two updating agents and
one frozen. Let each updating agent place weight $a$ on the frozen agent
$F$ and $1-a$ on the other, so that
$\AUU=(1-a)\left(\begin{smallmatrix}0&1\\1&0\end{smallmatrix}\right)$ and
$\AUF=a\one$. Then
\[
  Z=(I-\AUU)^{-1}=\frac{1}{a(2-a)}
    \begin{pmatrix}1&1-a\\1-a&1\end{pmatrix},
  \qquad H=Z\AUF=\one ,
\]
so each agent's reference is the frozen belief in full. Take degenerate
kernels, $P_v(y,y')=\tfrac12+c_v(y)-c_v(y')$, so that $R^*_v=c_v$ up to a constant. Then \eqref{eq:case4-frozen}
exponentiates to the Gibbs form \eqref{eq:gibbs}:
\[
  q^*_1(y)\;\propto\;q_F(y)\,
  \exp\Bigl(\frac{1}{a(2-a)}
    \Bigl[\frac{c_1(y)}{\beta_1}+(1-a)\frac{c_2(y)}{\beta_2}\Bigr]\Bigr),
\]
and symmetrically for agent~2 with the roles of $c_1$ and $c_2$ exchanged.
The frozen belief is the reference and the bracketed sum is the reward. Each agent ends at the frozen belief tilted by
a weighted sum of \emph{both} agents' scores, its own with weight $1$ and its neighbour's with $1-a$, all scaled by $1/(a(2-a))$: attending less to the
boundary, that is smaller $a$, amplifies both tilts.

\end{example}

The interior fixed point does not displace the atoms, which remain fixed points as long as every updating agent places some weight on the updating agents, its own self-weight included; the boundary only prevents interior trajectories from reaching them.

\begin{proposition}[Atoms under a frozen boundary]
\label{prop:case4-frozen-atoms}
Under the assumptions of Proposition~\ref{prop:case4-frozen}:
\begin{enumerate}[label=(\alph*),leftmargin=*]
\item if every updating agent places positive weight on the updating agents,
$a^{\Alg}_v>0$, then for each $y\in\Y$ the profile in which every updating
agent holds $\delta_y$ is a fixed point of \eqref{eq:case4} restricted to the
updating agents;
\item no trajectory started from an interior profile approaches any such
profile: there is $c>0$, depending on the initial profile and on the instance, with
$\|q_{v,t}-\delta_y\|_1\geq c$ for every updating agent $v$, every $y\in\Y$
and every $t$.
\end{enumerate}
\end{proposition}

\begin{proof}
\emph{(a).} Let every updating agent hold $\delta_y$. Since $a^{\Alg}_v>0$,
agent $v$ places positive weight on some updating agent, whose belief
vanishes off $y$, so its reference $\bar q_v\propto\prod_uq_u^{a_{vu}}$
vanishes off $y$ whatever the frozen beliefs, and so does
$q_v\propto\bar q_v\,e^{R_v/\beta_v}$. Hence $q_v=\delta_y$ again.

\emph{(b).} Fix $y_0\in\Y$ and write $x_{v,t}(y)=f_{v,t}(y)-f_{v,t}(y_0)$ for the
log-odds of an updating agent, which removes the normalisers; the beliefs
stay interior, so these are finite. For each $y$ the log-space update
restricted to the updating agents reads
\[
  \mathbf x_{t+1}(y)=\AUU\mathbf x_t(y)+\mathbf g_t(y),
  \qquad
  \mathbf g_t(y)=\AUF\mathbf x_{\Frz}(y)+B^{-1}\bigl[\mathbf R_t(y)-\mathbf R_t(y_0)\bigr],
\]
where $\mathbf x_{\Frz}(y)$ collects the frozen agents' log-odds, finite
because their beliefs are interior. The signals lie in $[0,1]$, so every entry
of $\mathbf g_t(y)$ is at most
$G(y)=\|\AUF\mathbf x_{\Frz}(y)\|_\infty+1/\beta_{\min}$ in absolute value.
Iterating, and using that $\AUU$ has non-negative entries, gives entrywise
\[
  |\mathbf x_t(y)|
  \;\leq\;\AUU^{\,t}|\mathbf x_0(y)|+\sum_{s<t}\AUU^{\,t-1-s}|\mathbf g_s(y)|
  \;\leq\;\|\mathbf x_0(y)\|_\infty\one+G(y)\,Z\one ,
\]
since $\AUU^{\,t}\one\leq\one$ for a substochastic matrix and
$\sum_{k<t}\AUU^{\,k}\leq Z$ by Lemma~\ref{lem:substoch}(a). Hence
$|x_{v,t}(y)|\leq K$ for every $v$, $y$ and $t$, with
$K=\max_y\bigl(\|\mathbf x_0(y)\|_\infty+G(y)\|Z\one\|_\infty\bigr)$. Then
$q_{v,t}(z)=e^{x_{v,t}(z)}/\sum_we^{x_{v,t}(w)}\geq e^{-2K}/|\Y|$ for every $z$,
and for every $y$
\[
  \|q_{v,t}-\delta_y\|_1=2\bigl(1-q_{v,t}(y)\bigr)\;\geq\;\frac{2(|\Y|-1)}{|\Y|}\,e^{-2K}=c>0 .
\]
\end{proof}

\paragraph{The uncoupled case: self-play.}
\label{app:uncoupled}

Setting $A=I$ leaves each agent facing only itself, so \eqref{eq:case4}
becomes multiplicative weights in self-play. Two known facts about that iteration describe what happens in this limit. Its iterates do not approach an interior equilibrium, since by \citet{bailey2018} the divergence from one is non-decreasing at any constant step size; its time average does converge, to the set of maximal lotteries. That is the setting \citet{freund1999} study: two
players of a zero-sum game each update by multiplicative weights, and because the update has regret $O(\sqrt T)$ against any fixed action when its step size is tuned to the horizon, the average of
play over $T$ rounds approaches the value of the game, which in this case is the
symmetric one with payoff $P_v-\tfrac12\one\one^\top$ and the agent plays both
sides, so its optimal strategies are the maximal lotteries of $P_v$, and the
regret bound places the time average within $O(1/t)$ of that set. Both are properties of the single-agent algorithm, not of the network, and we claim neither; the proposition contributes the identification of \eqref{eq:case4} at $A=I$ with that
algorithm.

In case~(II), by contrast, the uncoupled system has, when $\beta_v>\tfrac12$, a unique attracting fixed point at each agent, the regularised Nash policy $p^*_v$, and the implicit function theorem could
continue it to small $a$. The uncoupled system has no attracting fixed
point at all, so there is no fixed point to continue, and an expansion in the weight each agent places on others does not exist. The structural cause is
the one identified in Section~\ref{sec:answers}: with the reference aggregated
rather than fixed there is no anchor, the KL penalty is measured against a moving
target, and the effective regularisation vanishes. The phenomenon is the known non-convergence of multiplicative weights in zero-sum games, and nothing peculiar to a network, and
Theorem~\ref{thm:cesaro} shows that the time-average conclusion (c) extends to the coupled case, in the approximate form \eqref{eq:cesaro}.

\paragraph{Separate attention graphs.}
\label{app:two-graphs}

The model gives the two aggregates the same weights. The fixed-point theory
does not need that, as this section shows, although the convergence argument
of Theorem~\ref{thm:case4-directed} does; and an agent may well anchor to one
set of neighbours and be measured against another. In a distillation pipeline the teachers an agent
copies need not be the judges it is scored by; in \textsc{rlhf} the reference
policy is a checkpoint while the comparisons come from an annotator pool. Let
$A^{\mathcal R}$ and $A^{\mathcal C}$ both satisfy
Assumption~\ref{ass:graph}, and replace the two aggregates of
Section~\ref{sec:aggregates} by
\[
  \bar q_{v,t}\propto\prod_u q_{u,t}^{\,a^{\mathcal R}_{vu}},
  \qquad
  \check q_{v,t}=\sum_u a^{\mathcal C}_{vu}q_{u,t} .
\]
Write $\pi^{\mathcal R}$ and $D^{\mathcal R}$ for the stationary distribution and
deviation matrix of $A^{\mathcal R}$.

\begin{proposition}[The two roles may use different graphs]
\label{prop:two-graphs}
Assume that the reference is aggregated over a graph $A^{\mathcal R}$ and the
challenger over a graph $A^{\mathcal C}$, each satisfying
Assumption~\ref{ass:graph}.
\begin{enumerate}[label=(\alph*),leftmargin=*]
\item Cases~(I) and~(II) are as before, on their own graph: replace $A$ by
$A^{\mathcal R}$ throughout Section~\ref{sec:case3}, and by
$A^{\mathcal C}$ throughout Section~\ref{sec:case2}.
\item In case~(III), Propositions~\ref{prop:case4-fp}
and~\ref{prop:case4-spread} hold with $\pi$ and $D$ replaced by
$\pi^{\mathcal R}$ and $D^{\mathcal R}$ throughout, and with
$R^*_v=R_v(\cdot\,;\check q^*_v)$ evaluated at the
$A^{\mathcal C}$-challenger. In particular the collective policy is the
$\pi^{\mathcal R}$-weighted geometric mean, and the solvability condition
\eqref{eq:case4-solv} reads
$\sum_v(\pi^{\mathcal R}_v/\beta_v)R^*_v(y)=c$ on the support.
\end{enumerate}
\end{proposition}

\begin{proof}
For~(a), case~(I) never forms the challenger aggregate and case~(II) never
forms the reference aggregate, so each case sees one matrix and the arguments
are unaltered. For~(b), the proof of Proposition~\ref{prop:case4-fp} uses the
two aggregates only through $\sum_ua_{vu}=1$, which holds for both matrices,
and through the vanishing of $\bar q^*_v$ off the support, which involves
$A^{\mathcal R}$ alone; aggregating the pairwise condition with weights
$\pi^{\mathcal R}_v$ then gives the stated form. The proof of
Proposition~\ref{prop:case4-spread} uses row-stochasticity of the reference
weights to propagate the support along $A^{\mathcal R}$, and solves
$(I-A^{\mathcal R})\mathbf f^*=B^{-1}\mathbf R^*$ with the deviation matrix of
$A^{\mathcal R}$; the challenger enters only inside $\mathbf R^*$.
\end{proof}

The division of labour is then that the \emph{reference} graph supplies the weights, through $\pi^{\mathcal R}$ and $D^{\mathcal R}$, and the
\emph{challenger} graph supplies the signals those weights are applied to.
The convergence theory does not carry over as readily:
Theorem~\ref{thm:case4-directed} compares the signals two agents receive, and
with different challenger weights that comparison acquires a term in
$\tau_1(A^{\mathcal C})$ alongside the contraction at $\tau_1(A^{\mathcal R})$.
What the sharp rate is, and whether a temperature threshold is needed at all when the two graphs differ, we leave open.

\section{Graphs outside Assumption~\ref{ass:graph}}
\label{app:cross}

\begin{figure}[t]
\centering
\includegraphics[width=0.62\textwidth]{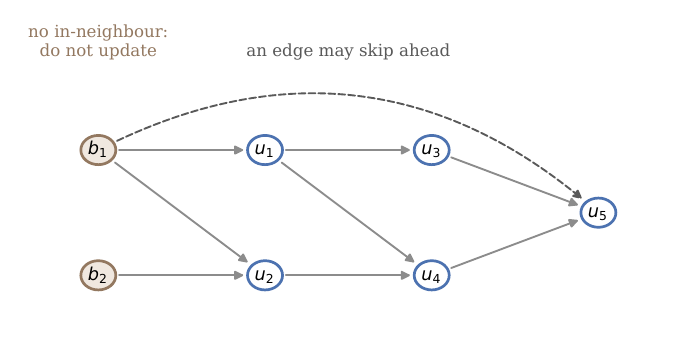}
\caption{A cascade. The arrows point along the direction influence flows, and the shaded
agents, having no in-neighbour, do not update.
Proposition~\ref{prop:layered} assumes only acyclicity, so an edge
may skip ahead, as the dashed one does, and the paths reaching an agent may
differ in length.}
\label{fig:layered}
\end{figure}

Assumption~\ref{ass:graph} asks the attention matrix to be primitive, which
makes influence circulate and gives a single stationary distribution $\pi$.
This section treats two departures from it. The first is a general
row-stochastic $A$, where the graph decomposes into closed components and the
existing results apply within each. The second is a cascade, in which
influence flows one way and, as Figure~\ref{fig:layered} shows, none of the
four dynamics induces a collective rule.

\subsection{Reducible graphs}
\label{app:reducible}

Assumption~\ref{ass:graph} and the cascades below are the two extremes: in
the first every agent reaches every other, in the second none reaches back.
This subsection treats a general row-stochastic $A$, of which both are special
cases, and shows that the graph decomposes so that the existing results apply
component by component.

Partition the agents into strongly connected components and call a component
\emph{closed} if no agent in it places weight outside it, \emph{transient}
otherwise. Write $\mathcal C_1,\dots,\mathcal C_K$ for the closed components,
$T$ for the transient agents, $A_{TT}$ for the block of $A$ among them, and
\[
  Z=(I-A_{TT})^{-1},\qquad
  h_{vk}=\Bigl[Z\,A_{T\mathcal C_k}\one\Bigr]_v ,
\]
so that $h_{vk}$ is the probability that the walk of
Lemma~\ref{lem:substoch}, started at $v$, is absorbed in $\mathcal C_k$;
$\sum_kh_{vk}=1$ because every transient agent reaches some closed component.

\begin{proposition}[Reducible graphs decompose]
\label{prop:reducible}
Assume that $A$ is row-stochastic and that each closed component is primitive.
\begin{enumerate}[label=(\alph*),leftmargin=*]
  \item Each closed component evolves independently of everything outside it,
        and Assumption~\ref{ass:graph} holds within it, so the results of
        Sections~\ref{sec:case3} to~\ref{sec:case4} apply to $\mathcal C_k$
        with its own stationary distribution $\pi^{(k)}$, computed from the
        sub-block of $A$ on $\mathcal C_k$.
  \item Under an aggregated reference and a fixed challenger, write
        \[
          \bar r_k(y)=\sum_{v\in\mathcal C_k}\frac{\pi^{(k)}_v}{\beta_v}\,r_v(y)
        \]
        for the influence-weighted reward of $\mathcal C_k$, and
        $S_k=\bar r_k(y)-\bar r_k(y')$ for the drift inside it. Every transient agent's log-odds
        drift at the $H$-weighted average of these,
        \[
          \phi_{v,t}=t\sum_{k}h_{vk}S_k+O(1),\qquad v\in T ,
        \]
        and $v$ concentrates on the maximiser of $\sum_kh_{vk}\bar r_k$ when it is unique.
  \item Under a fixed reference, Propositions~\ref{prop:case2-fp} and~\ref{prop:case2-geom} apply to $A$ as it is, since neither assumes Assumption~\ref{ass:graph}.
\end{enumerate}
\end{proposition}

Parts~(b) and~(c) cover the dynamics with a fixed challenger and with a fixed
reference. Case~(III) is covered by~(a) within each closed component and is
otherwise open: a transient agent's limit is not characterised there, because
the components it inherits from need not converge, and when they do the
inheritance is not linear. 

\begin{proof}
(a) The agents of a closed component place no weight outside it, by definition, so its sub-block is row-stochastic and its dynamics are those of
\eqref{eq:general} on $\mathcal C_k$ alone; primitivity is the hypothesis.
(b) Every transient agent reaches some closed component, so $A_{TT}$ is
substochastic with $\rho(A_{TT})<1$ by Lemma~\ref{lem:substoch}, and $Z$ and
$H$ are defined. Theorem~\ref{thm:case3-partial} does not apply directly,
because the boundary beliefs here move, so we argue
from the recursion. Write $\boldsymbol\phi_{T,t}$ for the transient log-odds
and $\boldsymbol\phi_{C,t}$ for those of the closed components. The transient
block obeys
\[
  \boldsymbol\phi_{T,t+1}
  =A_{TT}\boldsymbol\phi_{T,t}+A_{TC}\boldsymbol\phi_{C,t}
  +B_T^{-1}\Delta\mathbf r_T ,
\]
and by (a) with Theorem~\ref{thm:case3},
$\boldsymbol\phi_{C,t}=t\,\mathbf s+\mathbf c+\boldsymbol\epsilon_t$ with
$\mathbf s$ the vector whose entry for an agent of $\mathcal C_k$ is $S_k$,
$\mathbf c$ constant and $|\boldsymbol\epsilon_t|\leq E\one$ entrywise for some
$E$; the error tends to zero geometrically, but only its boundedness is used.
Put $\mathbf w=ZA_{TC}\mathbf s$ and
$\mathbf b=Z\bigl(A_{TC}\mathbf c+B_T^{-1}\Delta\mathbf r_T-\mathbf w\bigr)$. Then
$\mathbf w=A_{TT}\mathbf w+A_{TC}\mathbf s$ and
$\mathbf w+\mathbf b=A_{TT}\mathbf b+A_{TC}\mathbf c+B_T^{-1}\Delta\mathbf r_T$,
so $t\,\mathbf w+\mathbf b$ solves the transient recursion with
$\boldsymbol\epsilon_t$ removed, and the deviation
$\mathbf e_t=\boldsymbol\phi_{T,t}-(t\,\mathbf w+\mathbf b)$ satisfies
\[
  \mathbf e_{t+1}=A_{TT}\mathbf e_t+A_{TC}\boldsymbol\epsilon_t .
\]
Let $\mathbf z=Z\one$, so that $\mathbf z=\one+A_{TT}\mathbf z$ and
$\mathbf z\geq\one$. The rows of $A_{TT}$ and $A_{TC}$ are non-negative and sum
to at most one, so induction on $t$ gives, entrywise,
\[
  |\mathbf e_t|\;\leq\;X_0\one+E\,\mathbf z,
  \qquad X_0=\max_{v\in T}|e_{v,0}| ,
\]
since $A_{TT}(X_0\one+E\mathbf z)+EA_{TC}\one\leq X_0\one+E(A_{TT}\mathbf z+\one)$.
Hence $\boldsymbol\phi_{T,t}=t\,\mathbf w+O(1)$. As $\mathbf s$ is constant on
each component, $w_v=[ZA_{TC}\mathbf s]_v=\sum_kh_{vk}S_k$, giving
$\phi_{v,t}=t\sum_kh_{vk}S_k+O(1)$. Concentration on the maximiser then
follows as in Theorem~\ref{thm:case3}. (c) Neither Proposition~\ref{prop:case2-fp} nor Proposition~\ref{prop:case2-geom} assumes Assumption~\ref{ass:graph}.
\end{proof}

A reducible graph therefore induces one rule per closed component, so no single influence measure ranks
all the agents; and a transient agent implements a mixture of those rules
with weights $H_{v\cdot}$ fixed by the topology, which is the delegation
reading of the cascades below with closed components in place of a frozen
frontier. Cascades are the case $K=|\Frz|$ with every closed component a
single frozen agent, and Assumption~\ref{ass:graph} the case $K=1$ with
$T=\emptyset$.

\subsection{Cascades: when influence flows one way}
\label{app:cascade}
\label{sec:layered}

This subsection treats a graph in which each agent observes only agents upstream of it, the arrangement of a distillation cascade. There is then no stationary distribution common to the network, so no single rule
governs it; what replaces one is an exact solution agent by agent, each with its
own weights, given by the directed paths reaching it.

\begin{figure}[t]
\centering
\includegraphics[width=0.62\textwidth]{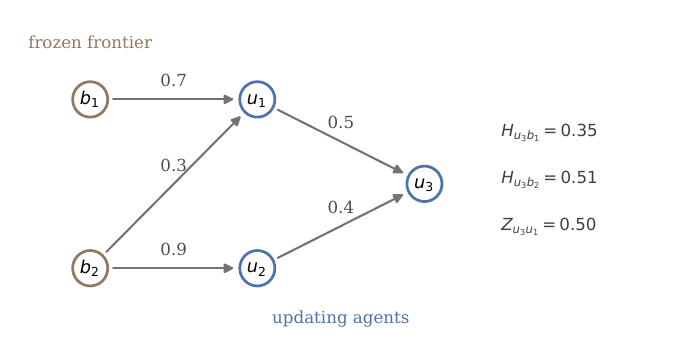}
\caption{The two matrices of Lemma~\ref{lem:substoch} on a small cascade,
arrows pointing along the direction influence flows and labelled with the attention paid. $h_{vb}$ collects the weight of all directed paths from $b$
to $v$: agent $u_3$ inherits $b_1$ with weight $0.5\times0.7=0.35$ and $b_2$
with weight $0.5\times0.3+0.4\times0.9=0.51$. $z_{vu}$ counts expected visits,
so $Z_{u_3u_1}=0.5$ is the attention $u_3$ pays $u_1$ along the single path
between them.}
\label{fig:paths}
\end{figure}

Figure~\ref{fig:paths} shows both matrices on a small cascade, where $Z$
collects the paths among the updating agents and $H$ the attention each pays
to the frozen ones.

\begin{proposition}[Cascades are solvable]
\label{prop:layered}
Assume that $G$ has no directed cycle through two or more distinct agents, while
each agent may place weight $a_{vv}\geq0$
on the belief it already holds. Let $\Frz$ be the set of agents with no in-neighbour other than
themselves and suppose every agent in $\Frz$ is frozen. For an updating
agent $v$ write
\[
  a_v=\sum_{u\neq v}a_{vu}=1-a_{vv}\;\in(0,1]
\]
for the weight it places on others, let $d(v)$ be the number of edges on the
longest directed path of distinct agents ending at $v$, and put
$L=\max_vd(v)$. Then:
\begin{enumerate}[label=(\alph*),leftmargin=*]
  \item \emph{(Forward substitution.)} For each of the four dynamics of
        Section~\ref{sec:fourcases} and every vector of temperatures, the
        fixed point is unique and obtained in any topological order: once the
        limits of the agents $v$ observes other than itself are known,
        $q^*_v$ is the unique solution of \eqref{eq:general} at $v$, with both
        aggregates evaluated on them and on $q^*_v$ itself.
  \item \emph{(Time to the limit.)} For the two dynamics with a fixed
        challenger, the limit is approached geometrically, at the rate
        $\rho(\AUU)^t=(1-\min_va_v)^t$ up to a factor polynomial in $t$. If no
        agent places weight on itself, so that $a_v\equiv1$, the limit is
        attained \emph{exactly} after $L$ rounds, for all four dynamics.
  \item \emph{(No hypotheses on temperature.)} Neither the threshold
        $\beta_v>\tfrac12$ of Proposition~\ref{prop:case2-fp}(b), nor the
        solvability condition \eqref{eq:case4-solv}, nor the contraction
        condition \eqref{eq:case4-threshold}, nor the unanimity condition
        \eqref{eq:unanimity} is required for~(a) or for the convergence
        stated in~(b).
  \item \emph{(Explicit visit counts and harmonic measure.)} The fundamental
        matrix of Lemma~\ref{lem:substoch} has entries
        \begin{equation}\label{eq:cascade-Z}
          z_{vu}=\frac{\mathbb P\bigl(\text{the walk from }v\text{ ever visits }u\bigr)}{a_u},
        \end{equation}
        and $h_{vb}$ is, as always, the probability that the walk from $v$ stops
        at $b\in\Frz$, so each row of $H$ is a distribution over the frontier.
        When $a_v\equiv1$ the walk visits each agent it reaches exactly once,
        $\AUU$ is nilpotent with $\AUU^{L}=0$, and $z_{vu}$ reduces to the total
        weight of the directed paths from $u$ to $v$; in a graph layered so that
        every edge joins consecutive layers, the rows of $H$ indexed by the
        $k$th layer are then the products $W_kW_{k-1}\cdots W_2$ of the
        inter-layer weight matrices.
  \item \emph{(Explicit limit, aggregated reference.)} For the two dynamics
        whose reference is aggregated, cases (I) and (III) of
        Section~\ref{sec:fourcases}, fix a reference response $y_0$ and write
        $\phi_v(y)=\log\bigl(q^*_v(y)/q^*_v(y_0)\bigr)$. Then for every
        updating $v$,
        \begin{equation}\label{eq:layered-explicit}
          \phi_v=\sum_{u\in\Alg}z_{vu}\,g_u+\sum_{b\in\Frz}h_{vb}\,\phi_b,
          \qquad
          g_u(y)=\frac{R_u(y;\check q^*_u)-R_u(y_0;\check q^*_u)}{\beta_u},
        \end{equation}
        a sum over the directed paths reaching $v$ of the frontier log-odds and
        of the quality signals met along the way. With an exogenous signal
        $g_u=(r_u-r_u(y_0))/\beta_u$ is a constant and
        \eqref{eq:layered-explicit} is a closed form in the data. With an
        endogenous signal $g_u$ depends on the limits of the agents $u$
        observes; when no agent places weight on itself these are supplied by
        \eqref{eq:layered-explicit} in topological order, and otherwise $g_u$
        also depends on $q^*_u$, which~(a) determines.
\end{enumerate}
\end{proposition}

\noindent
With an aggregated challenger and positive self-weights, (b) asserts nothing
about convergence, and the iteration need not converge: in our runs with a
single updating agent attending to one frozen agent, self-weight $0.9$, a
cyclic kernel and $\beta=0.2$, the iterates keep moving across the simplex,
in case~(II) and in case~(III), although the fixed point is unique.

\begin{proof}
Because no directed cycle passes through two distinct agents, the agents
admit a topological order in which every in-neighbour of $v$ other than $v$
itself precedes it. For an updating $v$, both $\bar q_{v,t}$ and
$\check q_{v,t}$ are formed from $q_{v,t}$ and from the beliefs of agents
preceding $v$, so the fixed-point equation for $v$ involves only $v$ and its
predecessors. Fix the predecessors' limits.

\emph{(a) and (c).} With a fixed challenger, or with $a_{vv}=0$, the signal
$R_v(\cdot\,;\check q_v)$ is then known, and the equation at $v$ determines
$q^*_v$ directly: with a fixed reference it is the Gibbs policy of $\mu_v$,
and with an aggregated one its log-odds form
$\phi_v=\sum_{u\neq v}a_{vu}\phi_u+a_{vv}\phi_v+g_v$ gives
$a_v\phi_v=\sum_{u\neq v}a_{vu}\phi_u+g_v$, which determines $\phi_v$ since
$a_v>0$. With an aggregated challenger and $a_{vv}>0$ the signal depends on
$q_v$ itself, and the equation at $v$ reads
\[
  \log q_v=\kappa_vP_vq_v+\mathbf c_v+\lambda\one ,
\]
with $\kappa_v=a_{vv}/(a_v\beta_v)$ for an aggregated reference and
$\kappa_v=a_{vv}/\beta_v$ for a fixed one, $\mathbf c_v$ fixed by the
predecessors and $\lambda$ a normaliser. A solution exists by Brouwer's
theorem~\citep[Ch.~6]{border1985}, applied to the continuous map
$q\mapsto\sigma(\kappa_vP_vq+\mathbf c_v)$ of $\Delta(\Y)$ into itself. If $q$
and $q'$ are two solutions, then
$\log q-\log q'=\kappa_vP_v(q-q')+(\lambda-\lambda')\one$, and taking the inner
product with $d=q-q'$, which sums to zero, gives
\[
  \sum_y\bigl(\log q(y)-\log q'(y)\bigr)\bigl(q(y)-q'(y)\bigr)
  =\kappa_v\,d^\top P_vd=\kappa_vB_{P_v}(d,d)=0 ,
\]
by the antisymmetry of $B_{P_v}$ shown in Section~\ref{supp:degenerate}. Each
term on the left is non-negative and vanishes only where $q(y)=q'(y)$, so
$q=q'$. Solving in topological order therefore determines every limit
uniquely, which is~(a). No step used a bound on any $\beta_v$ or a condition
on the kernels, which is~(c) for~(a).

\emph{(b).} With a fixed challenger the recursion on the updating block is
affine, with matrix $\AUU$ for an aggregated reference and none for a fixed
one. In the topological order $\AUU$ is triangular with diagonal entries
$a_{vv}=1-a_v$, so its spectrum is $\{1-a_v\}$ and
$\rho(\AUU)=1-\min_va_v<1$; the iteration converges at that rate, up to a
factor polynomial in $t$ from repeated diagonal entries, by
Lemma~\ref{lem:substoch}. If every $a_v=1$ the diagonal vanishes and $\AUU$
is nilpotent; a walk then leaves each agent immediately, so it visits agents
of strictly decreasing depth and $\AUU^{L}=0$. Exactness at round $d(v)$
follows by induction on the depth, for all four dynamics, since with no
self-weight an agent's belief at round $t+1$ is a function of its
predecessors' beliefs at round $t$ alone. The frozen agents never move, so a
agent at depth $1$ sees only them and attains the value assigned by~(a) after
one round; and if every agent of depth below $d$ has attained its value by
round $d-1$, then an agent at depth $d$, which observes only such agents,
attains its own at round $d$. No temperature condition was used, which
is~(c) for~(b).

\emph{(d).} $z_{vu}$ is the expected number of visits to $u$ before absorption
for the walk of Lemma~\ref{lem:substoch}. Each visit to $u$ is followed by a
further visit to $u$ exactly when the walk stays at $u$, which has probability
$a_{uu}=1-a_u$, and no other return is possible because a return would
traverse a cycle through distinct agents. The number of visits, given that
$u$ is reached at all, is therefore geometric with success probability $a_u$
and mean $1/a_u$, which is \eqref{eq:cascade-Z}; and $H=Z\AUF$ read entrywise
is the absorption probability, as in Lemma~\ref{lem:substoch}. When
$a_v\equiv1$ the geometric factor is $1$ and the reach probability is the
total weight of the directed paths from $u$ to $v$; in the layered case a path
from the $k$th layer to the frontier has exactly one edge per layer, so the
surviving term of $H$ is the product $W_kW_{k-1}\cdots W_2$.

\emph{(e).} With an aggregated reference the log form of \eqref{eq:general}
gives
$(I-\AUU)\boldsymbol\phi_{\Alg}=\mathbf g_{\Alg}+\AUF\boldsymbol\phi_{\Frz}$
on the updating block after differencing at $y$ and $y_0$ to remove the
normalisers; applying $Z=(I-\AUU)^{-1}$ yields \eqref{eq:layered-explicit}.
The readings of $g_u$ are immediate from $R_u(\cdot\,;\nu_u)=r_u$ in the
exogenous case and from~(a) in the endogenous one.
\end{proof}

\end{document}